\documentclass[runningheads]{llncs}

\usepackage{xcolor}

\usepackage{amsmath} 
\usepackage[T1]{fontenc}

\usepackage{graphicx,verbatim}
\usepackage{placeins}

\usepackage{booktabs}
\usepackage{multirow}
\usepackage{float}
\usepackage{adjustbox}
\usepackage{array}
\usepackage{ragged2e}
\usepackage{amsfonts}
\usepackage{xcolor}
\usepackage{makecell}
\usepackage{subcaption}
\usepackage{caption}
\usepackage{longtable}
\usepackage{pdflscape}
\usepackage{enumitem}
\usepackage[pass]{geometry}
\usepackage[hidelinks]{hyperref}
\hypersetup{pdftitle={Med-AR: Autoregressive Vision-Language Pretraining for Long-Tailed Chest X-Ray Classification and Uncertainty-Aware Evaluation},pdfauthor={Janhavi Prabhu et al.}}

\begin{document}
\title{Med-AR: Autoregressive Vision-Language Pretraining for Long-Tailed Chest X-Ray Classification and Uncertainty-Aware Evaluation}

\author{Janhavi Prabhu \and Sahil \and Akshay V \and Shivam Shukla \and Manoj Tadepalli \and Preetham Putha}
\authorrunning{Prabhu et al.}

\institute{
Qure.ai, Bangalore, India \\
\email{janhavi.prabhu@qure.ai} \\
\email{sahil@qure.ai} \\
\email{Akshay.V@qure.ai} \\
\email{shivam.shukla@qure.ai} \\
\email{manoj.tadepalli@qure.ai} \\
\email{preetham.putha@qure.ai}
}

\titlerunning{Med-AR for Long-Tailed Chest X-Ray Classification}
\maketitle

\begin{abstract}
Long-tailed chest X-ray classification requires visual representations that capture both common abnormalities and subtle, infrequent findings. We propose \textbf{Med-AR-8B and Med-AR-2B}, two radiology-native autoregressive vision-language models pretrained with structured reports, abnormality-focused text, and region annotations. We evaluate the transfer of their visual encoders to multi-label classification against contrastive, self-supervised, and supervised pretrained encoders, including Med-CLIP, CheXFound, EVA-Base, ARK, and BioViL-T, using a common ML-Decoder classification head. To assess fine-grained recognition, we also construct LLM-expanded, report-derived label sets for MIMIC-CXR and CheXpert. Across PadChest, MIMIC-CXR, and CheXpert, Med-AR-8B outperforms Med-CLIP in mean AUROC and AUPRC for head, medium, and tail findings. On MIMIC-CXR, it increases tail-label mean AUPRC from 0.1033 to 0.1441. Med-AR-2B achieves the strongest discrimination results on PadChest. Across the broader encoder comparison, a Med-AR variant achieves the highest mean AUROC and AUPRC in every reported prevalence group on each public dataset. Both Med-AR variants also achieve lower excess area under the risk--coverage curve than Med-CLIP on all three public datasets, indicating improved selective-prediction performance under the evaluated protocol. Internal results are metric-dependent, with Med-CLIP retaining advantages in overall and tail AUPRC and in selective prediction. These findings establish Med-AR as a strong pretraining recipe for long-tailed chest X-ray classification on the evaluated public benchmarks and demonstrate the value of assessing discrimination and selective prediction together.


\end{abstract}
\section{Introduction}

Chest X-rays (CXRs) are among the most frequently performed imaging studies for screening, triage, and longitudinal monitoring of thoracic disease, making robust automated abnormality recognition clinically important. However, multi-label CXR classification remains challenging because thoracic findings exhibit a highly long-tailed distribution: a small subset of common abnormalities accounts for the majority of positive samples, while many clinically significant findings appear only infrequently~\cite{holste2024towards,mimic-database}. Consequently, models may achieve strong aggregate metrics yet fail on rare conditions, where reliable performance is often most clinically necessary. This problem is further complicated by disease co-occurrence, reporting variability, label noise, and dataset heterogeneity.

Rare abnormality recognition in CXRs also differs from standard image-level classification. Many uncommon or clinically subtle findings are spatially localized, low contrast, or interpretable only in relation to surrounding anatomical structures. Although broad abnormalities such as cardiomegaly or large pleural effusions can often be identified from coarse global structure, other findings require fine-grained local evidence together with anatomical and contextual understanding. Effective long-tail CXR classification therefore depends on representations that preserve localized visual detail while simultaneously maintaining clinically meaningful global context, rather than relying solely on pooled image summaries~\cite{Huang2021-sf,Wu2023MedKLIP}.

To address these challenges, recent work has increasingly explored vision-language pretraining using paired radiology reports as supervision. Contrastive approaches, including CLIP, CXR-CLIP, Language over Labels, and MedCLIP, learn transferable visual representations by aligning image and text embeddings~\cite{Alec2021-qc,You2023-te,Wiehe2022-wu,Wang2022-vp}. These methods have demonstrated strong performance in medical imaging~\cite{boecking2022making,Huang2021-sf,Wu2023MedKLIP}. Global image--text alignment compresses report information into pooled representations, potentially limiting the supervision available for fine-grained findings. This limitation does not apply uniformly to contrastive learning: local and knowledge-guided approaches explicitly model finer-grained relationships. Our comparison focuses on the contrastive recipe evaluated here, rather than all possible contrastive formulations.

Autoregressive vision-language pretraining provides an alternative supervision paradigm. Rather than optimizing only global image-text agreement, autoregressive models condition on the image and generate textual outputs sequentially token by token~\cite{Moon2021-wz,Chen2024-ii,Bai2025-jv,Chen2023-fy,Zhu2025-bq,Kimi_Team2025-ez}. This formulation naturally supports richer and more heterogeneous supervision from the same study, including full reports, impression summaries, structured findings, region descriptions, bounding-box annotations, question answering, and step-by-step clinical reasoning. Because supervision is provided through token-level prediction rather than a single pooled similarity objective, rare findings contribute explicit gradients whenever they appear, potentially encouraging representations that better preserve fine-grained anatomical and contextual information. Autoregressive training therefore offers a scalable framework for integrating diverse forms of clinical supervision within a unified objective, without requiring separate task-specific formulations. In radiology, where interpretation depends not only on identifying abnormalities but also on understanding their anatomical location, prominence, and contextual relationships, multimodal next-token prediction may therefore produce more transferable visual representations for downstream long-tail recognition.

Despite growing interest in large autoregressive vision-language models, it remains unclear whether radiology-native autoregressive pretraining provides better downstream transfer than medical contrastive pretraining under controlled experimental conditions. Prior comparisons often differ simultaneously in architecture, pretraining corpus, optimization strategy, and downstream evaluation protocol, making it difficult to isolate the contribution of the pretraining objective itself. Furthermore, evaluations of medical vision-language models have primarily focused on aggregate discrimination performance, with comparatively less attention given to long-tail behavior and uncertainty-aware evaluation, both of which are critical for reliable clinical deployment~\cite{faghani2023quantifying,Zou2023-ig,Hullermeier2021-sf,Geifman2018-bh}.

In this work, we address this gap through a comparison of contrastive and autoregressive pretraining under a matched downstream architecture for long-tailed multi-label CXR classification. We compare MedCLIP-style contrastive pretraining and autoregressive vision-language pretraining using the InternViT architecture trained under each objective, followed by independent downstream fine-tuning with an ML-Decoder classification head, thereby controlling the downstream architecture while comparing complete pretraining recipes. Our autoregressive framework is trained in a radiology-native setting using heterogeneous supervision, including structured multi-section report templates, auxiliary abnormality-focused targets, and approximately 100k region annotations. To evaluate rare-finding transfer more rigorously, we additionally construct LLM-expanded label sets for MIMIC-CXR and CheXpert that increase the granularity of downstream evaluation. Experiments across an internal dataset, MIMIC-CXR, CheXpert, and PadChest demonstrate that autoregressive pretraining achieves stronger downstream discrimination on public benchmarks, improves performance on multiple tail abnormalities, and yields lower EAURC on the public datasets, while the contrastive baseline retains lower EAURC on the internal dataset~\cite{whata2024uncertainty2,Geifman2018-bh}.

\noindent\textbf{Our contributions are threefold:}
\begin{itemize}
    \item We compare contrastive and autoregressive pretraining recipes for long-tailed chest X-ray classification, using separate InternViT encoders independently fine-tuned with the same ML-Decoder architecture.
    
    \item We develop a radiology-native autoregressive pretraining framework that integrates heterogeneous supervision from structured reports, auxiliary text targets, and region-level annotations, enabling the encoder to learn representations beyond global image-report alignment.
    
    \item We provide a comprehensive evaluation across internal and public CXR datasets, including head-tail analysis, LLM-expanded long-tail benchmarks for MIMIC-CXR and CheXpert, and uncertainty-aware assessment using risk-coverage metrics.
\end{itemize}

\section{Related Work}

\subsection{Long-tailed chest X-ray classification}
Long-tailed recognition in chest X-rays has received increasing attention, particularly following the introduction of the CXR-LT benchmark and challenge, which demonstrated that strong aggregate performance can conceal poor recognition of rare findings~\cite{Holste2022-fo,holste2024towards}. Existing approaches primarily address imbalance during supervised downstream training through techniques such as re-weighted losses, asymmetric objectives, resampling, data augmentation, and architectural modifications including multi-view fusion~\cite{10094303,Ben-Baruch2020-pm,Park2023-nt,Kim2023-kp}. Although these methods improve classification under skewed label distributions, they focus largely on optimization at the downstream stage rather than on the representations learned during pretraining.

This limitation is particularly important in chest radiography, where rare abnormalities are often subtle, spatially localized, and difficult to distinguish using coarse global features alone. Consequently, there has been growing interest in leveraging radiology reports as an additional source of supervision, since textual descriptions provide clinically meaningful context that may help models learn more transferable representations for rare and fine-grained findings. This motivation has led to the increasing adoption of vision-language pretraining methods in medical imaging.

\subsection{Contrastive pretraining}
Medical vision-language pretraining commonly uses radiology reports as weak supervision for learning transferable visual representations. Early and widely adopted methods follow CLIP-style contrastive alignment, including CXR-CLIP, Language over Labels, and MedCLIP, which learn a shared image--text embedding space from paired or weakly paired image-report studies~\cite{Alec2021-qc,You2023-te,Wiehe2022-wu,Wang2022-vp}. By incorporating report semantics during pretraining, these approaches improve representation quality beyond purely supervised image classification.

Building on this paradigm, subsequent work has explored richer forms of alignment to better capture the structure and granularity of radiology reports. For example, GLoRIA introduces both global and local alignment objectives between report words and image regions, while MedKLIP incorporates knowledge-enhanced triplet encoding derived from medical reports to guide visual learning from chest X-rays~\cite{Huang2021-sf,boecking2022making,Wu2023MedKLIP}. Similar trends have also emerged outside medical imaging, where methods such as DenseCLIP and FineCLIP improve dense and fine-grained recognition by modeling local visual--textual correspondences rather than relying solely on pooled embeddings~\cite{Rao2022DenseCLIP,Jing2024FineCLIP}. In parallel, recent studies use large language models to extract structured disease information from reports, improving weak supervision and report understanding~\cite{wei2024enhancing}.

Collectively, these works suggest that richer supervision and finer-grained image-text interactions can improve representation learning. Global alignment remains a common component, although local and knowledge-guided contrastive methods also exploit finer-grained supervision. This has motivated growing interest in autoregressive vision-language modeling, where supervision is provided through sequential next-token prediction rather than pooled embedding alignment alone.

\subsection{Autoregressive pretraining}
Autoregressive and generative objectives have emerged as a strong alternative for multimodal representation learning. Early work such as VirTex demonstrated that generating captions from semantically rich annotations can produce transferable visual features, while SimVLM introduced a simplified large-scale pretraining framework based on prefix language modeling~\cite{Desai2021VirTex,Wang2021SimVLM}. Subsequent methods further expanded this paradigm. CoCa combines contrastive alignment with autoregressive caption generation within a unified framework, and Emu extends next-token prediction to interleaved image--text sequences~\cite{Yu2022CoCa,Sun2024Emu}. More recently, AIM and AIMv2 showed that large autoregressively pretrained vision encoders can scale effectively and transfer well across classification, localization, and grounding tasks~\cite{ElNouby2024AIM,Fini2025AIMv2}.

A key advantage of autoregressive supervision is that it naturally supports dense and heterogeneous training signals. Instead of optimizing a single global similarity objective, autoregressive models can learn from full reports, impression summaries, structured findings, region descriptions, and other clinically relevant textual outputs. This property is particularly appealing in radiology, where understanding subtle abnormalities often depends on contextual and localized information. In the medical domain, generative pretraining has therefore been explored for joint image-text understanding and report generation~\cite{Moon2021-wz}, while recent multimodal systems such as Florence-VL and InternVL3 further demonstrate the increasing role of autoregressive vision-language modeling~\cite{Chen2024-ii,Zhu2025-bq}.

Despite this progress, there remains limited evidence on whether autoregressive pretraining provides advantages over medical contrastive pretraining for long-tailed multi-label chest X-ray classification under controlled experimental settings. In particular, prior comparisons often differ simultaneously in architecture, optimization strategy, and downstream evaluation protocol, making it difficult to isolate the effect of the pretraining objective itself.

\subsection{Uncertainty-aware evaluation in medical imaging}
Beyond discrimination performance alone, reliable deployment of medical AI systems also requires robust uncertainty estimation. This is especially important in long-tailed classification settings, where rare findings are underrepresented during training and therefore more susceptible to unreliable predictions. Prior work in machine learning and radiology distinguishes between aleatoric uncertainty, which arises from inherent noise or ambiguity in the data, and epistemic uncertainty, which reflects uncertainty in the model parameters due to limited or biased observations~\cite{Hullermeier2021-sf,faghani2023quantifying}. In chest radiography, both forms of uncertainty are clinically relevant because findings may be visually ambiguous, sparsely represented, or confounded by overlapping pathology.

Recent work has increasingly emphasized that uncertainty should be evaluated jointly with prediction reliability and clinical decision risk rather than treated as a secondary metric~\cite{Zou2023-ig,Sensoy2021-fy}. Beyond calibration alone, selective prediction and risk-coverage analysis assess whether a model's confidence can appropriately rank its own errors, enabling uncertain predictions to be deferred to clinicians when necessary~\cite{Geifman2018-bh,Geifman2019SelectiveNetAD}. However, uncertainty-aware evaluation remains relatively underexplored in studies of medical vision-language pretraining, particularly in the long-tailed multi-label setting where rare diseases are both more difficult to recognize and more prone to overconfident failure modes.

\section{Methodology}
This study investigates how different vision-language pretraining paradigms influence representation learning and downstream transfer in long-tailed chest X-ray classification. The problem is particularly challenging because rare thoracic abnormalities are often subtle, sparsely represented, and frequently co-occur with more common findings~\cite{holste2024towards,mimic-database}. In such settings, robust recognition depends on representations that preserve both localized visual detail and broader clinical context. The contrastive recipe considered here aligns pooled image and text representations, whereas autoregressive pretraining learns by sequentially predicting clinically grounded text conditioned on the image. This autoregressive formulation naturally supports heterogeneous supervision signals, including report generation, structured findings, region descriptions, question answering, and bounding-box annotations within a unified training framework. If such supervision improves representation quality, autoregressive pretraining may provide a scalable approach for incorporating increasingly diverse forms of clinical supervision. To evaluate these differences, we perform a comparison under matched downstream architecture between contrastive and autoregressive pretraining under matched downstream evaluation settings, described in the following sections.
\begin{figure}
\centering
\includegraphics[width=0.99\linewidth]{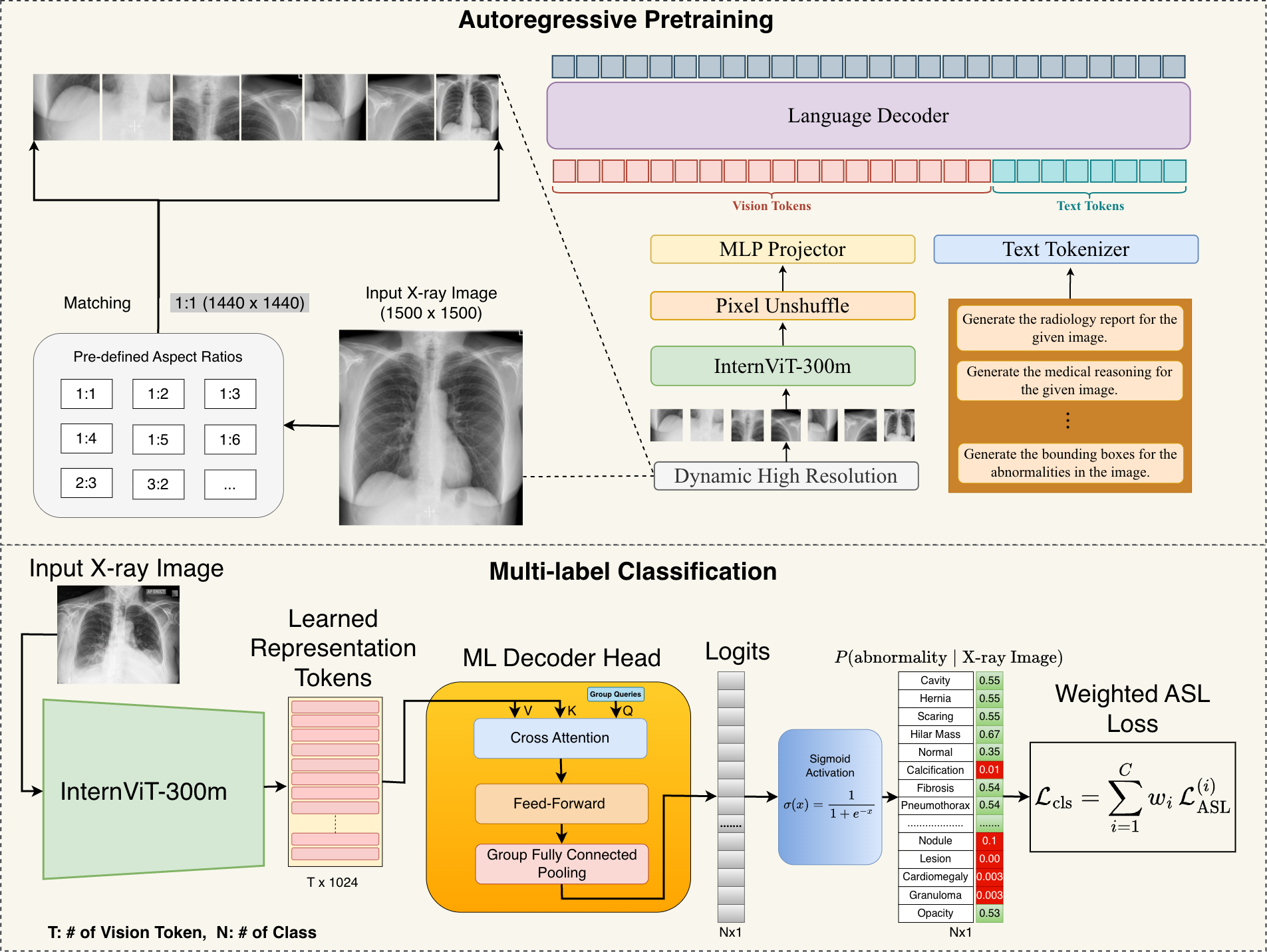}
\caption{Overview of the proposed framework. \textbf{Top:} In autoregressive pretraining, each chest X-ray is dynamically tiled at high resolution according to a matched aspect ratio and encoded by InternViT-300M. The visual features are transformed into vision tokens through pixel unshuffle and an MLP projector, then combined with prompt-conditioned text tokens in a language decoder. The model is trained with heterogeneous radiology-native supervision, including report generation, medical reasoning, structured abnormality outputs, and region-level descriptions such as abnormality bounding boxes. \textbf{Bottom:} For downstream long-tailed multi-label classification, the pretrained InternViT encoder is transferred and coupled with an ML-Decoder head, which predicts logits for all abnormalities and is optimized with weighted asymmetric loss.}
\label{fig:model_architecture}
\end{figure}

\subsection{Overview}
We study the effect of the pretraining paradigm on long-tailed multi-label chest X-ray (CXR) classification by comparing two vision-language pretraining strategies under a controlled downstream setup: (i) MedCLIP-style contrastive pretraining~\cite{Wang2022-vp}, and (ii) autoregressive pretraining using an InternVL3-style vision-language model~\cite{Zhu2025-bq}. To control the downstream architecture, both approaches use independently pretrained instances of the same image-backbone family and are fine-tuned with the same classification head, namely ML-Decoder~\cite{Ridnik2021-yk} shown in Figure \ref{fig:model_architecture}.

Formally, let $x$ denote a chest X-ray image and let $y \in \{0,1\}^C$ denote the multi-label target vector over $C$ abnormalities. In both settings, a pretrained vision encoder $E(\cdot)$ first maps the image to a set of visual features,
\begin{equation}
    \mathbf{z} = E(x),
\end{equation}
which are then passed to independently trained instances of the same multi-label classifier architecture $g(\cdot)$ to obtain per-label probabilities,
\begin{equation}
    \hat{\mathbf{p}} = g(\mathbf{z}) \in [0,1]^C.
\end{equation}
The transferred encoders share an architecture but are pretrained with different objectives and supervision. In particular, the autoregressive recipe additionally uses structured and region-level targets. This design controls the downstream architecture but does not isolate the effect of the pretraining objective from the supervision, initialization, or optimization recipe.

\subsection{Shared Vision Backbone Architecture}

For both paradigms, we use InternViT-300M as the visual backbone. InternViT is a transformer-based image encoder designed for high-resolution visual understanding and defines the common encoder architecture in our comparison. In the autoregressive setup, the encoder operates within the InternVL3 framework~\cite{Zhu2025-bq}; in the contrastive setup, an independent encoder is trained using our MedCLIP-style contrastive recipe~\cite{Wang2022-vp}. By transferring the same encoder family to the downstream task, we reduce architectural confounding while comparing the transfer performance of the two recipes.

\subsection{Autoregressive Vision-Language Pretraining}

Our autoregressive pretraining setup is implemented using InternVL3~\cite{Zhu2025-bq}, as illustrated at the top of Fig.~\ref{fig:model_architecture}, where an InternViT image encoder is trained jointly with a language decoder via conditional next-token prediction. Given an image $x$ and a textual prompt $q$, the model generates a target sequence
\begin{equation}
    t = (t_1, t_2, \dots, t_T),
\end{equation}
which may correspond to a structured report, an impression-style summary, a compact abnormality list, or a region-grounded textual description. The training objective maximizes the conditional likelihood of the target sequence:
\begin{equation}
    \mathcal{L}_{\mathrm{AR}}
    = - \sum_{k=1}^{T} \log p(t_k \mid x, q, t_{<k}).
\end{equation}

This objective couples visual representation learning with language generation. The image is encoded into a sequence of visual tokens that condition the decoder, and the encoder is trained to produce features that support accurate and coherent text generation. This motivates the hypothesis that the learned representations retain structured and fine-grained information; classification results alone do not directly establish visual grounding.

A key advantage of this formulation is that it naturally supports multiple forms of supervision through prompt-conditioned generation. The same image can be used to generate different clinically meaningful outputs, such as detailed findings, concise impressions, abnormality summaries, or region-level descriptions. This provides dense, token-level supervision across multiple views of the data, rather than relying on a single global alignment signal.

To further enrich supervision, we construct diverse training targets from radiology reports and auxiliary annotations. These include: (i) section-wise report templates, (ii) short impression-like summaries, (iii) compact abnormality-focused outputs, and (iv) region-level descriptions derived from bounding-box annotations. Such heterogeneous supervision is intended to associate localized image patterns with corresponding text, potentially benefiting subtle and spatially specific abnormalities.

To study how the strength of the language modeling objective affects the learned visual features, we pretrain two models with identical InternViT encoders but different decoder scales. One model uses a 2B-parameter language decoder (Med-AR-2B), while the other uses an 8B-parameter decoder (Med-AR-8B). A larger decoder can model more complex linguistic patterns, but may also dominate training or overfit to dataset-specific text, potentially limiting the generality of the visual features. Whether decoder capacity changes the information retained by the vision encoder remains an empirical question.

After pretraining, the language decoder is discarded and only the vision encoder is transferred to the downstream classification task. Both Med-AR-2B and Med-AR-8B are evaluated under an identical fine-tuning setup and compared against an InternViT encoder pretrained with MedCLIP, enabling a comparison of the resulting encoders under matched downstream architecture.

\subsection{MedCLIP-Style Contrastive Pretraining}

Our contrastive baseline combines InternViT with BiomedVLP-CXR-BERT and uses the MedCLIP semantic-matching objective~\cite{Wang2022-vp,10204115}. We retain the name Med-CLIP in the result tables for this implementation. Let $v=f_{\mathrm{img}}(x)$ and $u=f_{\mathrm{text}}(r)$ denote image and report embeddings. Medical semantic information from report-derived labels determines soft matching targets for image--text pairs, rather than assigning a positive target only to the paired report and treating every other report as a negative.

The bidirectional semantic-matching objective is
\begin{equation}
\mathcal{L}_{\mathrm{MedCLIP}}=\frac{1}{2}\left(\mathcal{L}_{\mathrm{img}\rightarrow\mathrm{text}}+\mathcal{L}_{\mathrm{text}\rightarrow\mathrm{img}}\right),
\end{equation}
where each directional term is a cross-entropy between the semantic matching targets and the similarity-based predicted matching distribution. This is distinct from standard paired InfoNCE: images and reports with related medical labels can receive nonzero matching targets even when they are not an original pair. After pretraining, the visual encoder is transferred to downstream classification.

The baseline aligns pooled visual and textual representations with semantic supervision. The autoregressive recipe additionally uses structured generation targets and region annotations. The comparison therefore evaluates complete pretraining recipes rather than isolating the objective or establishing a limitation of all contrastive methods.

\subsection{Downstream Fine-Tuning with ML-Decoder}

After pretraining, all three vision encoders are fine-tuned on downstream multi-label classification using ML-Decoder~\cite{Ridnik2021-yk} as shown in bottom of Fig \ref{fig:model_architecture}. ML-Decoder is a transformer-based prediction head designed for efficient large-scale multi-label recognition. Instead of learning an independent classifier for each label, it uses a fixed set of learnable queries that interact with the encoded image features to produce label predictions.

Let $\mathbf{z} \in \mathbb{R}^{N \times d}$ denote the output token sequence of the vision encoder, where $N$ is the number of visual tokens and $d$ is the feature dimension. ML-Decoder maps $\mathbf{z}$ to class logits
\begin{equation}
    \mathbf{s} = h(\mathbf{z}) \in \mathbb{R}^{C},
\end{equation}
followed by sigmoid activation to obtain class probabilities
\begin{equation}
    \hat{p}_i = \sigma(s_i), \qquad i = 1,\dots,C.
\end{equation}

A key advantage of ML-Decoder is that it scales efficiently to large label spaces while still modeling label dependencies and co-occurrence structure. This makes it particularly suitable for long-tailed CXR classification, where abnormalities often co-occur and the number of classes can be large in fine-grained evaluation settings.

\subsection{Loss Function for Long-Tailed Multi-Label Classification}

To address both inter-class and intra-class imbalance during downstream training, we combine class-specific weighting with asymmetric loss (ASL)~\cite{Ben-Baruch2020-pm,Kim2023-kp}.

\paragraph{Class-specific weighting.}
Let $y_i \in \{0,1\}$ denote the ground-truth label for class $i$, and let $\rho_i$ denote the positive sample ratio of class $i$ in the training set. We define a class-aware sample weight as
\begin{equation}
    w_i = y_i \exp(1-\rho_i) + (1-y_i)\exp(\rho_i).
\end{equation}
This weighting increases the contribution of rare positive classes while preserving the contribution of negative examples.

\paragraph{Asymmetric loss.}
For each class $i$, the asymmetric loss is defined as
\begin{equation}
\mathcal{L}_{\mathrm{ASL}}^{(i)}
=
- y_i (1-\hat{p}_i)^{\gamma_+} \log(\hat{p}_i)
- (1-y_i)\, \hat{p}_{m,i}^{\gamma_-} \log(1-\hat{p}_{m,i}),
\end{equation}
where
\begin{equation}
    \hat{p}_{m,i} = \max(\hat{p}_i - m, 0).
\end{equation}
Here, $\gamma_+$ and $\gamma_-$ are focusing parameters for positive and negative samples, respectively, and $m$ is a probability margin used to suppress easy negatives. Following prior long-tailed CXR classification work~\cite{Kim2023-kp}, we use $\gamma_+=1$, $\gamma_-=4$, and $m=0.05$.

\paragraph{Final training objective.}
The final downstream loss is computed as the weighted sum over classes:
\begin{equation}
    \mathcal{L}_{\mathrm{cls}}
    =
    \sum_{i=1}^{C} w_i \, \mathcal{L}_{\mathrm{ASL}}^{(i)}
\end{equation}
This objective jointly addresses inter-class imbalance, by up-weighting rare classes, and intra-class imbalance, by down-weighting easy negatives. As a result, it is well suited for long-tailed multi-label CXR classification.

\subsection{Long-tailed evaluation protocol}
\label{sec:headmidtail}

We study the downstream task as multi-label abnormality classification on four datasets: an internal clinical dataset, MIMIC-CXR, CheXpert and PadChest. This multi-dataset setup is important because long-tailed behavior can be sensitive to dataset composition, reporting style, disease prevalence, and annotation granularity. Evaluating on both internal and public datasets therefore helps distinguish in-domain gains from broader transfer behavior.

To make the long-tail analysis explicit, we partition labels separately within each dataset according to their frequency in the \emph{training split}. Let $c_i$ denote the number of positive training examples for label $i$. Because the label-frequency distribution is strongly long-tailed, we first transform counts into log space:
\begin{equation}
x_i = \log(1 + c_i).
\end{equation}
We then compute dataset-specific thresholds from the empirical 33rd and 67th percentiles of the log-count distribution:
\begin{equation}
T_1 = Q_{0.33}(x), \qquad T_2 = Q_{0.67}(x).
\end{equation}
To obtain interpretable thresholds in the original count space, we map these quantiles back using the inverse transform:
\begin{equation}
C_1 = \exp(T_1) - 1, \qquad C_2 = \exp(T_2) - 1.
\end{equation}
Finding labels are then assigned to three groups directly according to their positive training counts:
\begin{equation}
\text{tail}: c_i < C_1, \qquad
\text{medium}: C_1 \leq c_i < C_2, \qquad
\text{head}: c_i \geq C_2.
\end{equation}

This procedure provides a dataset-adaptive partition of the label space while applying the same rule across all benchmarks. Because the log transform preserves ordering, the groups principally reflect relative label-frequency ranks within each dataset, with boundary details determined by quantile interpolation and ties. Tail membership therefore does not imply the same absolute prevalence across datasets. The resulting head/medium/tail split enables finer-grained analysis of whether improvements from pretraining are concentrated on frequent labels or extend to intermediate- and low-frequency abnormalities. The dataset-specific count thresholds used in this work are summarized in Table~\ref{tab:head_medium_tail_thresholds}.

For MIMIC-CXR and CheXpert, we further expand the evaluation space through an LLM-based label extraction pipeline, producing a finer-grained long-tailed benchmark than the standard label set. This expanded label space, shown in Figure~\ref{fig:mimic_distri} and Figure~\ref{fig:chex_distri}, is not introduced as a separate task; rather, it serves as a more sensitive probe of representation quality under rare-label transfer, where differences between pretraining paradigms may be difficult to detect using only a small number of coarse disease categories.

\section{Experiments}
\FloatBarrier

\subsection{Dataset Preparation}
\subsubsection{Pre-training:}
For pre-training we use an internal training dataset of 1.4 million chest X-rays and their corresponding reports, obtained from various hospitals, anonymized and cleaned. For MedCLIP pretraining, we leverage paired image--text data and use prompt engineering to design an effective instruction for extracting detailed disease labels from radiology reports, enabling fine-grained annotation of more than 150 findings, including rare abnormalities. The LLM (Qwen3~\cite{Qwen3TechnicalReport}) was prompted to produce a JSON object listing all present findings, guided by a curated list of over 150 finding categories. Downstream label expansion is described separately in Section~\ref{subsec:tag_extract}. For autoregressive pre-training, we create a structured JSON template consisting of 15 subheadings covering the various anatomical regions to look at in a chest radiograph. Under each subheading, we include further categorizations for abnormalities. In addition to the 15 headings, we also add \textit{Findings}, \textit{Impression}, \textit{Summary}, \textit{Reasoning} and \textit{Prominence Score} sections. The prominence score section provides a dictionary of the report-derived findings along with a generated prominence score, from 0--5. We use Qwen3 to extract this information from the reports and output the response in the structured template format. In cases when an abnormality is not mentioned or is not present, we instruct the LLM to use a negation statement. This procedure conflates non-mention with explicit absence and is a source of weak-supervision noise. The reasoning and prominence fields are generated from reports; they are not independently adjudicated visual assessments. Furthermore, we incorporate expert-annotated bounding boxes for approximately 20 thoracic findings (e.g., opacity, nodule, pneumothorax, mass), represented as four-point coordinates. For each supervision type: global report parsing, structured abnormality extraction, prominence scoring, and region-level bounding-box, we design dedicated prompt templates for Qwen3, ensuring consistent JSON outputs across these heterogeneous data sources which are then used for pretraining InternVL3 model.
\subsubsection{Finetuning:}
We conduct separate training experiments on four large-scale CXR datasets: an internal collection of chest radiographs (Internal), PadChest \cite{Bustos2019-wt}, CheXpert \cite{irvin2019chexpertlargechestradiograph} and MIMIC-CXR \cite{PhysioNet-mimic-cxr-2.1.0}; the splits in Table~\ref{Train_Test} contain 1,872,398 training images and 297,608 test images in total. For each dataset, models are trained exclusively on its corresponding training split and evaluated on its own held-out test set, as summarized in Table~\ref{Train_Test}. 
\begin{itemize}
    \item \textbf{Internal Dataset:} The training split includes approximately 1.4M chest X-rays from multiple hospitals. An LLM fine-tuned on radiologist annotations extracts 40 report-derived labels, including hilar mass. 
    \item \textbf{PadChest:} Includes 160K CXRs labeled with 174 UMLS-derived findings. We use the 29 most frequent labels for training and the held-out split reported in Table~\ref{Train_Test}. Volume loss is the least prevalent retained finding. Tail analysis is restricted to these retained labels, rather than the full PadChest vocabulary.
    \item \textbf{MIMIC-CXR:} A public dataset with 173K CXRs (frontal) and reports. Our LLM extracts $\sim$150 tags for granularity. We use 110 finding labels (with training prevalence > 0.0002) for training. Bullae is the least common retained finding. We plan to release the derived labels and split identifiers subject to the source dataset terms.  
    \item \textbf{CheXpert:} A public dataset with approximately 160K retained CXRs (frontal) and reports. Our LLM extracts $\sim$150 tags for granularity. We use 115 finding labels (with training prevalence > 0.0002) for training. Left ventricular assist device (LVAD) is the least common retained label. We plan to release the derived labels and split identifiers subject to the source dataset terms.
\end{itemize}
\begin{table}
\centering
\begin{tabular}{|c|c|c|}
\hline
\textbf{Dataset} & \textbf{Training Samples} & \textbf{Testing Samples} \\
\hline
Internal & 1,432,373 & 242,965 \\
MIMIC-CXR & 152,240 & 21,466 \\
PadChest & 138,235 & 22,626 \\
CheXpert & 149,550 & 10,551 \\
\hline
\end{tabular}
\caption{Dataset statistics, showing the number of training and testing samples for each dataset used in fine-tuning.}
\label{Train_Test}
\end{table}
\begin{figure}[ht]
    \centering
    \includegraphics[width=\linewidth,height=0.80\textheight,keepaspectratio]{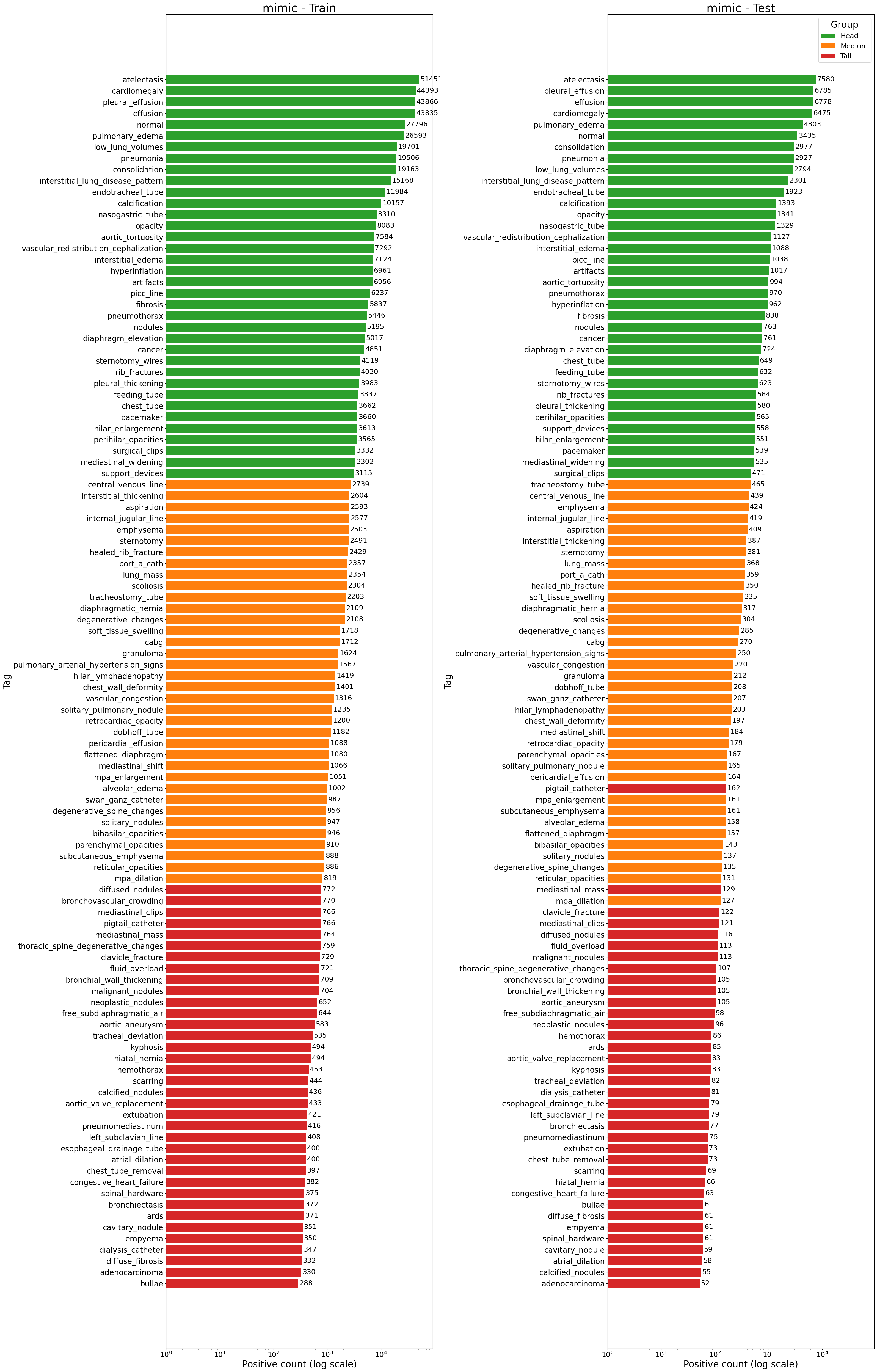}
    \caption{Distribution of extracted MIMIC-CXR tags. The classification label space contains 110 labels; available per-label results are listed in Supplementary Section~\ref{supp:mimic}. The data is split into head, medium, and tail tags. Train prevalence $\geq 0.02\%$; log-count thresholds: tail $< 1106$, medium $< 4210$, head $\geq 4210$.}
    \label{fig:mimic_distri}
\end{figure}

\begin{figure}[ht]
    \centering
    \includegraphics[width=\linewidth,height=0.80\textheight,keepaspectratio]{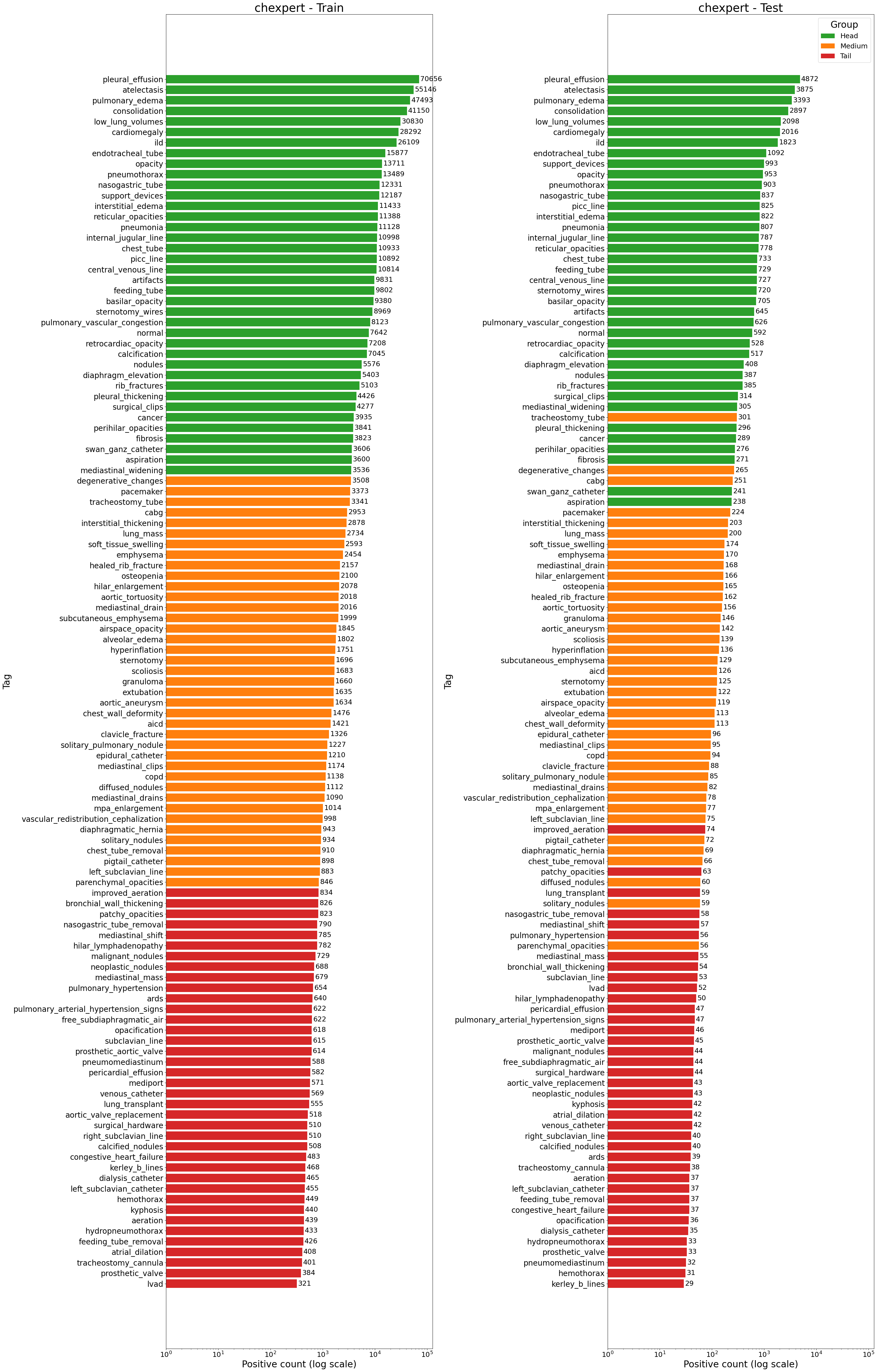}
    \caption{Distribution of the CheXpert tag extraction dataset with 115 tags. The data is split into head, medium, and tail tags. Train prevalence $\geq 0.02\%$; log-count thresholds: tail $< 841$, medium $< 3519$, head $\geq 3519$.}
    \label{fig:chex_distri}
\end{figure}

\begin{figure}[ht]
    \centering

    \begin{subfigure}{\linewidth}
        \centering
        \includegraphics[width=\linewidth,height=0.32\textheight,keepaspectratio]{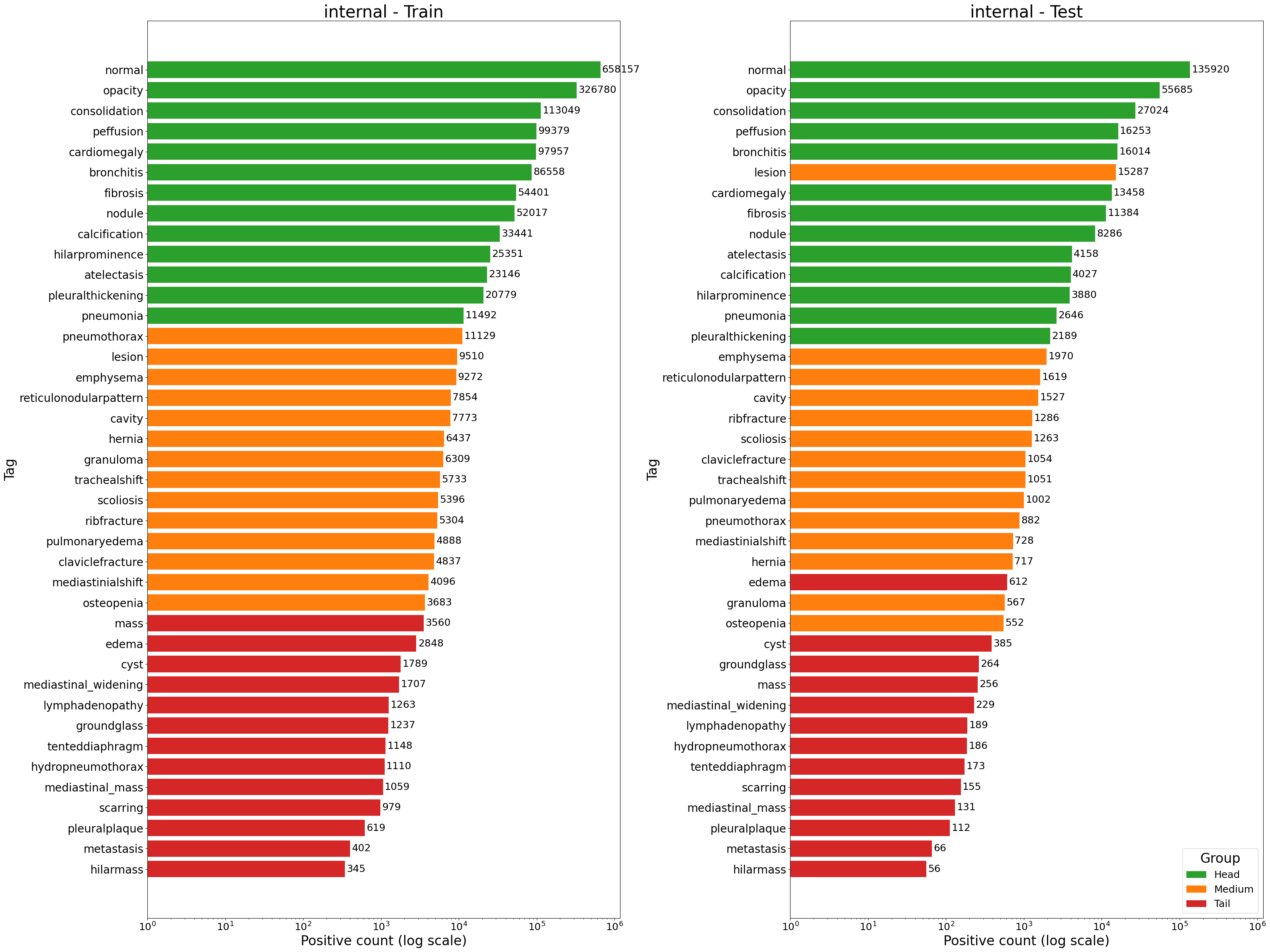}
        \caption{Distribution of the Internal dataset with 40 tags. The data is split into head, medium, and tail tags. Train prevalence $\geq 0.02\%$; log-count thresholds: tail $< 3667$, medium $< 11176$, head $\geq 11176$.}
        \label{fig:internal_distri}
    \end{subfigure}

    \vspace{0.8em}

    \begin{subfigure}{\linewidth}
        \centering
        \includegraphics[width=\linewidth,height=0.32\textheight,keepaspectratio]{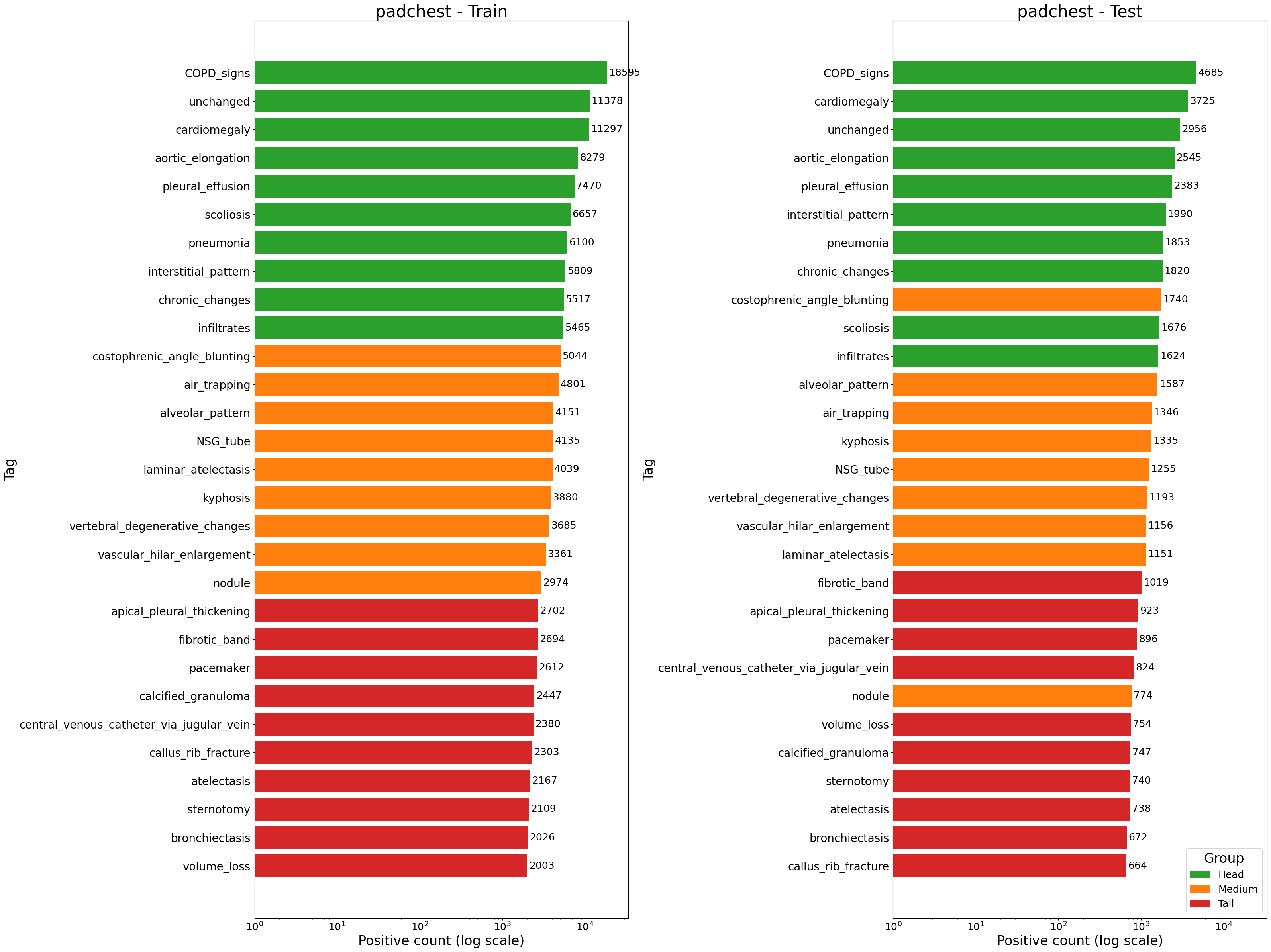}
        \caption{Distribution of the PadChest dataset with 29 tags. The data is split into head, medium, and tail tags. Train prevalence $\geq 0.02\%$; log-count thresholds: tail $< 2765$, medium $< 5361$, head $\geq 5361$.}
        \label{fig:padchest_distri}
    \end{subfigure}

    \caption{Tag distribution comparison for the Internal and PadChest datasets.}
    \label{fig:dataset_distri_combined}
\end{figure}

\FloatBarrier
\vspace{-1em}
The distribution of each tag in the train and test set is given in Figures \ref{fig:mimic_distri} \ref{fig:chex_distri} \ref{fig:internal_distri} and \ref{fig:padchest_distri} and training set thresholds for grouping into head, medium and tail for each dataset is mentioned in Table \ref{tab:head_medium_tail_thresholds} and discussed in Sec \ref{sec:headmidtail}.

\begin{table}
\centering
\caption{Dataset-specific thresholds used to partition labels into head, medium, and tail groups based on the number of positive training examples. Thresholds are derived from the long-tailed label distribution of each dataset after filtering labels with training prevalence $\geq 0.02\%$ as discussed in Section \ref{sec:headmidtail}. }
\label{tab:head_medium_tail_thresholds}
\setlength{\tabcolsep}{8pt}
\renewcommand{\arraystretch}{1.2}
\begin{tabular}{lccc}
\toprule
\textbf{Dataset} & \textbf{Tail} & \textbf{Medium} & \textbf{Head} \\
\midrule
Internal
& $c_i < 3667$
& $3667 \leq c_i < 11176$
& $c_i \geq 11176$ \\

MIMIC-CXR
& $c_i < 1106$
& $1106 \leq c_i < 4210$
& $c_i \geq 4210$ \\

PadChest
& $c_i < 2765$
& $2765 \leq c_i < 5361$
& $c_i \geq 5361$ \\

CheXpert
& $c_i < 841$
& $841 \leq c_i < 3519$
& $c_i \geq 3519$ \\

\bottomrule
\end{tabular}
\end{table}

\subsection{LLM-based Label Extraction and Concordance Analysis}
\label{subsec:tag_extract}

To expand the label space of MIMIC-CXR and CheXpert, we derived structured labels from free-text radiology reports using a large language model (LLM). The prompt specified a comprehensive set of thoracic abnormalities and required the model to return their presence or absence in a structured JSON format. Positive findings were restricted to those supported by the report, with explicit instructions to avoid inference beyond the documented text. We additionally included an \texttt{other\_findings} field to capture abnormalities outside the predefined vocabulary. While this improved coverage, it introduced lexical variability across models, including alternative terminology and singular/plural forms; these free-form outputs were therefore mapped to a canonical label vocabulary during post-processing.

\begin{figure}
    \centering
    \includegraphics[width=\linewidth]{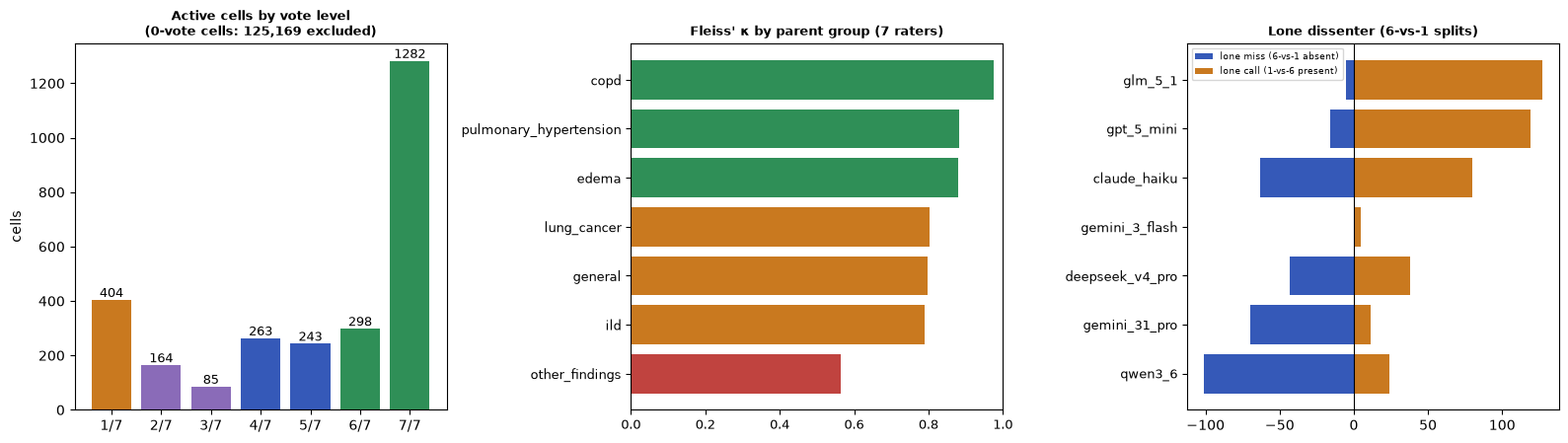}
    \caption{Inter-model concordance of LLM-derived labels. \textbf{Left:} Distribution of votes across seven models for active report--tag pairs, where the vote denotes the number of models (1--7) identifying a finding as present; unanimous-negative pairs (0/7) are excluded. \textbf{Middle:} Fleiss' $\kappa$ by parent finding group, showing strong agreement across clinical categories. \textbf{Right:} Model-specific 6-vs-1 disagreements, distinguishing lone misses from lone positive calls.
}
    \label{fig:concordance_fleiss}
\end{figure}

We evaluated inter-model extraction concordance on a 1,000-report subset using seven LLMs: DeepSeek V4 Pro, Gemini 3.1 Pro, Gemini 3 Flash, GLM 5.1, Qwen3, Claude Haiku, and GPT-5 Mini. Because the resulting multi-label matrix was highly sparse, with approximately 98\% of report--label pairs corresponding to shared negatives, raw percentage agreement would substantially overestimate concordance. We therefore evaluated chance-corrected agreement using Cohen's $\kappa$ for pairwise model comparisons and Fleiss' $\kappa$ for agreement across the complete seven-model panel. Mean pairwise Cohen's $\kappa$ was approximately 0.83 at the individual-label level and increased to 0.87 after related labels were collapsed into parent finding groups, indicating that a substantial component of apparent disagreement resulted from differences in label granularity. Consistently, parent-group Fleiss' $\kappa$ showed strong seven-model concordance across the clinical finding groups ($\kappa=0.79$--$0.98$), whereas \texttt{other\_findings} showed lower agreement ($\kappa=0.56$), reflecting its more heterogeneous and open-ended vocabulary (Figure \ref{fig:concordance_fleiss}).

Residual disagreement was dominated by \emph{synonym routing}, whereby models identified the same underlying finding but assigned different keys. Common examples included \texttt{effusion} versus \texttt{pleural\_effusion} and the nodule family (\texttt{nodules}, \texttt{solitary\_nodules}, and \texttt{solitary\_pulmonary\_nodule}). When \texttt{solitary\_pulmonary\_nodule} was missed under its specific key, \texttt{nodules} was assigned in 96\% of such cases; similarly, \texttt{pleural\_effusion} was assigned in 95\% of cases in which \texttt{effusion} was missed. The open-ended \texttt{other\_findings} field further contributed to this naming variability, reinforcing the need for canonicalization. One notable exception was \texttt{artifacts}, which exhibited model-specific over-extraction rather than synonym substitution and consequently behaved as a source of annotation noise.

For large-scale downstream label expansion, we selected Gemini 3 Flash for its inclusive extraction behavior relative to the other models. This stage is separate from Qwen3-based construction of the internal pretraining targets. General and specific labels may both be emitted and are examined during canonicalization. Without expert reference labels, this comparison does not establish clinical precision or recall. Inter-model agreement measures reproducibility of report extraction, and shared model errors can remain undetected. Parent-group agreement may also conceal errors in fine-grained distinctions. The expanded labels should therefore be interpreted as report-derived weak supervision rather than independently validated image-level ground truth. Supplementary Section~\ref{sec:llm_label_expansion} details parsing, normalization, and concordance analysis.

\subsection{Implementation}

\textbf{Pretraining} MedCLIP is pretrained using AdamW (lr=2e-5) with a one-cycle scheduler. The text encoder is BiomedVLP-CXR-BERT, and the image encoder is InternViT (embedding size=512).
\\
For training the VLM, we use a learning rate of 1e-5 with AdamW and a warmup ratio of 0.05. \\
\textbf{Fine-Tuning} ML-Decoder with k=N (no. of classes), where k is size of group queries serves as the classification head, processing the final hidden state of the image encoder (embedding size = 1024). The model is fine-tuned using Asymmetric Loss, AdamW, and a OneCycle scheduler (batch size = 2304, max lr = 5e-05). We used 8XH100 GPUs for both MedCLIP and Med-AR training. 

\subsection{Evaluation Metrics and Statistical Analysis}

\paragraph{Evaluation metrics.}
We evaluate downstream performance using AUROC, AUPRC, sensitivity, specificity, and Excess Area Under the Risk--Coverage Curve (EAURC). Since the task is multi-label classification, each abnormality is treated as a one-vs-rest binary prediction problem, and metrics are computed per label before being averaged over all labels as well as over the head, medium, and tail subsets.

We report AUROC as a threshold-independent measure of ranking quality, and AUPRC as a complementary metric that is especially informative under class imbalance, where positive cases may be rare~\cite{Saito2015PRC}. To characterize thresholded operating performance, we also report sensitivity and specificity, which quantify the ability to recover positive cases while avoiding false alarms. For public-dataset operating-point results, thresholds are selected per label by maximizing Youden's $J$ on the validation set, as described in Supplementary Section~\ref{supp:operating_points}.

To assess uncertainty-aware reliability, we use EAURC, a selective-prediction metric that evaluates how well model confidence ranks predictions by difficulty ~\cite{Geifman2018-bh}. Lower EAURC indicates a lower excess selective risk under the evaluated confidence ordering. This metric does not measure probability calibration or establish clinical deployment readiness; in sparse multi-label tasks, ordinary error risk can be dominated by negative examples.

\paragraph{Statistical significance tests.}
We use two complementary statistical tests to assess whether differences between pretraining paradigms are systematic. First, we apply the Wilcoxon signed-rank test to paired per-label AUROC and AUPRC values~\cite{Wilcoxon1945}. This test is appropriate because the comparison is paired at the label level, and the distribution of per-label metric differences is not guaranteed to be Gaussian. It therefore provides a robust non-parametric assessment of whether one pretraining strategy tends to outperform the other across all labels, or within the head and tail subsets.

Second, we use DeLong's test to compare label-wise AUROCs between models~\cite{DeLong1988}. DeLong's test is designed for correlated ROC curves and is well suited here because the competing models are evaluated on the same test cases for each label. Unlike the Wilcoxon test, which summarizes paired differences across labels, DeLong's test identifies whether the AUROC gap for a specific abnormality is statistically significant.

The reported significance comparisons are interpreted as nominal, exploratory evidence: correlated labels, multiple comparisons, repeated studies, and training variability limit inference. Non-significance does not establish equivalence, and smaller $p$-values do not measure larger effect sizes. Together, these tests provide complementary evidence: the Wilcoxon signed-rank test assesses whether one pretraining paradigm is consistently better across labels, while DeLong's test identifies which individual abnormalities show significant AUROC differences.

\section{Results and Discussion}
\label{sec:res}

\subsection{Main quantitative comparison}
\label{subsec:mainres}

Table~\ref{tab:groupwise_overall_head_medium_tail} summarizes mean AUROC and mean AUPRC across overall, head, medium, and tail label groups for the four evaluation benchmarks. A clear pattern emerges on the three public datasets, although the preferred autoregressive variant differs across them.

On PadChest, autoregressive pretraining consistently outperforms Med-CLIP across all prevalence groups, with Med-AR-2B achieving the best results throughout. Specifically, mean AUROC improves from 0.8657 to 0.8800 overall, from 0.8425 to 0.8571 for head labels, from 0.8531 to 0.8674 for medium labels, and from 0.9002 to 0.9143 for tail labels. The corresponding AUPRC values also increase consistently, from 0.4310 to 0.4595 overall, from 0.4428 to 0.4706 for head labels, from 0.3678 to 0.3941 for medium labels, and from 0.4762 to 0.5072 for tail labels.

A similar but even stronger trend is observed on MIMIC-CXR, where Med-AR-8B is the best-performing model across all groups. Overall mean AUROC increases from 0.8777 for Med-CLIP to 0.8971 for Med-AR-8B, while overall mean AUPRC rises from 0.2315 to 0.2769. The gains are visible across all prevalence regimes: for head labels, AUROC improves from 0.8756 to 0.8918 and AUPRC from 0.4010 to 0.4432; for medium labels, AUROC improves from 0.8841 to 0.9028 and AUPRC from 0.1902 to 0.2435; and for tail labels, AUROC improves from 0.8733 to 0.8967 and AUPRC from 0.1033 to 0.1441.

CheXpert provides a complementary test of this trend and highlights a stronger role for decoder scale. Med-AR-8B achieves the best AUROC and AUPRC in every prevalence group, lifting overall AUROC from 0.8585 (Med-CLIP) to 0.8759 and overall AUPRC from 0.2142 to 0.2391, with consistent gains for head labels (0.8433 to 0.8565 AUROC; 0.3710 to 0.3978 AUPRC), medium labels (0.8790 to 0.8920 AUROC; 0.1802 to 0.2054 AUPRC), and tail labels (0.8526 to 0.8788 AUROC; 0.0922 to 0.1151 AUPRC). In contrast, Med-AR-2B does not surpass Med-CLIP on this benchmark and is lowest across every prevalence group for both AUROC and AUPRC. Autoregressive supervision therefore remains beneficial on CheXpert, but the advantage is concentrated in the larger decoder, and a smaller autoregressive model is insufficient to outperform a strong contrastive baseline on this dataset.

As shown in Section~\ref{subsec:stat}, paired per-label Wilcoxon results support most of the Med-AR-8B improvements on the public datasets at the nominal 0.05 level. The exception is PadChest medium-label AUPRC ($p=0.0645$). These tests describe the observed label-wise differences and do not isolate the contribution of the pretraining objective.

The internal benchmark presents a more nuanced picture and is best described as broadly comparable between the two paradigms, with a modest edge for autoregressive pretraining in AUROC but a mixed pattern in AUPRC. In terms of AUROC, Med-AR-2B performs best across all groups, improving from 0.9081 to 0.9135 overall, from 0.8947 to 0.9008 for head labels, from 0.9069 to 0.9124 for medium labels, and from 0.9227 to 0.9274 for tail labels. However, the AUPRC results are less uniform. Med-AR-2B achieves the best performance for the head and medium groups, increasing AUPRC from 0.4289 to 0.4415 and from 0.2434 to 0.2452, respectively, whereas Med-CLIP remains strongest overall and on the tail subset, with AUPRC values of 0.2862 overall and 0.1897 on tail labels. Consistent with this mixed aggregate pattern, the statistical analysis in subsection~\ref{subsec:stat} shows that several internal comparisons do not reach significance, especially for AR-8B and for the tail subset, supporting the interpretation that Med-AR and Med-CLIP are broadly comparable on the internal benchmark rather than cleanly separated.

The preferred autoregressive variant is dataset-dependent. Med-AR-2B achieves the highest AUROC on Internal and the best discrimination on PadChest, whereas Med-AR-8B is strongest on MIMIC-CXR and CheXpert. These results indicate that decoder scale interacts with the training recipe and downstream dataset; they do not establish the objective itself as the primary source of improvement.

\begin{table*}[t]
\centering
\scriptsize
\setlength{\tabcolsep}{3pt}
\renewcommand{\arraystretch}{1.12}
\caption{Mean AUROC and AUPRC for overall, head, medium, and tail tags across datasets. The best value within each metric block is shown in bold, and the second-best value is underlined.}
\label{tab:groupwise_overall_head_medium_tail}
\resizebox{\linewidth}{!}{%
\begin{tabular}{llcccccc}
\toprule
\multirow{2}{*}{\textbf{Dataset}} 
& \multirow{2}{*}{\textbf{Group}}
& \multicolumn{3}{c}{\textbf{mAUROC}}
& \multicolumn{3}{c}{\textbf{mAUPRC}} \\
\cmidrule(lr){3-5} \cmidrule(lr){6-8}
& 
& \makecell[c]{\textbf{Med-CLIP}}
& \makecell[c]{\textbf{Med-AR-8B}\\[-0.55ex]\smash{\scriptsize(ours)}}
& \makecell[c]{\textbf{Med-AR-2B}\\[-0.55ex]\smash{\scriptsize(ours)}}
& \makecell[c]{\textbf{Med-CLIP}}
& \makecell[c]{\textbf{Med-AR-8B}\\[-0.55ex]\smash{\scriptsize(ours)}}
& \makecell[c]{\textbf{Med-AR-2B}\\[-0.55ex]\smash{\scriptsize(ours)}} \\
\midrule
\multirow{4}{*}{Internal}
& Overall & 0.9081 & \underline{0.9111} & \textbf{0.9135} & \textbf{0.2862} & 0.2754 & \underline{0.2842} \\
& Head    & 0.8947 & \underline{0.8983} & \textbf{0.9008} & 0.4289 & \underline{0.4367} & \textbf{0.4415} \\
& Medium  & 0.9069 & \underline{0.9100} & \textbf{0.9124} & \underline{0.2434} & 0.2292 & \textbf{0.2452} \\
& Tail    & 0.9227 & \underline{0.9253} & \textbf{0.9274} & \textbf{0.1897} & 0.1639 & \underline{0.1689} \\
\midrule
\multirow{4}{*}{PadChest}
& Overall & 0.8657 & \underline{0.8722} & \textbf{0.8800} & 0.4310 & \underline{0.4469} & \textbf{0.4595} \\
& Head    & 0.8425 & \underline{0.8497} & \textbf{0.8571} & 0.4428 & \underline{0.4583} & \textbf{0.4706} \\
& Medium  & 0.8531 & \underline{0.8595} & \textbf{0.8674} & 0.3678 & \underline{0.3818} & \textbf{0.3941} \\
& Tail    & 0.9002 & \underline{0.9060} & \textbf{0.9143} & 0.4762 & \underline{0.4940} & \textbf{0.5072} \\
\midrule
\multirow{4}{*}{MIMIC}
& Overall & 0.8777 & \textbf{0.8971} & \underline{0.8859} & 0.2315 & \textbf{0.2769} & \underline{0.2541} \\
& Head    & 0.8756 & \textbf{0.8918} & \underline{0.8846} & 0.4010 & \textbf{0.4432} & \underline{0.4264} \\
& Medium  & 0.8841 & \textbf{0.9028} & \underline{0.8881} & 0.1902 & \textbf{0.2435} & \underline{0.2175} \\
& Tail    & 0.8733 & \textbf{0.8967} & \underline{0.8851} & 0.1033 & \textbf{0.1441} & \underline{0.1185} \\
\midrule
\multirow{4}{*}{CheXpert}
& Overall & \underline{0.8585} & \textbf{0.8759} & 0.8422 & \underline{0.2142} & \textbf{0.2391} & 0.2039 \\
& Head    & \underline{0.8433} & \textbf{0.8565} & 0.8310 & \underline{0.3710} & \textbf{0.3978} & 0.3614 \\
& Medium  & \underline{0.8790} & \textbf{0.8920} & 0.8586 & \underline{0.1802} & \textbf{0.2054} & 0.1682 \\
& Tail    & \underline{0.8526} & \textbf{0.8788} & 0.8365 & \underline{0.0922} & \textbf{0.1151} & 0.0829 \\
\bottomrule
\end{tabular}%
}
\end{table*}

\begin{figure*}[t]
\centering

\begin{subfigure}{0.32\textwidth}
    \centering
    \includegraphics[width=\linewidth,height=0.15\textheight,keepaspectratio]{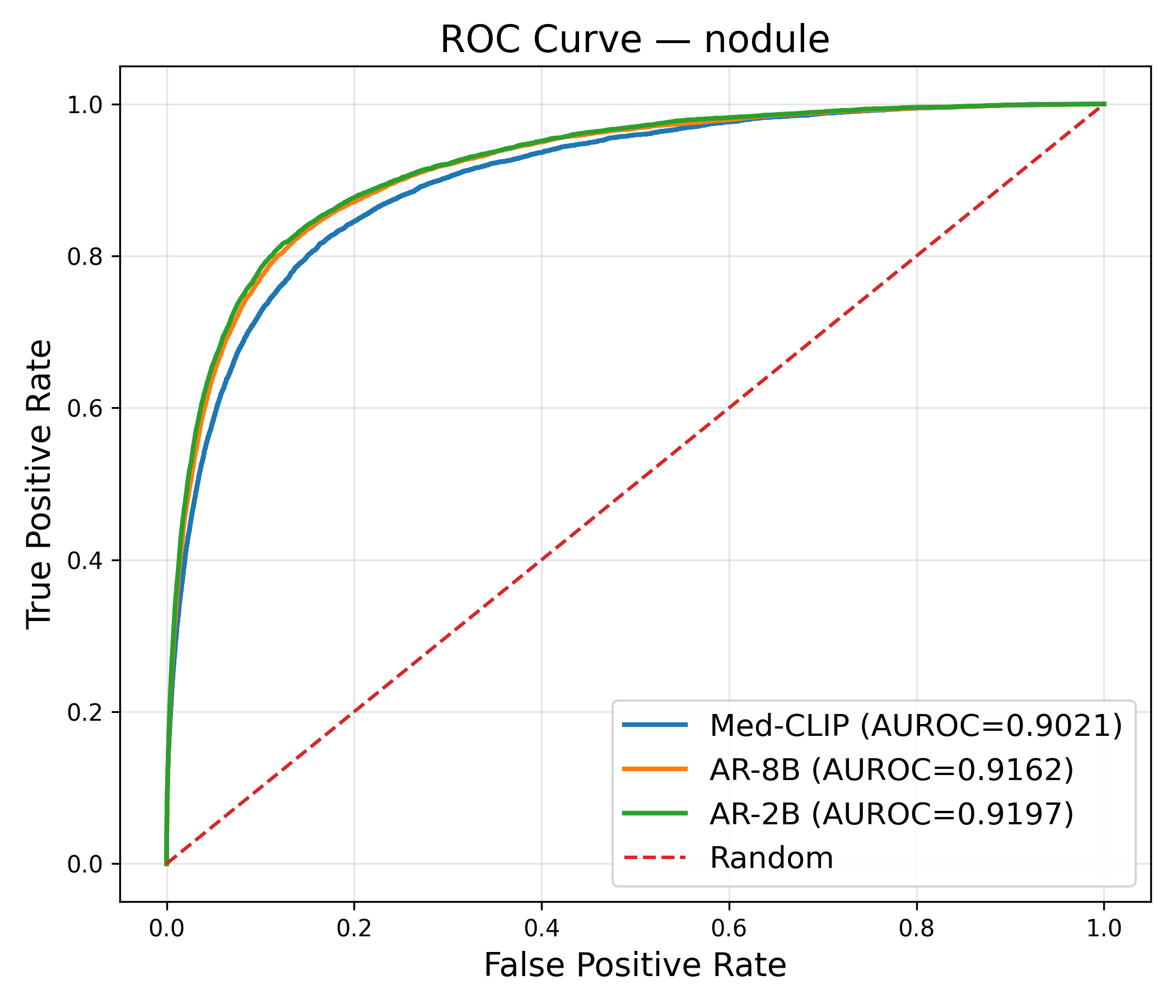}
    \caption{Internal -- Head (\textit{nodule})}
\end{subfigure}
\hfill
\begin{subfigure}{0.32\textwidth}
    \centering
    \includegraphics[width=\linewidth,height=0.15\textheight,keepaspectratio]{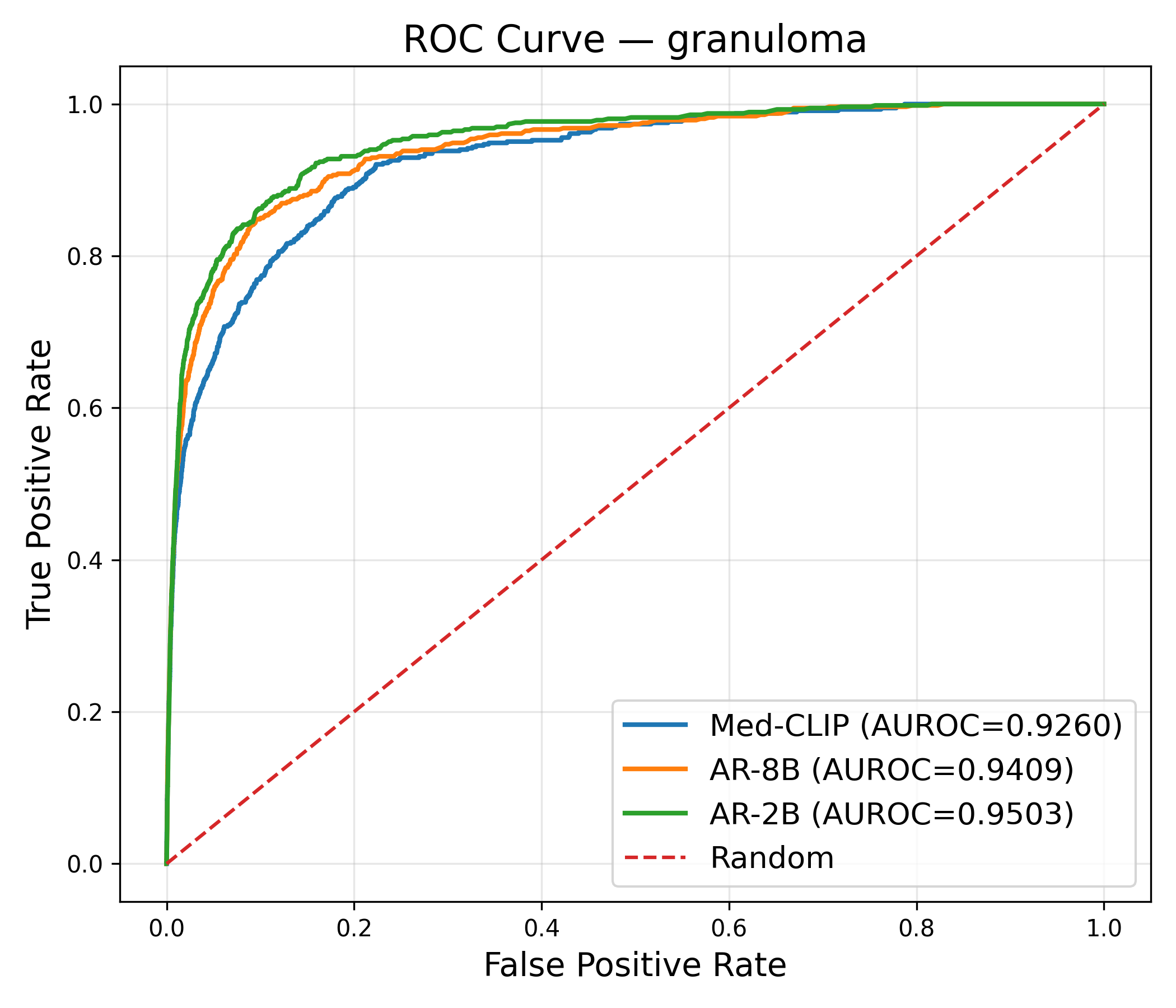}
    \caption{Internal -- Medium (\textit{granuloma})}
\end{subfigure}
\hfill
\begin{subfigure}{0.32\textwidth}
    \centering
    \includegraphics[width=\linewidth,height=0.15\textheight,keepaspectratio]{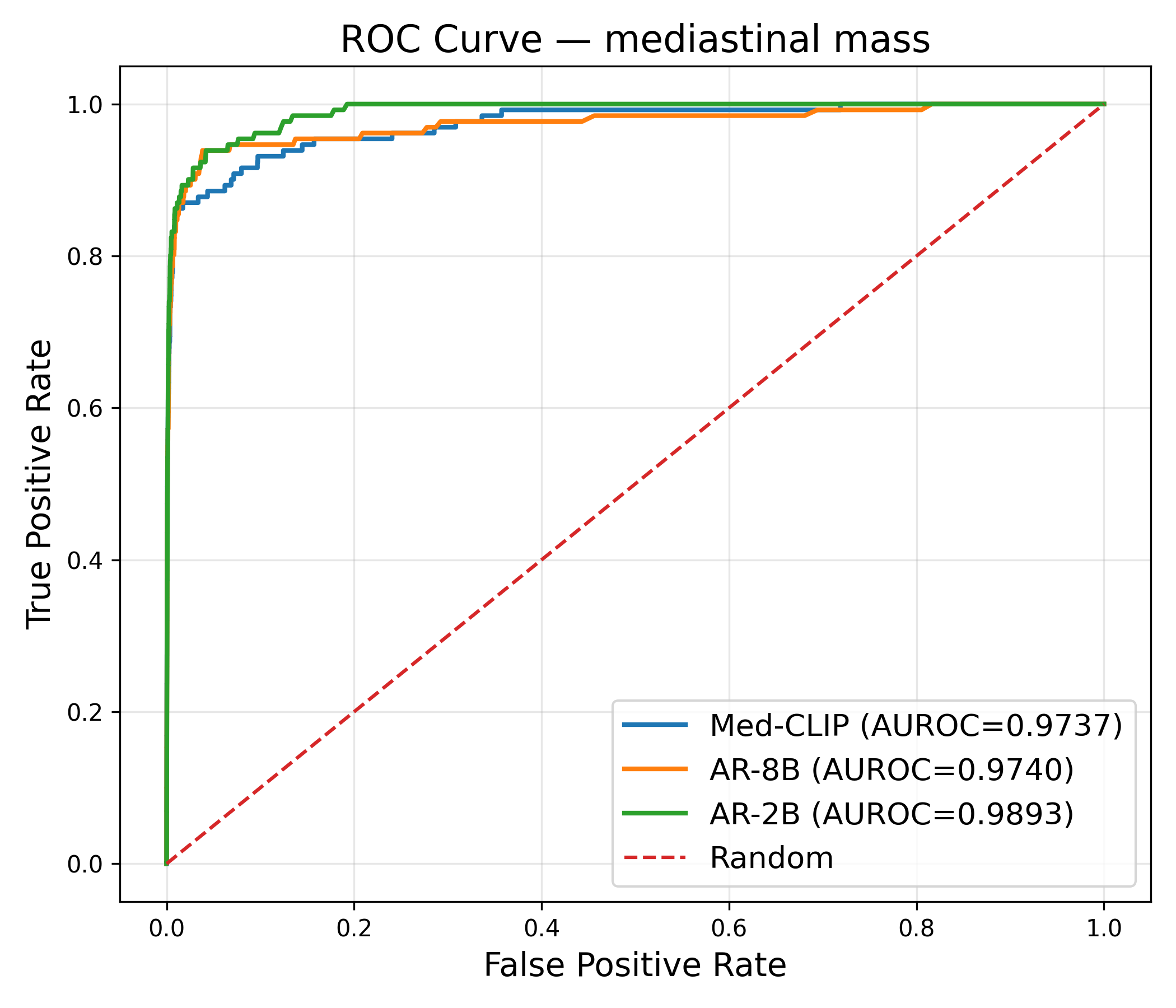}
    \caption{Internal -- Tail (\textit{mediastinal mass})}
\end{subfigure}

\vspace{0.15cm}

\begin{subfigure}{0.32\textwidth}
    \centering
    \includegraphics[width=\linewidth,height=0.15\textheight,keepaspectratio]{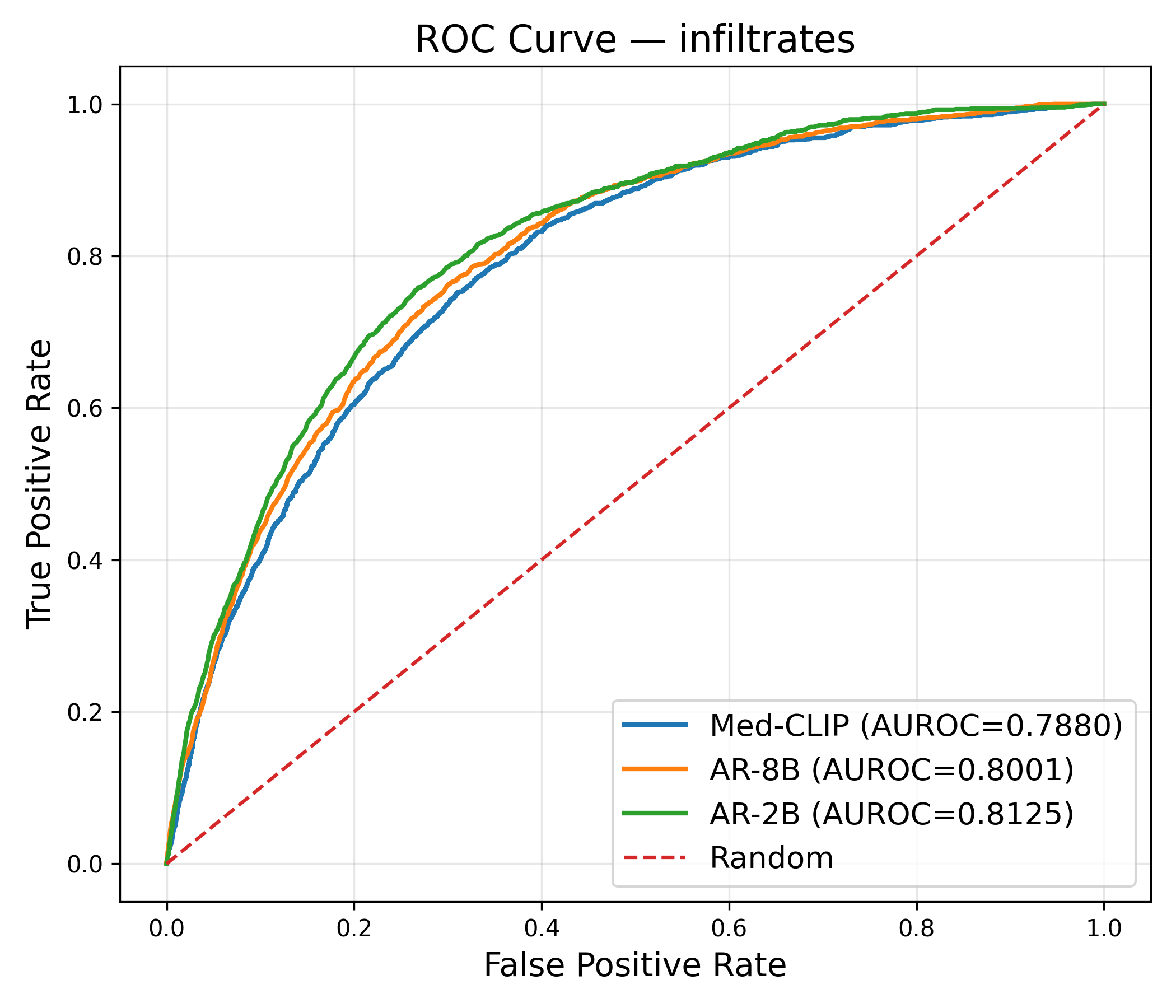}
    \caption{PadChest -- Head (\textit{infiltrates})}
\end{subfigure}
\hfill
\begin{subfigure}{0.32\textwidth}
    \centering
    \includegraphics[width=\linewidth,height=0.15\textheight,keepaspectratio]{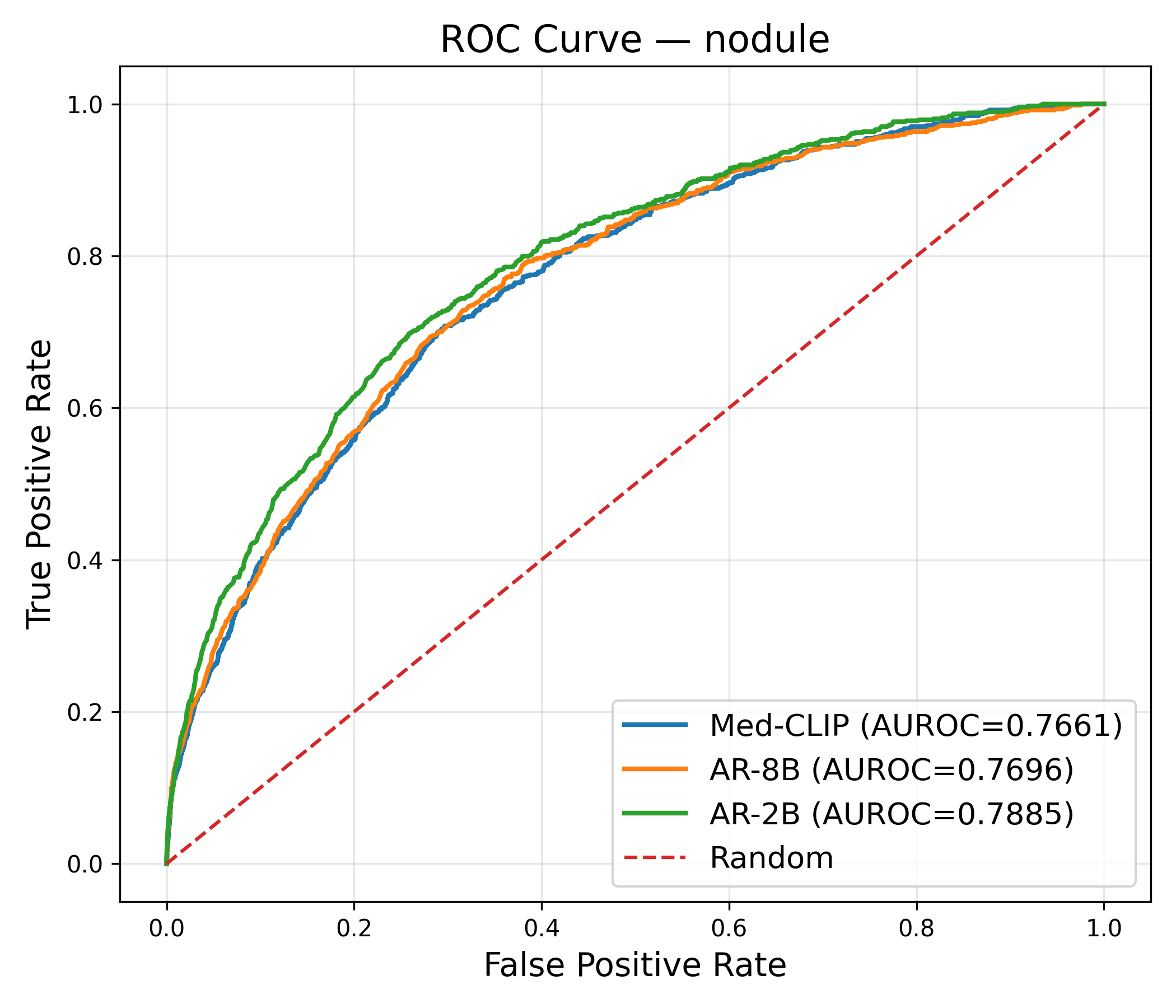}
    \caption{PadChest -- Medium (\textit{nodule})}
\end{subfigure}
\hfill
\begin{subfigure}{0.32\textwidth}
    \centering
    \includegraphics[width=\linewidth,height=0.15\textheight,keepaspectratio]{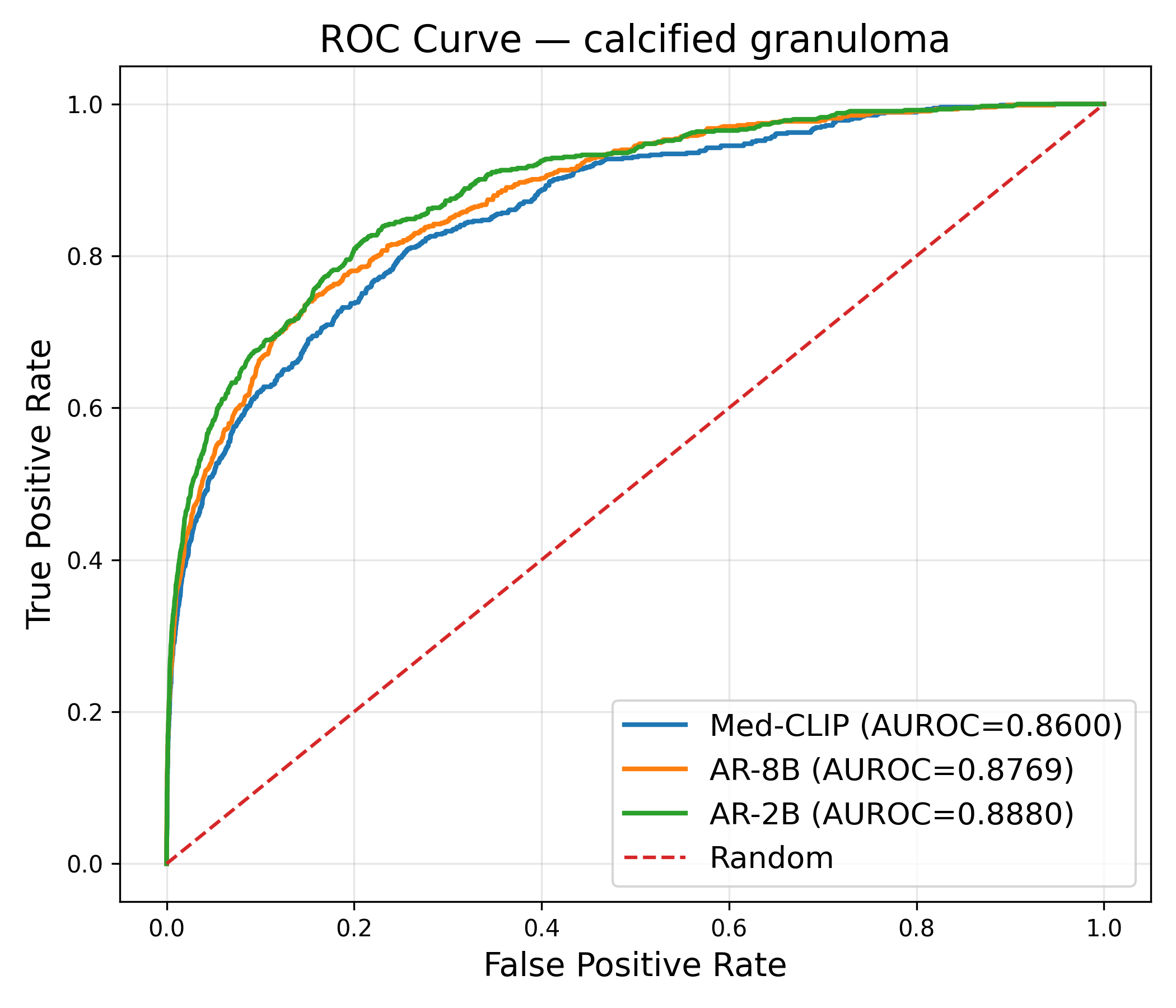}
    \caption{PadChest -- Tail (\textit{calcified granuloma})}
\end{subfigure}

\vspace{0.15cm}

\begin{subfigure}{0.32\textwidth}
    \centering
    \includegraphics[width=\linewidth,height=0.15\textheight,keepaspectratio]{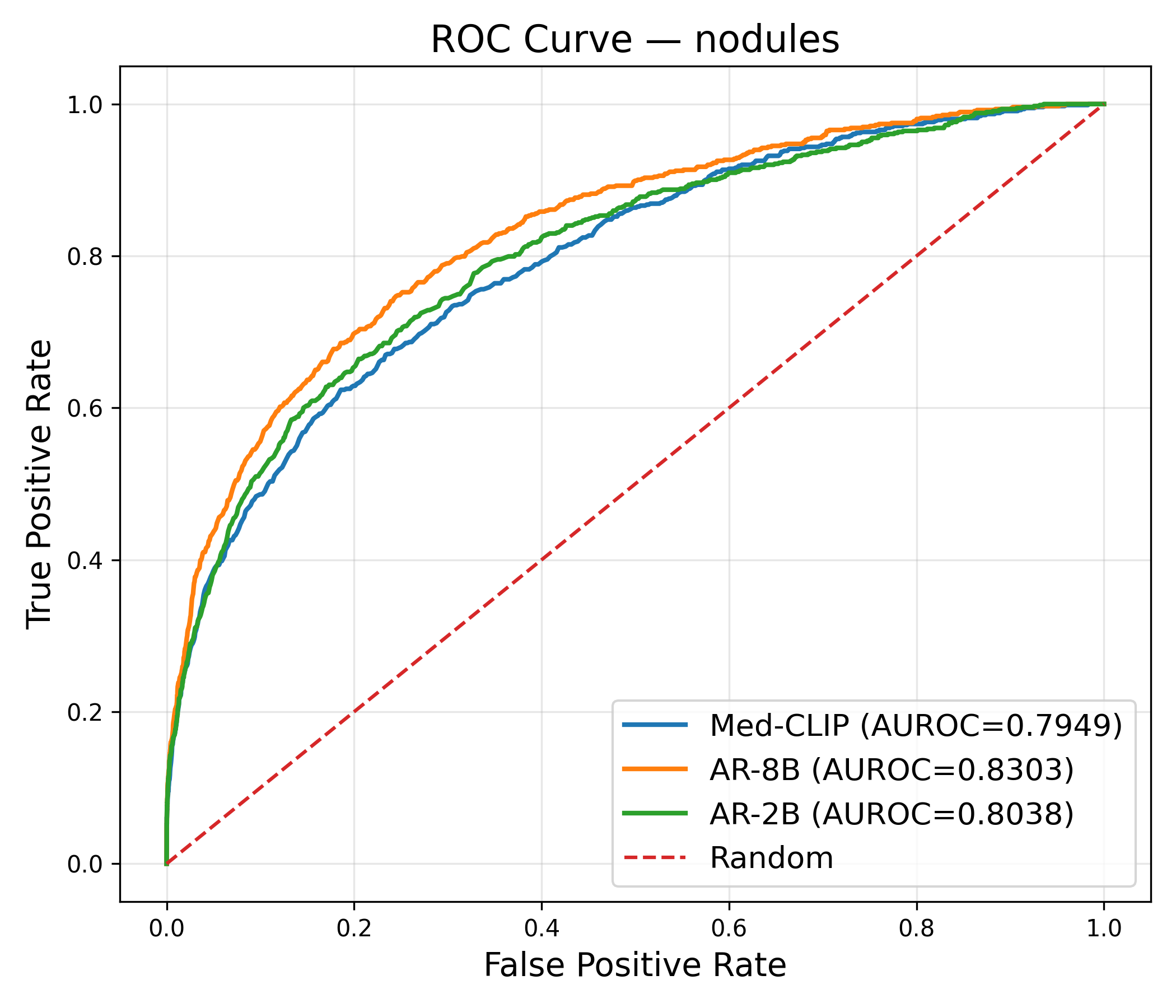}
    \caption{MIMIC-CXR -- Head (\textit{nodules})}
\end{subfigure}
\hfill
\begin{subfigure}{0.32\textwidth}
    \centering
    \includegraphics[width=\linewidth,height=0.15\textheight,keepaspectratio]{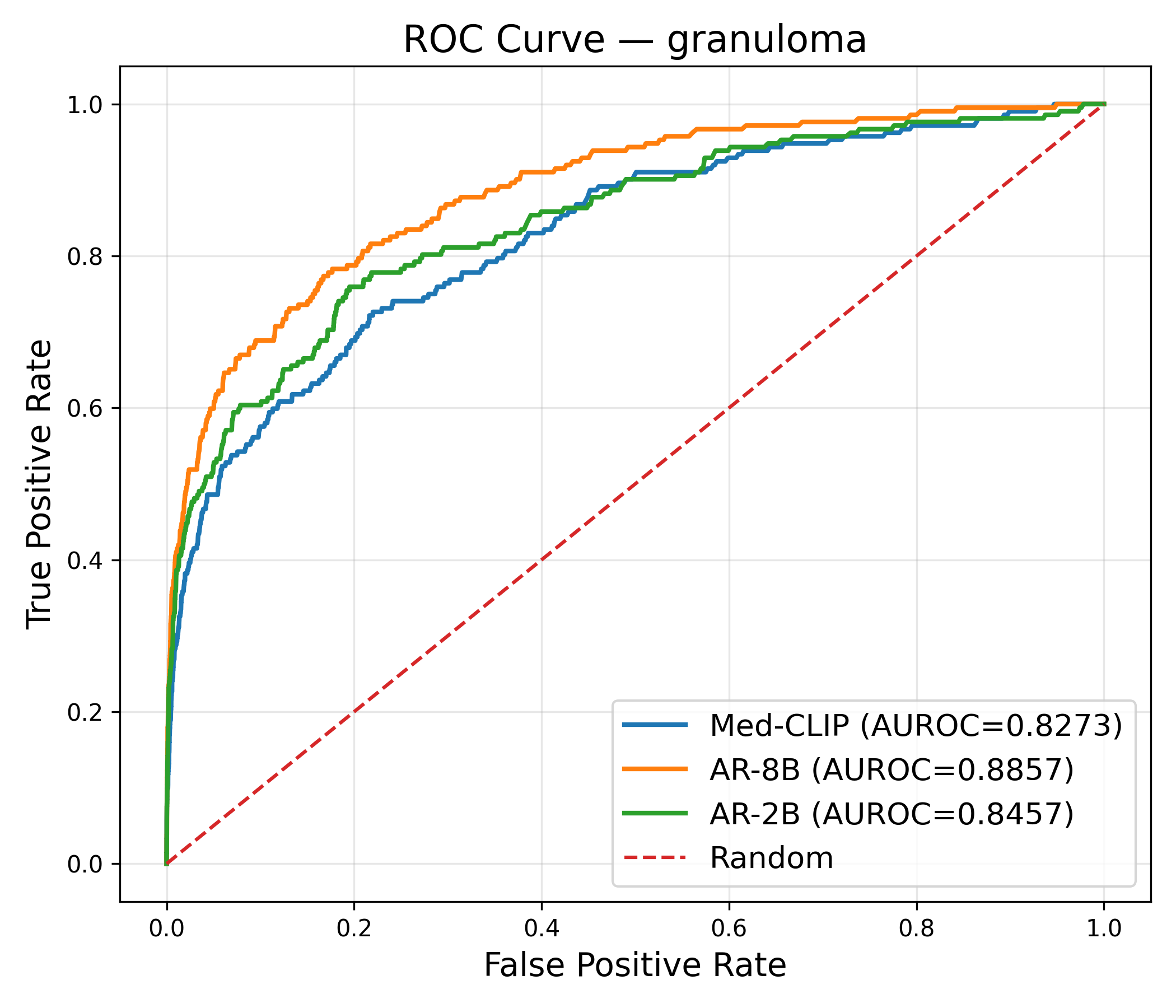}
    \caption{MIMIC-CXR -- Medium (\textit{granuloma})}
\end{subfigure}
\hfill
\begin{subfigure}{0.32\textwidth}
    \centering
    \includegraphics[width=\linewidth,height=0.15\textheight,keepaspectratio]{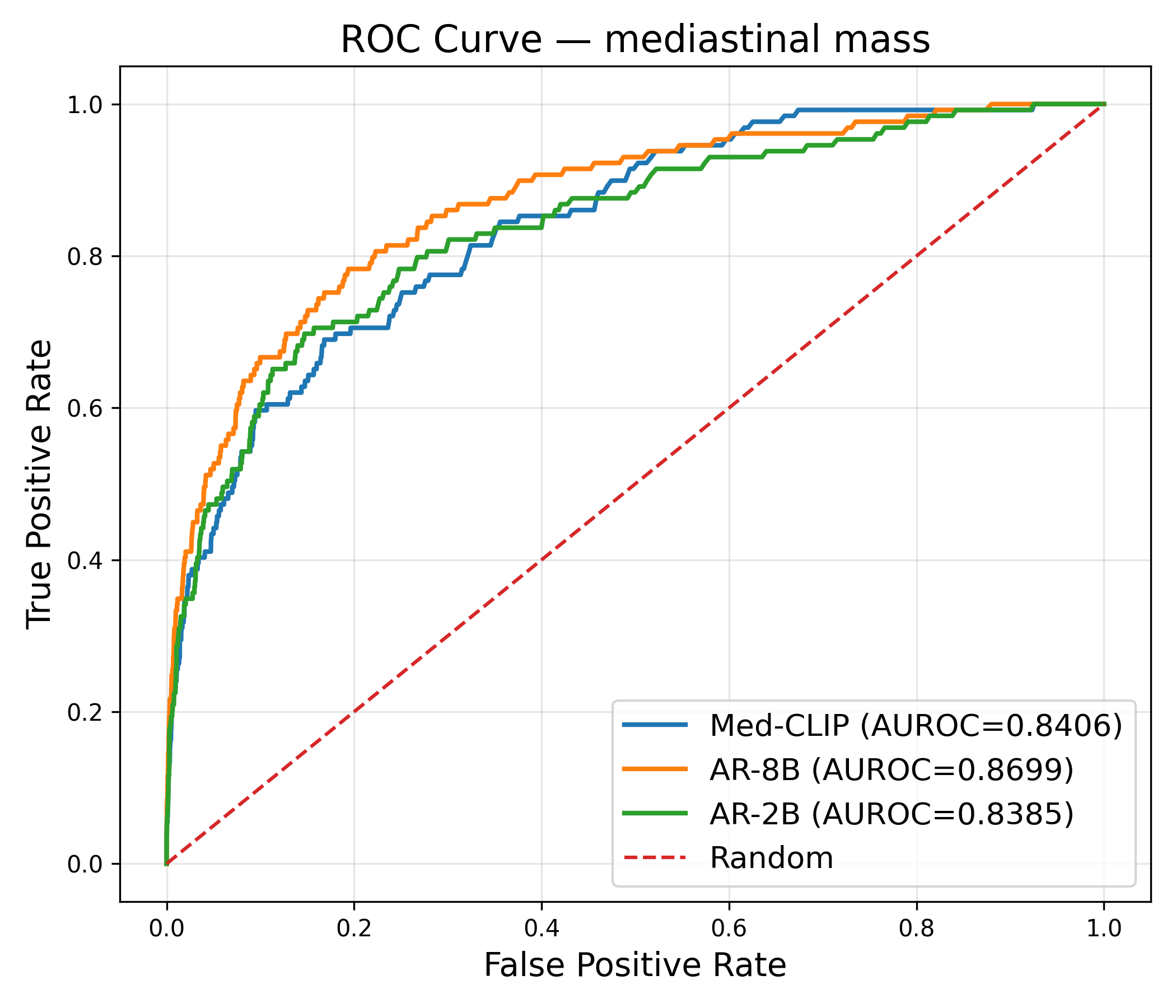}
    \caption{MIMIC-CXR -- Tail (\textit{mediastinal mass})}
\end{subfigure}

\vspace{0.15cm}

\begin{subfigure}{0.32\textwidth}
    \centering
    \includegraphics[width=\linewidth,height=0.15\textheight,keepaspectratio]{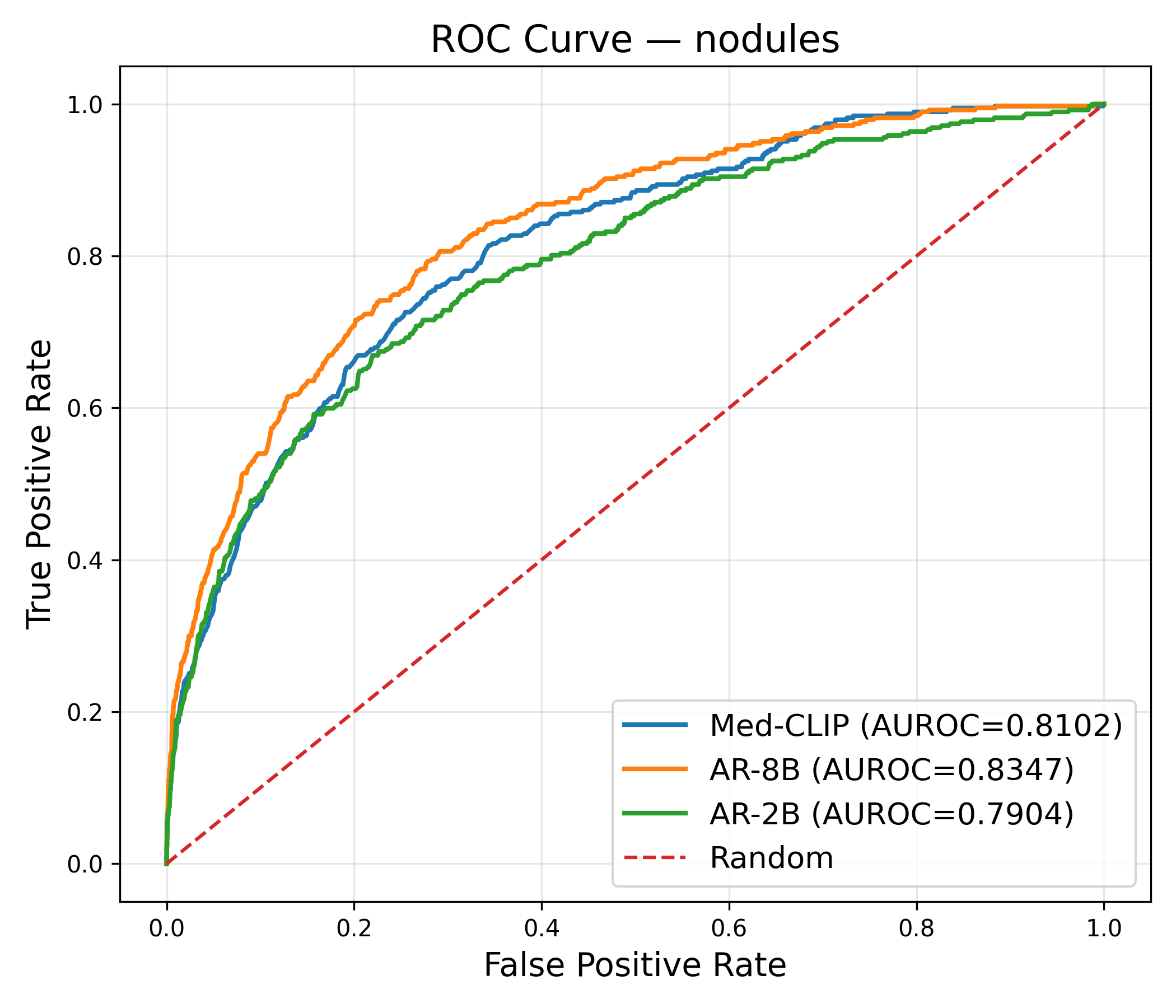}
    \caption{CheXpert -- Head (\textit{nodules})}
\end{subfigure}
\hfill
\begin{subfigure}{0.32\textwidth}
    \centering
    \includegraphics[width=\linewidth,height=0.15\textheight,keepaspectratio]{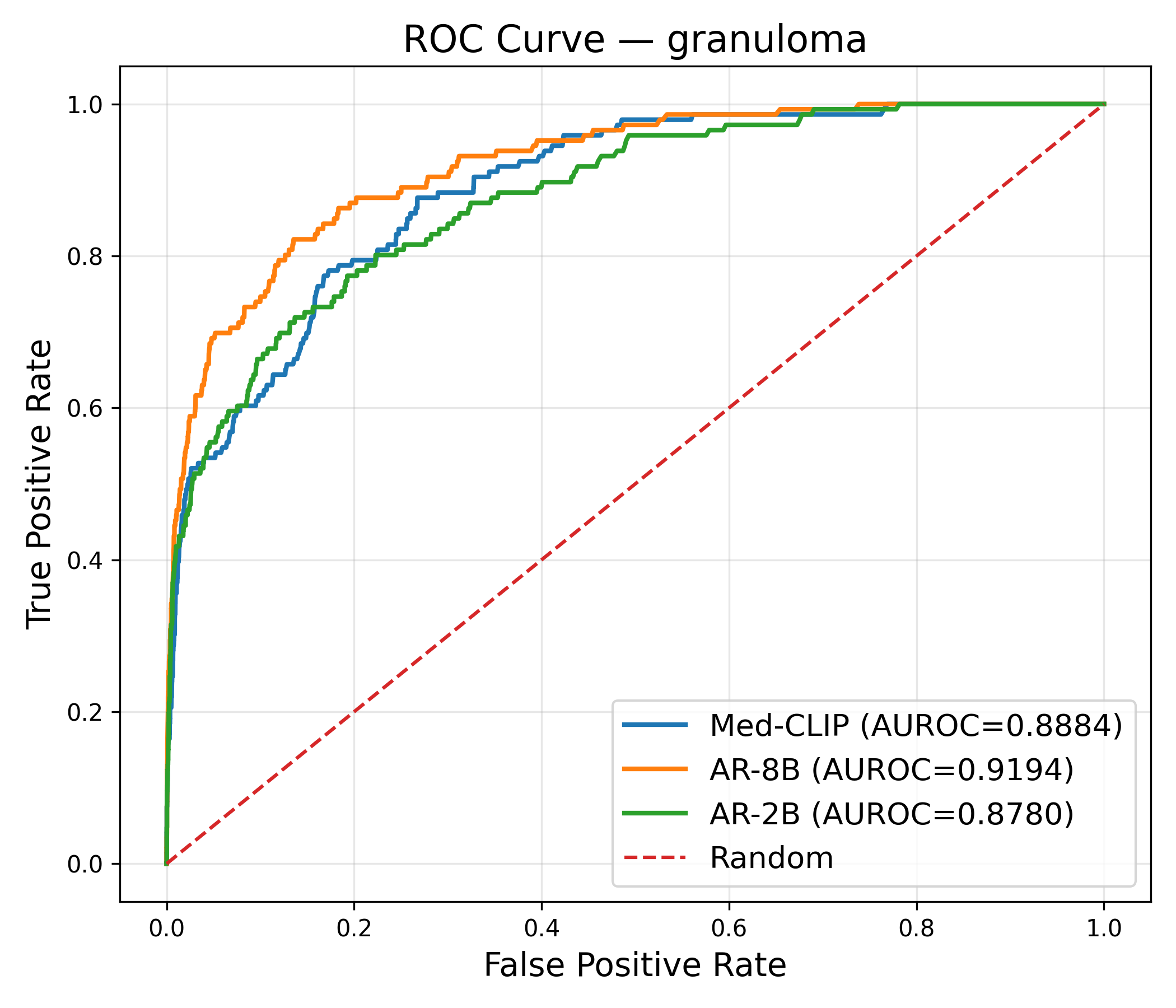}
    \caption{CheXpert -- Medium (\textit{granuloma})}
\end{subfigure}
\hfill
\begin{subfigure}{0.32\textwidth}
    \centering
    \includegraphics[width=\linewidth,height=0.15\textheight,keepaspectratio]{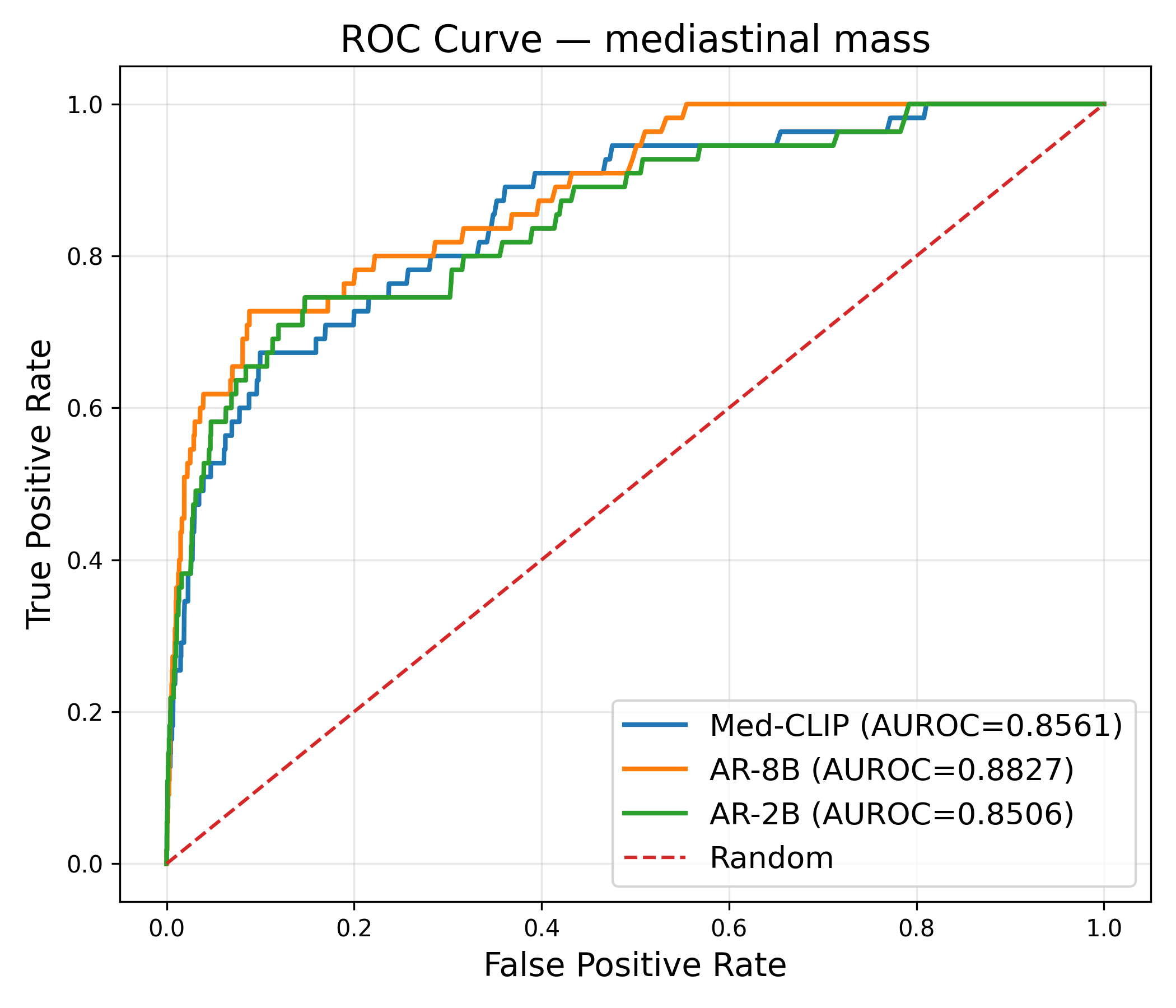}
    \caption{CheXpert -- Tail (\textit{mediastinal mass})}
\end{subfigure}

\caption{ROC curves comparing Med-CLIP, Med-AR-8B, and Med-AR-2B across four datasets. Rows correspond to Internal, PadChest, MIMIC-CXR, and CheXpert, while columns correspond to representative head-, medium-, and tail-prevalence abnormalities, respectively.}
\label{fig:roc_curves_4x3}
\end{figure*}

\subsection{Statistical significance across labels}
\label{subsec:stat}

To determine whether the aggregate improvements in Table~\ref{tab:groupwise_overall_head_medium_tail} are systematic across labels rather than driven by a small subset of abnormalities, we applied the Wilcoxon signed-rank test to paired per-label AUROC and AUPRC values. Table~\ref{tab:wilcoxon_auroc_auprc_head_medium_tail} reports results for the overall, head, medium, and tail groups, comparing both Med-AR-8B and Med-AR-2B against Med-CLIP.

The public benchmarks generally support the performance trends in Table~\ref{tab:groupwise_overall_head_medium_tail}. On PadChest, both Med-AR variants have nominally significant overall gains in AUROC and AUPRC. Med-AR-2B also has nominally significant gains in every subgroup; Med-AR-8B does so except for medium-label AUPRC ($p=0.0645$). On MIMIC-CXR, both variants have nominally significant AUROC and AUPRC gains in all groups. Effect magnitudes should be read from the metric differences, rather than inferred from the relative sizes of the $p$-values.

CheXpert shows the most asymmetric pattern between the two AR variants. Med-AR-8B achieves highly significant improvements over Med-CLIP across the overall benchmark and all prevalence groups, with AUROC p-values ranging from $2.17\mathrm{e}{-15}$ overall to $3.01\mathrm{e}{-06}$ on the tail subset, and AUPRC p-values from $1.08\mathrm{e}{-12}$ overall to $1.27\mathrm{e}{-03}$ on the tail subset. In contrast, Med-AR-2B fails to significantly outperform Med-CLIP on CheXpert: p-values are near 1.00 for all overall, head, and medium comparisons, and no significant AUPRC gain is observed on any subset. This directly corroborates the aggregate finding that Med-AR-2B does not surpass Med-CLIP on CheXpert and shows that the behavior is systematic at the per-label level rather than a metric-averaging artifact. Taken together with PadChest and MIMIC-CXR, these results indicate that the overall advantage of autoregressive pretraining on public benchmarks is broad and label-consistent, but depends critically on decoder scale: Med-AR-8B has nominally significant overall gains on all three public datasets, whereas Med-AR-2B has such gains on PadChest and MIMIC-CXR but not CheXpert.

The internal dataset presents a more nuanced picture, consistent with the broadly comparable aggregate results reported in the previous subsection. For Med-AR-8B, none of the Wilcoxon comparisons reach significance, indicating that its small aggregate improvements over Med-CLIP are not consistently expressed at the per-label level. In contrast, Med-AR-2B shows statistically significant AUROC gains for the overall, head, and medium groups, and also achieves a significant AUPRC improvement on the head subset. However, neither autoregressive variant shows significant gains on the internal tail subset, and most AUPRC comparisons remain non-significant. This pattern supports a balanced interpretation of the internal benchmark: Med-AR remains competitive and often favorable, particularly for AUROC and for the head and medium groups, but the differences relative to Med-CLIP are not uniformly strong enough to support the same clear superiority observed on the public datasets.

\FloatBarrier
\begin{table}[H]
\centering
\caption{Wilcoxon signed-rank test results comparing autoregressive and CLIP-based pretraining using per-label AUROC and AUPRC across datasets for all, head, medium, and tail tags.}
\label{tab:wilcoxon_auroc_auprc_head_medium_tail}
\scriptsize
\setlength{\tabcolsep}{5pt}
\renewcommand{\arraystretch}{1.08}
\begin{tabular}{lllcccc}
\toprule
\textbf{Dataset} & \makecell{\textbf{Tag}\\\textbf{Group}} & \textbf{Model} & \makecell{\textbf{AUROC}\\\textbf{Stat.}} & \makecell{\textbf{AUROC}\\\textbf{p-val}} & \makecell{\textbf{AUPRC}\\\textbf{Stat.}} & \makecell{\textbf{AUPRC}\\\textbf{p-val}} \\
\midrule
\multirow{8}{*}{Internal}
& \multirow{2}{*}{All}    & AR-8B & 352.0 & 0.1578   & 533.0 & 0.9080 \\
&                         & AR-2B & 215.0 & 0.0023   & 495.0 & 0.7982 \\
& \multirow{2}{*}{Head}   & AR-8B & 27.0  & 0.0594   & 39.0  & 0.2131 \\
&                         & AR-2B & 17.0  & 0.0123   & 25.0  & 0.0453 \\
& \multirow{2}{*}{Medium} & AR-8B & 31.0  & 0.1698   & 64.0  & 0.9045 \\
&                         & AR-2B & 21.0  & 0.0471   & 57.0  & 0.7928 \\
& \multirow{2}{*}{Tail}   & AR-8B & 57.0  & 0.6196   & 82.0  & 0.9710 \\
&                         & AR-2B & 37.0  & 0.1788   & 90.0  & 0.9933 \\
\midrule
\multirow{8}{*}{PadChest}
& \multirow{2}{*}{All}    & AR-8B & 48.0 & 4.36e-05 & 43.0 & 2.35e-05 \\
&                         & AR-2B & 1.0  & 3.73e-09 & 4.0  & 1.30e-08 \\
& \multirow{2}{*}{Head}   & AR-8B & 6.0  & 0.0137   & 9.0  & 0.0322 \\
&                         & AR-2B & 0.0  & 9.77e-04 & 0.0  & 9.77e-04 \\
& \multirow{2}{*}{Medium} & AR-8B & 3.0  & 0.0098   & 9.0  & 0.0645 \\
&                         & AR-2B & 0.0  & 0.00195  & 0.0  & 0.00195 \\
& \multirow{2}{*}{Tail}   & AR-8B & 7.0  & 0.0186   & 1.0  & 0.00195 \\
&                         & AR-2B & 1.0  & 0.00195  & 1.0  & 0.00195 \\
\midrule
\multirow{8}{*}{MIMIC}
& \multirow{2}{*}{All}    & AR-8B & 206.0  & 1.04e-17 & 172.0  & 4.33e-18 \\
&                         & AR-2B & 1032.0 & 8.42e-10 & 1164.0 & 8.90e-09 \\
& \multirow{2}{*}{Head}   & AR-8B & 0.0    & 1.46e-11 & 1.0    & 2.91e-11 \\
&                         & AR-2B & 13.0   & 1.28e-09 & 60.0   & 1.43e-06 \\
& \multirow{2}{*}{Medium} & AR-8B & 10.0   & 1.56e-10 & 16.0   & 6.15e-10 \\
&                         & AR-2B & 189.0  & 0.0038   & 158.0  & 7.75e-04 \\
& \multirow{2}{*}{Tail}   & AR-8B & 53.0   & 6.49e-07 & 43.0   & 1.88e-07 \\
&                         & AR-2B & 127.0  & 4.08e-04 & 185.0  & 0.00959 \\
\midrule
\multirow{8}{*}{CheXpert}
& \multirow{2}{*}{All}    & AR-8B & 524.0  & 2.17e-15 & 818.0  & 1.08e-12 \\
&                         & AR-2B & 5418.0 & 1.00e+00 & 4175.0 & 0.9905 \\
& \multirow{2}{*}{Head}   & AR-8B & 3.0    & 1.82e-11 & 31.0   & 8.64e-09 \\
&                         & AR-2B & 643.0  & 1.00e+00 & 555.0  & 0.9969 \\
& \multirow{2}{*}{Medium} & AR-8B & 72.0   & 6.41e-07 & 92.0   & 4.08e-06 \\
&                         & AR-2B & 671.0  & 1.00e+00 & 477.0  & 0.8876 \\
& \multirow{2}{*}{Tail}   & AR-8B & 81.0   & 3.01e-06 & 167.0  & 0.00127 \\
&                         & AR-2B & 549.0  & 0.9958   & 408.0  & 0.7072 \\
\bottomrule
\end{tabular}
\end{table}
\FloatBarrier

Overall, the Wilcoxon analysis reinforces the central narrative of the paper. On PadChest, MIMIC-CXR, and CheXpert, autoregressive pretraining---most consistently through Med-AR-8B---has nominally significant improvements over Med-CLIP in most prevalence-group comparisons, whereas on the internal benchmark the two paradigms are better characterized as broadly comparable, with the strongest support for Med-AR concentrated in Med-AR-2B and in the higher-prevalence groups.

\subsection{Label-wise AUROC differences}

\begin{figure*}[!t]
    \centering

    \begin{subfigure}{0.9\textwidth}
        \centering
        \includegraphics[width=\linewidth]{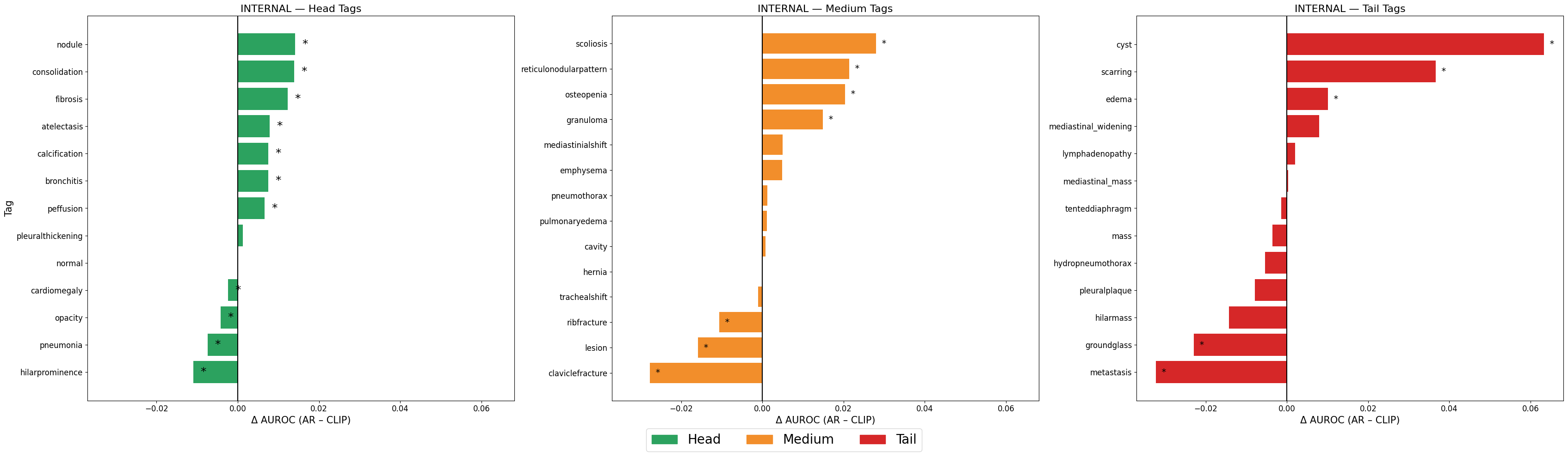}
        \caption{Internal}
        \label{fig:delong_internal}
    \end{subfigure}

    \vspace{0.12cm}

    \begin{subfigure}{0.9\textwidth}
        \centering
        \includegraphics[width=\linewidth]{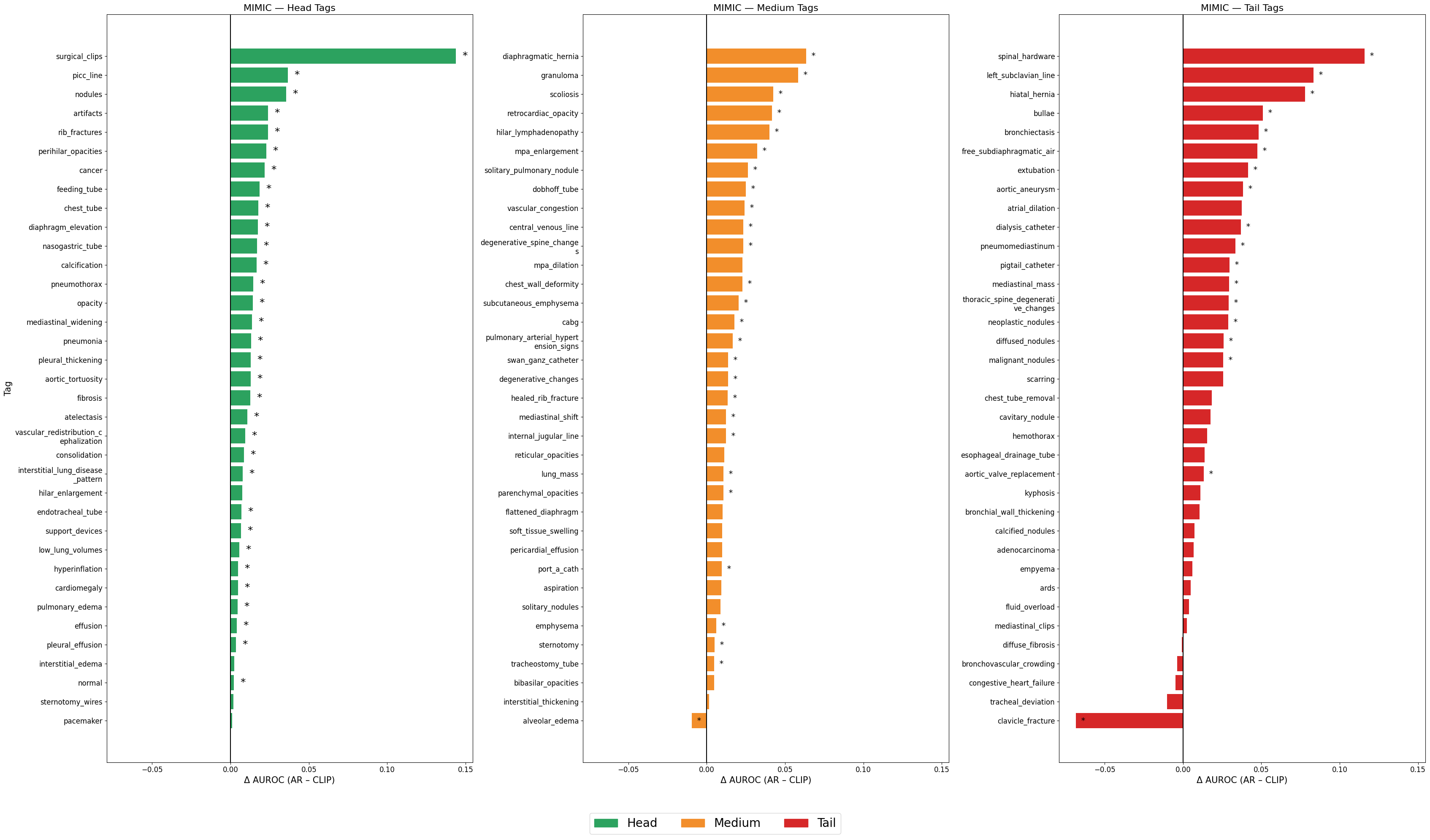}
        \caption{MIMIC-CXR}
        \label{fig:delong_mimic}
    \end{subfigure}

    \caption{Differences in AUROC, $\Delta$, between autoregressive and Med-CLIP pretraining for (a) Internal and (b) MIMIC-CXR. Statistically significant differences under the DeLong test are marked with (*).}
    \label{fig:delong_internal_mimic}
\end{figure*}

\begin{figure*}[!t]
    \centering

    \begin{subfigure}{0.9\textwidth}
        \centering
        \includegraphics[width=\linewidth]{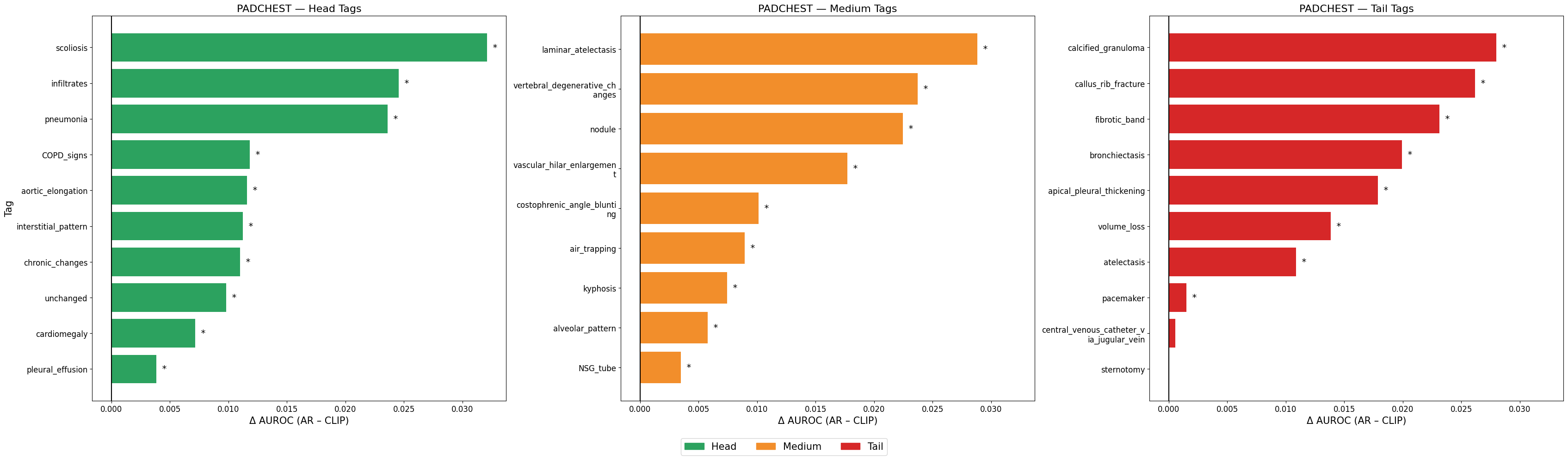}
        \caption{PadChest}
        \label{fig:delong_padchest}
    \end{subfigure}

    \vspace{0.12cm}

    \begin{subfigure}{0.9\textwidth}
        \centering
        \includegraphics[width=\linewidth]{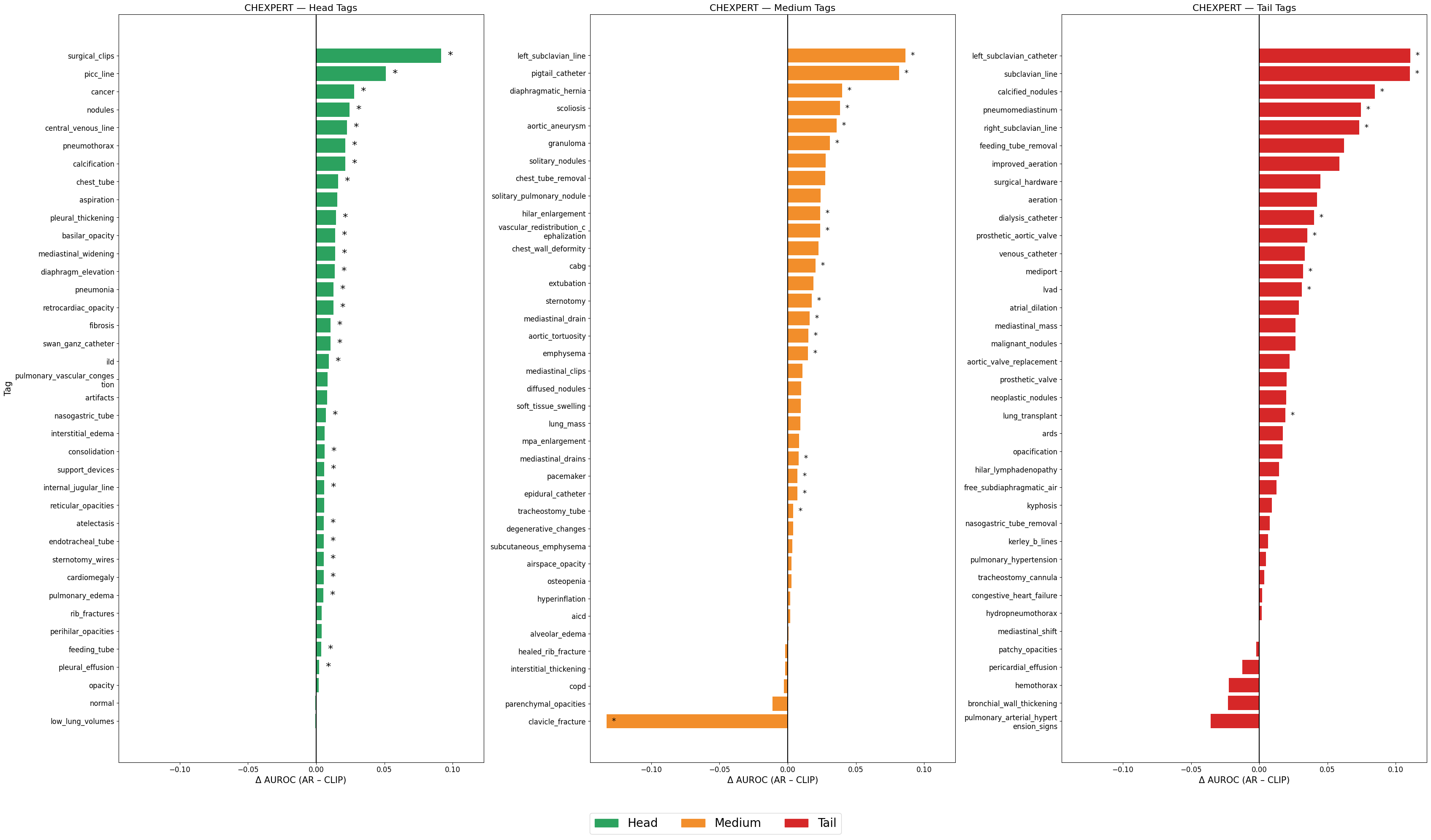}
        \caption{CheXpert}
        \label{fig:delong_chexpert}
    \end{subfigure}

    \caption{Differences in AUROC, $\Delta$, between autoregressive and Med-CLIP pretraining for (a) PadChest and (b) CheXpert. Statistically significant differences under the DeLong test are marked with (*).}
    \label{fig:delong_padchest_chexpert}
\end{figure*}

We further examined class-wise behavior using DeLong's test for paired AUROCs. Figures~\ref{fig:delong_internal_mimic} and~\ref{fig:delong_padchest_chexpert} visualize the AUROC difference between autoregressive pretraining and Med-CLIP for each abnormality, separately for the head, medium, and tail subsets in the Internal, MIMIC-CXR, PadChest, and CheXpert benchmarks. Statistically significant differences under DeLong's test are marked in the figures.

The public datasets again show a clear advantage for autoregressive pretraining. On MIMIC-CXR, the majority of labels across the head, medium, and tail groups show positive AUROC differences, and many of these gains are statistically significant. The largest improvements are particularly evident among several medium and tail abnormalities, supporting the view that autoregressive supervision is especially beneficial when clinically relevant findings are less frequent or semantically more specific. PadChest exhibits a similarly favorable pattern, with consistently positive shifts across all three prevalence groups and multiple significant improvements among both common and infrequent abnormalities. CheXpert follows the same trend when compared against Med-AR-8B: positive AUROC differences dominate the head, medium, and tail subsets, and a substantial fraction of labels reach statistical significance, including several rare abnormalities. This label-level evidence complements the Wilcoxon analysis and confirms that the Med-AR-8B gains on CheXpert are distributed across many abnormalities rather than concentrated in a handful of high-prevalence classes. Consistent with the earlier aggregate and Wilcoxon findings, Med-AR-2B on CheXpert does not produce a comparable distribution of positive significant differences, again indicating that the CheXpert benefit is tied to decoder scale.

The internal dataset is more heterogeneous. In the head and medium groups, the class-wise AUROC differences mostly favor autoregressive pretraining, consistent with the Wilcoxon results for AR-2B. However, the internal tail subset remains mixed, with some abnormalities benefiting from AR and others continuing to favor Med-CLIP. This label-wise heterogeneity explains why the internal benchmark, despite favorable aggregate metrics for AR, yields more limited statistical support at the tail level. Overall, the DeLong analysis reinforces the central pattern of the paper: autoregressive pretraining produces broad and consistent gains on the public datasets, while the internal benchmark shows a more selective advantage concentrated in higher-prevalence groups and clinically important individual findings.

\subsection{Uncertainty-aware evaluation}

Table~\ref{tab:eaurc_clip_ar8b_ar2b_head_mid_tail} reports Excess Area Under the Risk--Coverage Curve (EAURC), where lower values indicate better uncertainty ranking and, therefore, more effective selective prediction. This analysis complements the discrimination results by testing whether model confidence is informative about prediction correctness, rather than only whether the predicted scores separate positives from negatives.

The public datasets again favor autoregressive pretraining, although the best-performing variant varies across them. On MIMIC-CXR, Med-AR-8B achieves the lowest EAURC across all groups, reducing the overall value from 0.196 for Med-CLIP to 0.169, with consistent improvements for head labels (0.365 to 0.334), medium labels (0.186 to 0.146), and tail labels (0.038 to 0.027). Med-AR-2B also improves over Med-CLIP in every group, although it remains slightly weaker than Med-AR-8B. A similar trend is observed on PadChest, where Med-AR-8B again provides the strongest uncertainty ranking, reducing EAURC from 0.314 to 0.217 overall, from 0.397 to 0.306 for head labels, from 0.351 to 0.240 for medium labels, and from 0.198 to 0.107 for tail labels. Med-AR-2B is consistently second-best on both datasets.

The CheXpert results add an important and somewhat unexpected nuance. Both autoregressive variants outperform Med-CLIP on EAURC across all prevalence groups, but here \emph{Med-AR-2B}, not Med-AR-8B, achieves the best uncertainty ranking. Specifically, Med-AR-2B reduces EAURC from 0.212 (Med-CLIP) to 0.162 overall, from 0.406 to 0.356 for head labels, from 0.190 to 0.113 for medium labels, and from 0.040 to 0.018 for tail labels, with Med-AR-8B second-best throughout. This is particularly notable because Med-AR-2B has \emph{lower} AUROC and AUPRC than Med-CLIP on CheXpert. In other words, stronger discrimination does not automatically translate into better uncertainty ranking: despite being a weaker classifier on CheXpert, Med-AR-2B produces confidence scores that rank its own errors more effectively than either Med-CLIP or Med-AR-8B. This dissociation between discrimination and selective-prediction behavior highlights the value of evaluating these two axes independently when comparing pretraining paradigms.

In contrast, the internal benchmark shows the opposite pattern. Here, Med-CLIP achieves the lowest EAURC overall and within each subgroup, with values of 0.014 overall, 0.036 for head labels, 0.004 for medium labels, and 0.002 for tail labels. Both Med-AR variants perform worse on this benchmark, with Med-AR-8B ahead of Med-AR-2B overall and for head labels, and Med-AR-2B ahead for medium and tail labels. Both remain above Med-CLIP in every group. This is consistent with the more mixed behavior already observed on the internal dataset in the previous subsections: although Med-AR remains competitive in discrimination metrics, those gains do not translate into better confidence ordering in this more heterogeneous setting.

Taken together, the uncertainty-aware evaluation reinforces the broader narrative of the paper while adding two important qualifications. First, on all three public benchmarks, Med-AR-8B improves AUROC, AUPRC, and EAURC relative to Med-CLIP under the evaluated protocol; however, the optimal AR variant for uncertainty ranking is not always the one that maximizes discrimination, as clearly demonstrated on CheXpert. Second, on the internal dataset Med-CLIP retains a clear advantage in EAURC, suggesting that stronger representation transfer and better uncertainty ordering need not always coincide. These observations make uncertainty-aware evaluation an essential complementary axis for comparing pretraining paradigms in clinical imaging.

\begin{table}[ht]
\centering
\setlength{\tabcolsep}{10pt}
\renewcommand{\arraystretch}{1.3}
\caption{Excess Area Under the Risk--Coverage Curve (EAURC) for Med-CLIP, Med-AR-8B, and Med-AR-2B across datasets. Values are averaged over all labels (Overall) and separately over head, medium, and tail labels. Lower values indicate better uncertainty ranking. The best value in each row is shown in bold, and the second-best value is underlined.}
\label{tab:eaurc_clip_ar8b_ar2b_head_mid_tail}
\begin{tabular}{|l|c|c|c|c|}
\hline
\hline
\textbf{Dataset} 
& \textbf{Group}  
& \makecell[c]{\textbf{Med-CLIP} \\ \textbf{EAURC}} 
& \makecell[c]{\textbf{Med-AR-8B} \\[-0.55ex]\smash{\scriptsize(ours)} \\ \textbf{EAURC}}
& \makecell[c]{\textbf{Med-AR-2B} \\[-0.55ex]\smash{\scriptsize(ours)} \\ \textbf{EAURC}} \\
\hline
\multirow{4}{*}{Internal} 
& Overall & \textbf{0.014} & \underline{0.070} & 0.073 \\
& Head    & \textbf{0.036} & \underline{0.173} & 0.194 \\
& Medium  & \textbf{0.004} & 0.018 & \underline{0.016} \\
& Tail    & \textbf{0.002} & 0.022 & \underline{0.013} \\
\hline
\multirow{4}{*}{MIMIC} 
& Overall & 0.196 & \textbf{0.169} & \underline{0.179} \\
& Head    & 0.365 & \textbf{0.334} & \underline{0.345} \\
& Medium  & 0.186 & \textbf{0.146} & \underline{0.165} \\
& Tail    & 0.038 & \textbf{0.027} & \underline{0.028} \\
\hline
\multirow{4}{*}{PadChest} 
& Overall & 0.314 & \textbf{0.217} & \underline{0.274} \\
& Head    & 0.397 & \textbf{0.306} & \underline{0.358} \\
& Medium  & 0.351 & \textbf{0.240} & \underline{0.315} \\
& Tail    & 0.198 & \textbf{0.107} & \underline{0.153} \\
\hline
\multirow{4}{*}{CheXpert} 
& Overall & 0.212 & \underline{0.191} & \textbf{0.162} \\
& Head    & 0.406 & \underline{0.384} & \textbf{0.356} \\
& Medium  & 0.190 & \underline{0.159} & \textbf{0.113} \\
& Tail    & 0.040 & \underline{0.031} & \textbf{0.018} \\
\hline
\hline
\end{tabular}
\end{table}
\subsection{Comparison of pretrained encoders}

\FloatBarrier

\begin{table*}[t]
\centering
\scriptsize
\setlength{\tabcolsep}{2.6pt}
\renewcommand{\arraystretch}{1.03}
\caption{Comparison of vision encoders initialized from different methods across CheXpert, MIMIC-CXR, and PadChest using mean AUROC and mean AUPRC. All encoders are fine-tuned with the same ML-Decoder head. Within each dataset, metric, and prevalence group, the best result is shown in bold and the second-best is underlined. H=head, M=medium, and T=tail.}
\label{tab:encoder_ablation_multidataset_compact}
\begin{tabular}{ll|cc|cc|cc}
\toprule
\multirow{2}{*}{\textbf{Method}} & \multirow{2}{*}{\textbf{Grp}} 
& \multicolumn{2}{c|}{\textbf{CheXpert}} 
& \multicolumn{2}{c|}{\textbf{MIMIC-CXR}} 
& \multicolumn{2}{c}{\textbf{PadChest}} \\
& & \textbf{mAUROC} & \textbf{mAUPRC} & \textbf{mAUROC} & \textbf{mAUPRC} & \textbf{mAUROC} & \textbf{mAUPRC} \\
\midrule

\multirow{3}{*}{EVA-Base}
& H & 0.7981 & 0.2918 & 0.8250 & 0.3013 & 0.7961 & 0.3463 \\
& M & 0.8223 & 0.0996 & 0.8115 & 0.0804 & 0.7934 & 0.2480 \\
& T & 0.7843 & 0.0333 & 0.7952 & 0.0326 & 0.8185 & 0.2581 \\
\midrule

\multirow{3}{*}{RAD-DINO}
& H & 0.7632 & 0.2473 & 0.8440 & 0.3287 & 0.7777 & 0.3109 \\
& M & 0.7847 & 0.0614 & 0.8380 & 0.1032 & 0.7764 & 0.2224 \\
& T & 0.7460 & 0.0193 & 0.8289 & 0.0397 & 0.8056 & 0.2406 \\
\midrule

\multirow{3}{*}{ARK}
& H & 0.8430 & 0.3680 & 0.8837 & 0.4202 & 0.8419 & 0.4384 \\
& M & 0.8593 & 0.1451 & 0.8809 & 0.1817 & 0.8510 & 0.3633 \\
& T & 0.8325 & 0.0649 & 0.8816 & 0.1111 & 0.9047 & 0.4747 \\
\midrule

\multirow{3}{*}{CheXFound}
& H & 0.8429 & 0.3692 & 0.8809 & 0.4098 & 0.8487 & 0.4558 \\
& M & 0.8670 & 0.1665 & 0.8849 & 0.1987 & 0.8584 & 0.3770 \\
& T & 0.8384 & 0.0815 & 0.8736 & 0.1008 & 0.8950 & 0.4476 \\
\midrule

\multirow{3}{*}{BiomedCLIP}
& H & 0.7392 & 0.2360 & 0.8582 & 0.3719 & 0.6630 & 0.2034 \\
& M & 0.7604 & 0.0715 & 0.8536 & 0.1542 & 0.6705 & 0.1471 \\
& T & 0.7311 & 0.0269 & 0.8425 & 0.0791 & 0.7241 & 0.1970 \\
\midrule

\multirow{3}{*}{MedKLIP}
& H & 0.7847 & 0.2786 & 0.8112 & 0.2839 & 0.8132 & 0.3812 \\
& M & 0.7980 & 0.0777 & 0.7866 & 0.0621 & 0.8172 & 0.3045 \\
& T & 0.7411 & 0.0201 & 0.7628 & 0.0298 & 0.8344 & 0.3406 \\
\midrule

\multirow{3}{*}{BioViL-T}
& H & 0.6604 & 0.1676 & 0.7117 & 0.1881 & 0.6626 & 0.1858 \\
& M & 0.6217 & 0.0200 & 0.6424 & 0.0207 & 0.6632 & 0.1510 \\
& T & 0.6113 & 0.0072 & 0.5833 & 0.0061 & 0.7104 & 0.2087 \\
\midrule

\multirow{3}{*}{MedCLIP}
& H & \underline{0.8433} & \underline{0.3710} & 0.8756 & 0.4010 & 0.8425 & 0.4428 \\
& M & \underline{0.8790} & \underline{0.1802} & 0.8841 & 0.1902 & 0.8531 & 0.3678 \\
& T & \underline{0.8526} & \underline{0.0922} & 0.8733 & 0.1033 & 0.9002 & 0.4762 \\
\midrule

\multirow{3}{*}{\makecell[c]{Med-AR-2B\\[-0.45ex]\scriptsize(ours)}}
& H & 0.8310 & 0.3614 & \underline{0.8846} & \underline{0.4264} & \textbf{0.8571} & \textbf{0.4706} \\
& M & 0.8586 & 0.1682 & \underline{0.8881} & \underline{0.2175} & \textbf{0.8674} & \textbf{0.3941} \\
& T & 0.8365 & 0.0829 & \underline{0.8851} & \underline{0.1185} & \textbf{0.9143} & \textbf{0.5072} \\
\midrule

\multirow{3}{*}{\makecell[c]{Med-AR-8B\\[-0.45ex]\scriptsize(ours)}}
& H & \textbf{0.8565} & \textbf{0.3978} & \textbf{0.8918} & \textbf{0.4432} & \underline{0.8497} & \underline{0.4583} \\
& M & \textbf{0.8920} & \textbf{0.2054} & \textbf{0.9028} & \textbf{0.2435} & \underline{0.8595} & \underline{0.3818} \\
& T & \textbf{0.8788} & \textbf{0.1151} & \textbf{0.8967} & \textbf{0.1441} & \underline{0.9060} & \underline{0.4940} \\
\bottomrule
\end{tabular}
\end{table*}

The uncertainty-aware evaluation above compares Med-AR directly against Med-CLIP. We next broaden that comparison to ask whether the observed gains persist when the image encoder is initialized from a wider set of representation learning strategies, while keeping the downstream classifier fixed. To this end, we replace the pretrained encoder with alternatives spanning self-supervised, supervised, biomedical contrastive, knowledge-enhanced, and radiology-specific vision--language pretraining, and fine-tune all of them under the same ML-Decoder downstream setup on CheXpert, MIMIC-CXR, and PadChest.

Specifically, we compare RAD-DINO, an image-only self-supervised medical encoder trained beyond text supervision~\cite{PerezGarcia2025RADDINO}; ARK (Ark$^{+}$), a supervised chest radiography foundation model trained from heterogeneous expert-labeled datasets without manual label consolidation~\cite{ma2025fully,ma2023foundation,ma2026arkppp}; BiomedCLIP, a large-scale biomedical contrastive model pretrained on scientific image--text pairs~\cite{Zhang2023BiomedCLIP}; MedKLIP, a knowledge-enhanced radiology vision--language pretraining method that aligns medical entities with image patches~\cite{Wu2023MedKLIP}; the EVA-based encoder (EVA-Base in the tables); BioViL-T, a radiology-specific contrastive vision--language model from the BiomedVLP family that builds on the CXR-BERT text encoder and improved domain-specific textual semantics~\cite{10204115}; CheXFound, a self-supervised chest X-ray foundation model pretrained on a large curated CXR corpus and originally paired with a global--local representation integration strategy for downstream adaptation~\cite{Yang2025CheXFound}; and our proposed Med-AR-2B and Med-AR-8B encoders. Table~\ref{tab:encoder_ablation_multidataset_compact} summarizes the resulting performance across the three public benchmarks. Label-wise results are organized by dataset in Supplementary Sections~\ref{supp:internal}--\ref{supp:padchest}.

Table~\ref{tab:encoder_ablation_multidataset_compact} shows that the autoregressive encoders remain the strongest representation learners under a fixed downstream setup, and this conclusion now holds across all three public datasets. On CheXpert, Med-AR-8B achieves the best mAUROC and mAUPRC for the head, medium, and tail subsets, with MedCLIP consistently second-best and ARK/CheXFound close behind on the head subset. On MIMIC-CXR, Med-AR-8B is again best across all prevalence groups, while Med-AR-2B is second-best throughout, clearly ahead of the non-autoregressive baselines. On PadChest, Med-AR-2B is the strongest model across head, medium, and tail subsets, with Med-AR-8B second-best and ARK/ CheXFound leading among the non-autoregressive alternatives. Among the non-AR baselines, ARK and CheXFound are the most competitive on the head and tail groups, whereas the other radiology-specific and biomedical baselines, including MedKLIP, EVA-Base, and BioViL-T, remain clearly below the best Med-AR variants on both ranking-based metrics for every dataset.

\FloatBarrier

This comparison extends the analysis beyond a single contrastive baseline under a common downstream head. It is not a component ablation: the pretrained systems differ in architecture, pretraining data, and optimization. The results support transfer performance under the evaluated setup, without attributing differences solely to the pretraining objective or excluding interactions with the downstream head.
\subsection{Discussion}
\label{subsec:discussion}

Taken together, the results support five main observations.

First, on all three public benchmarks---PadChest, MIMIC-CXR, and CheXpert---autoregressive pretraining improves mAUROC and mAUPRC over Med-CLIP with Med-AR-8B across all prevalence groups, with nominal statistical support in most comparisons, with the strongest and most consistent gains delivered by Med-AR-8B. On MIMIC-CXR and CheXpert, Med-AR-8B is best overall and in every prevalence subgroup; on PadChest, the smaller Med-AR-2B is best while Med-AR-8B remains a strong second-best. The consistency of this pattern across aggregate metrics, paired Wilcoxon tests, and label-wise DeLong comparisons suggests a broad improvement in representation transfer rather than an artifact of a few abnormalities. The tail-label gains are consistent with the hypothesis that structured and region-aware autoregressive supervision benefits rare-finding recognition, but do not directly demonstrate the proposed visual-grounding mechanism.

Second, the internal benchmark is more nuanced. Med-AR-2B shows a modest mAUROC edge with partial statistical support, but mAUPRC results are mixed, tail-label separation is not uniform, and Med-CLIP retains a clear EAURC advantage in every subgroup. The two paradigms are therefore best described as broadly comparable on internal data, suggesting that the benefit of autoregressive pretraining is partly conditioned by dataset properties such as reporting consistency, label construction quality, and institutional heterogeneity.

Third, the EAURC analysis reveals that stronger discrimination does not guarantee better confidence-based risk ordering. Med-AR improves EAURC on all three public benchmarks, but CheXpert provides a striking example where discrimination and selective-prediction behavior decouple: Med-AR-2B has lower AUROC and AUPRC than Med-CLIP yet achieves the best EAURC across every CheXpert subgroup. On the internal dataset, the opposite dissociation occurs, with Med-CLIP achieving the best EAURC despite Med-AR matching or exceeding it on discrimination metrics. Together, these findings underscore the importance of evaluating selective-prediction behavior as a distinct axis rather than assuming it follows discrimination gains.

Fourth, the encoder comparison across CheXpert, MIMIC-CXR, and PadChest shows that Med-AR encoders outperform not only Med-CLIP but also a broad set of self-supervised, supervised, biomedical contrastive, knowledge-enhanced, and radiology-specific VLP baselines under the same ML-Decoder head, supporting the effectiveness of the complete pretraining recipe under this downstream setup.

Finally, the preferred autoregressive variant is dataset-dependent: Med-AR-2B leads on Internal and PadChest, while Med-AR-8B leads on MIMIC-CXR and CheXpert. This is most consequential on CheXpert, where Med-AR-2B fails to outperform Med-CLIP on discrimination while Med-AR-8B does so decisively. This suggests an interaction between decoder capacity, the pretraining recipe, and the downstream task. The present experiments do not isolate the objective as the primary driver of the gains.

Overall, these findings identify radiology-native autoregressive pretraining as a strong foundation for long-tailed CXR classification on public benchmarks, while the mixed internal results highlight that representation superiority is not absolute and depends on both dataset characteristics and the evaluation axis under consideration.

\subsection{Limitations}
The comparison controls the downstream architecture, but the pretraining recipes differ in supervision and may differ in initialization and compute. No matched-supervision component ablation is available, so the results cannot establish a causal advantage of next-token prediction alone. The additional pretrained-encoder comparison also does not equalize architecture or pretraining data.

The expanded MIMIC-CXR and CheXpert labels are derived from reports. Inter-model concordance does not establish expert-level correctness, and unmentioned findings may be encoded as negatives. Generated reasoning and prominence targets may introduce additional noise. The 1,000-report concordance sample provides limited evidence for very rare labels, and the retained PadChest vocabulary excludes many less frequent findings. Each dataset is fine-tuned separately; these experiments do not constitute deployment on an unseen institution without adaptation.

The statistical analyses do not quantify training-seed variability or establish equivalence for non-significant comparisons. Label dependencies and multiple testing warrant caution in interpreting nominal significance. EAURC assesses selective prediction under the evaluated protocol rather than calibration or clinical utility. Independent expert validation, fuller reporting of data provenance and split construction, and prospective evaluation remain necessary to assess clinical applicability.

\section{Conclusion and Future Work}

We investigated whether the choice of vision--language pretraining recipe---contrastive versus autoregressive---affects downstream transfer for long-tailed multi-label chest X-ray classification. To reduce downstream architectural confounding, we compared MedCLIP-style contrastive pretraining and InternVL3-based autoregressive pretraining (Med-AR-2B and Med-AR-8B) using a shared InternViT vision architecture and a common ML-Decoder classification head, and evaluated all models on an internal benchmark together with three public datasets: PadChest, MIMIC-CXR, and CheXpert.

Across the three public benchmarks, Med-AR-8B improves downstream AUROC and AUPRC over Med-CLIP in each prevalence group, with nominally significant paired per-label gains in most comparisons. Med-AR-8B performs best on MIMIC-CXR and CheXpert, while Med-AR-2B performs best on PadChest. The encoder comparison also favors Med-AR under the evaluated downstream setup. These observations support the complete autoregressive pretraining recipe, while the separate contributions of objective, supervision, and decoder scale remain unresolved.

On the internal benchmark, the two paradigms are better characterized as broadly comparable: Med-AR-2B shows a modest AUROC edge with partial statistical support, but AUPRC results are mixed and Med-CLIP retains a clear advantage in selective-prediction behavior as measured by EAURC. More broadly, our uncertainty-aware analysis uncovers a consequential decoupling between discrimination and confidence ranking. On CheXpert, Med-AR-2B achieves the best EAURC across all prevalence groups despite lower AUROC and AUPRC than Med-CLIP, while on the internal dataset the opposite dissociation occurs. Stronger discrimination therefore does not guarantee better confidence-based risk ordering, and prevalence-aware, uncertainty-aware evaluation should be treated as a first-class axis---rather than a secondary metric---when comparing medical vision--language pretraining strategies.

Several directions remain open. First, disentangling the contribution of the autoregressive objective itself from its richer supervision---structured multi-section report templates, abnormality-focused auxiliary outputs, and region-level annotations---will require component-wise ablations that we leave to future work. Second, dedicated strategies for rare-label learning, such as curriculum scheduling, meta-learning, and focused autoregressive pretraining on curated rare-finding corpora, may further close the gap on tail abnormalities and reduce residual variability across decoder scales. Third, integrating calibration techniques such as temperature scaling~\cite{Kull2019-wn,Xie2024-ii}, conformal prediction~\cite{Karimi2024-ev,Angelopoulos2020-cc}, and post-hoc meta-models~\cite{Shen2023-fp} would complement EAURC and support further assessment of reliability before clinical deployment. Finally, extending the LLM-based label-expansion framework to additional datasets and imaging modalities, and embedding the resulting confidence estimates into clinical workflows such as triage and radiologist--AI collaboration, are important next steps toward assessing real-world impact.

In summary, radiology-native autoregressive pretraining provides a strong foundation for long-tailed chest X-ray classification on public benchmarks, particularly for underrepresented findings. Its advantages, however, are not uniform across datasets or evaluation axes, and the Med-AR configuration that is best for discrimination need not coincide with the one that is best for selective prediction. These observations argue for a joint assessment of discrimination and uncertainty behavior when comparing medical vision--language pretraining paradigms, and position radiology-native Med-AR as a starting point for further research on reliable chest X-ray classification.

\section{Statements and Declarations}
\begin{credits}
\subsubsection{Author Contributions}
All authors contributed to the conception and design of the study. 
Janhavi Prabhu led the model training, methodology development, experimental design, and manuscript writing. 
Sahil contributed to model training, dataset preparation, label extraction and manuscript writing. 
Akshay Valsaraj contributed to implementation support and model development. 
Manoj Tadepalli and Preetham Putha provided oversight, technical guidance, and critical revisions of the manuscript. 
All authors reviewed and approved the final version of the manuscript.
\subsubsection{Data Availability} We plan to release the derived MIMIC-CXR and CheXpert label annotations and split identifiers following publication, subject to the source datasets' access and licensing requirements. Original images and reports remain subject to their respective access conditions.

\subsubsection{\discintname} The authors have no competing interests to declare that are relevant to the content of this article.
\end{credits}
\bibliographystyle{splncs04}
\bibliography{citations}
\clearpage
\newgeometry{left=18mm,right=18mm,top=20mm,bottom=20mm}
\pagestyle{plain}
\captionsetup[longtable]{width=\linewidth}
\setcounter{section}{0}
\setcounter{subsection}{0}
\setcounter{figure}{0}
\setcounter{table}{0}
\setcounter{equation}{0}
\renewcommand{\thesection}{S\arabic{section}}
\renewcommand{\thesubsection}{S\arabic{section}.\arabic{subsection}}
\renewcommand{\thefigure}{S\arabic{figure}}
\renewcommand{\thetable}{S\arabic{table}}
\renewcommand{\theequation}{S\arabic{equation}}
\renewcommand{\theHsection}{supp.\arabic{section}}
\renewcommand{\theHfigure}{supp.\arabic{figure}}
\renewcommand{\theHtable}{supp.\arabic{table}}
\renewcommand{\theHequation}{supp.\arabic{equation}}
\begin{center}
{\Large\bfseries Supplementary Material}\\[0.5em]
{\bfseries Med-AR: Autoregressive Vision-Language Pretraining for Long-Tailed Chest X-Ray Classification and Uncertainty-Aware Evaluation}
\end{center}
This supplement provides label-construction details and the existing per-label results. Results are grouped by dataset so that discrimination and operating-point metrics can be inspected together. In the tables, $*$ denotes tail labels and $\dagger$ denotes medium-frequency labels; unmarked labels are head labels. Bold and underlined entries indicate the highest and second-highest distinct displayed values, respectively, with ties sharing rank. These marks are descriptive, not significance tests.

\begin{center}
\begin{tabular}{ll}
\toprule
Section & Contents\\
\midrule
\ref{supp:methods} & Target preparation and label concordance\\
\ref{supp:overview} & Cross-dataset summary and operating points\\
\ref{supp:internal} & Internal: consolidated per-label metrics\\
\ref{supp:chexpert} & CheXpert: all five per-label metrics\\
\ref{supp:mimic} & MIMIC-CXR: all five per-label metrics\\
\ref{supp:padchest} & PadChest: all five per-label metrics\\
\bottomrule
\end{tabular}
\end{center}

\section{Target Preparation and Label Concordance}\label{supp:methods}
\subsection{Pretraining Target Structure}
The autoregressive targets use 15 anatomical subheadings with finding-specific fields, together with Findings, Impression, Summary, Reasoning, and Prominence Score sections. Prominence scores range from 0 to 5 and are generated from report text. Qwen3 is used to construct these report-derived targets; expert bounding-box annotations provide separate region-level supervision for approximately 20 findings. Non-mention is mapped to negation in the reported template construction, which can introduce false-negative supervision. The reasoning and prominence fields are synthetic targets, not independent expert image annotations. This description summarizes the target structure and does not specify an exact reproducible prompt.

\subsection{LLM-Based Label Expansion Pipeline}
\label{sec:llm_label_expansion}

\paragraph{Parsing and construction of the presence matrix.}
Each LLM response was parsed from its hierarchical JSON representation into tuples of
\texttt{(parent group, child tag, presence)}. Although the extraction schema additionally permitted attributes such as size, location, and characteristics, only the binary \texttt{presence} field was used for label construction and concordance analysis. Child labels were indexed jointly with their parent group to avoid collisions between identically named labels occurring in different sections of the schema.

For a successfully parsed report, a tag explicitly marked as present was assigned a value of 1. Tags belonging to the model vocabulary but not emitted for that report were treated as implicitly absent (0). In contrast, reports for which the JSON response could not be parsed were treated as missing rather than negative and were excluded from the corresponding comparisons. This distinction prevented parsing failures from being interpreted as evidence for absence. A zero denotes the extraction convention and does not independently establish image-level absence.

\paragraph{Vocabulary normalization.}
The predefined schema was supplemented with an open-ended \texttt{other\_findings} field to preserve findings not anticipated by the original taxonomy. Consequently, the raw outputs did not share a perfectly fixed vocabulary. In the full concordance analysis, the seven models collectively produced 5,586 distinct keys, whereas only 193 canonical tags were emitted by all seven models; most of the remaining variation originated from \texttt{other\_findings}. This expansion primarily reflected lexical fragmentation--for example, singular/plural forms such as \texttt{chest\_tube} and \texttt{chest\_tubes}--rather than thousands of distinct radiographic concepts.

We therefore normalized the extracted vocabulary before constructing the final label space. Formatting variants and alternative representations of the same concept were first standardized, after which semantically equivalent labels were mapped to a canonical label. Importantly, normalization was performed at the level of clinical concepts rather than by string similarity alone. Labels were therefore merged only when they represented the same finding or a difference in naming granularity, while clinically related but distinct findings remained separate. Mapping to a parent group for concordance analysis should not be interpreted as evidence that the child concepts are interchangeable.

\paragraph{Synonym and hierarchical mapping.}
In the absence of ground-truth annotations for the expanded label set, concordance across multiple LLMs was used as a check of inter-model reproducibility. Agreement among models was interpreted as supporting evidence that the extracted labels were reproducible rather than as a mechanism for defining the label mappings themselves. The concordance analysis also provided insight into the nature of residual disagreement. In particular, many apparent discrepancies occurred between synonymous or hierarchically related labels. For example, when \texttt{solitary\_pulmonary\_nodule} was not emitted, \texttt{nodules} was emitted in 96\% of such cases; similarly, \texttt{pleural\_effusion} was emitted in 95\% of cases in which \texttt{effusion} was absent. These observations indicate that a substantial fraction of tag-level disagreement reflected differences in terminology or label granularity rather than failure to identify the underlying radiographic finding, and informed inspection of the normalization and mapping procedure without establishing clinical correctness.

\begin{figure}
    \centering
    \includegraphics[width=\linewidth]{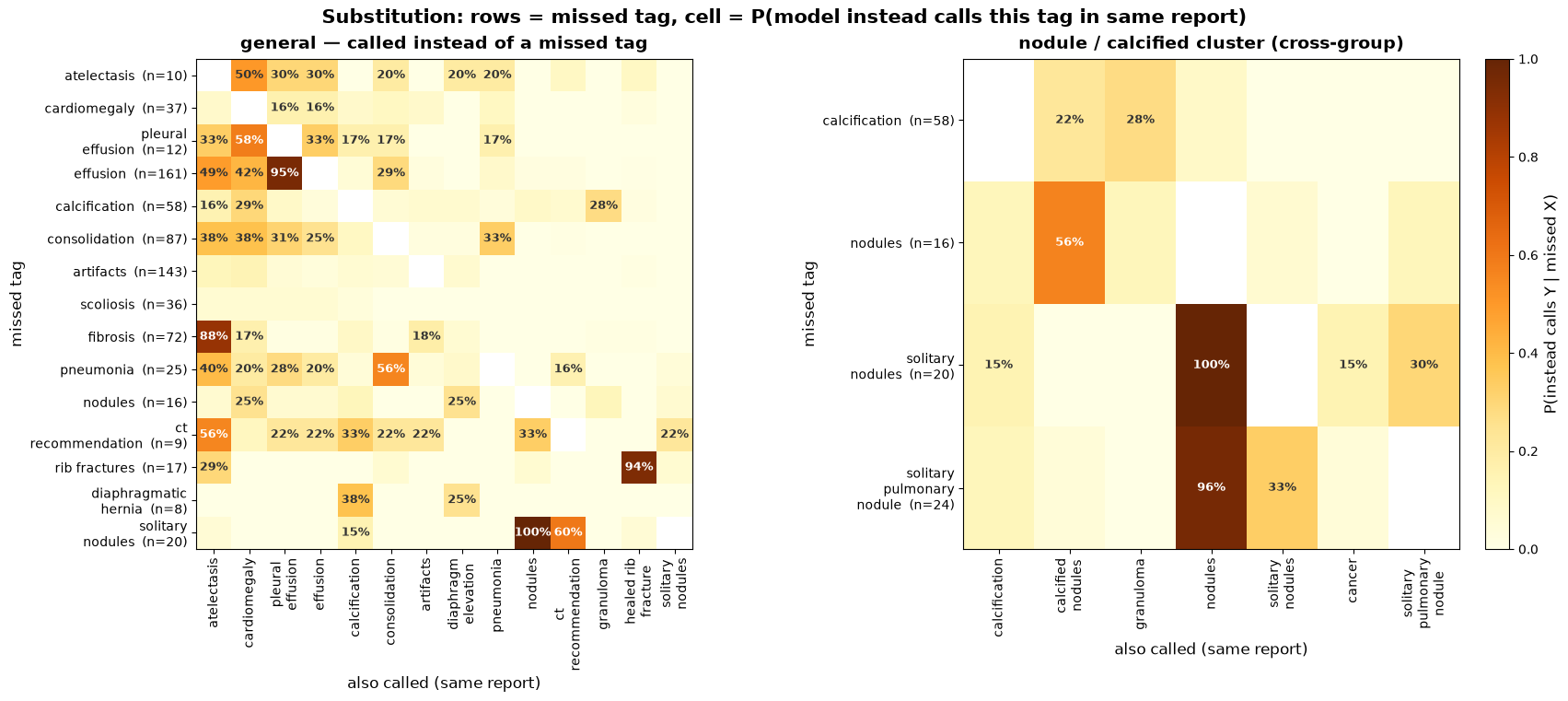}
    \caption{\textbf{Substitution patterns among discordant labels.}
For each missed tag (rows), cells show the probability that the same model assigned an alternative tag (columns) in the same report. High substitution rates identify cases where apparent disagreement reflects alternative label routing rather than failure to detect the underlying finding, as observed for \texttt{effusion}/\texttt{pleural\_effusion} and the nodule family.}
    \label{fig:substitution_concordance}
\end{figure}

We consequently examined related label families for normalization and parent-level concordance analysis. These families include both synonyms and related but non-equivalent findings; parent-level agreement does not establish equivalence at the child-label level. These included the nodule family; \texttt{effusion}/\texttt{pleural\_effusion}; low-lung-volume variants; COPD-related terminology such as hyperinflation and emphysema; ILD/fibrotic terminology; and closely related edema terminology. The mapping was deliberately conservative. In particular, \texttt{nodule}, \texttt{mass}, and \texttt{opacity} were maintained as separate concepts despite their potential co-occurrence, since substitutions among these categories can represent genuine differences in radiographic interpretation rather than lexical variation. Similarly, the airspace-opacity family (e.g., consolidation, pneumonia, perihilar opacity, and edema) was not indiscriminately collapsed, as the concordance analysis identified this group as one of the remaining sources of substantive clinical disagreement.

\begin{figure}
    \centering
    \includegraphics[width=\linewidth]{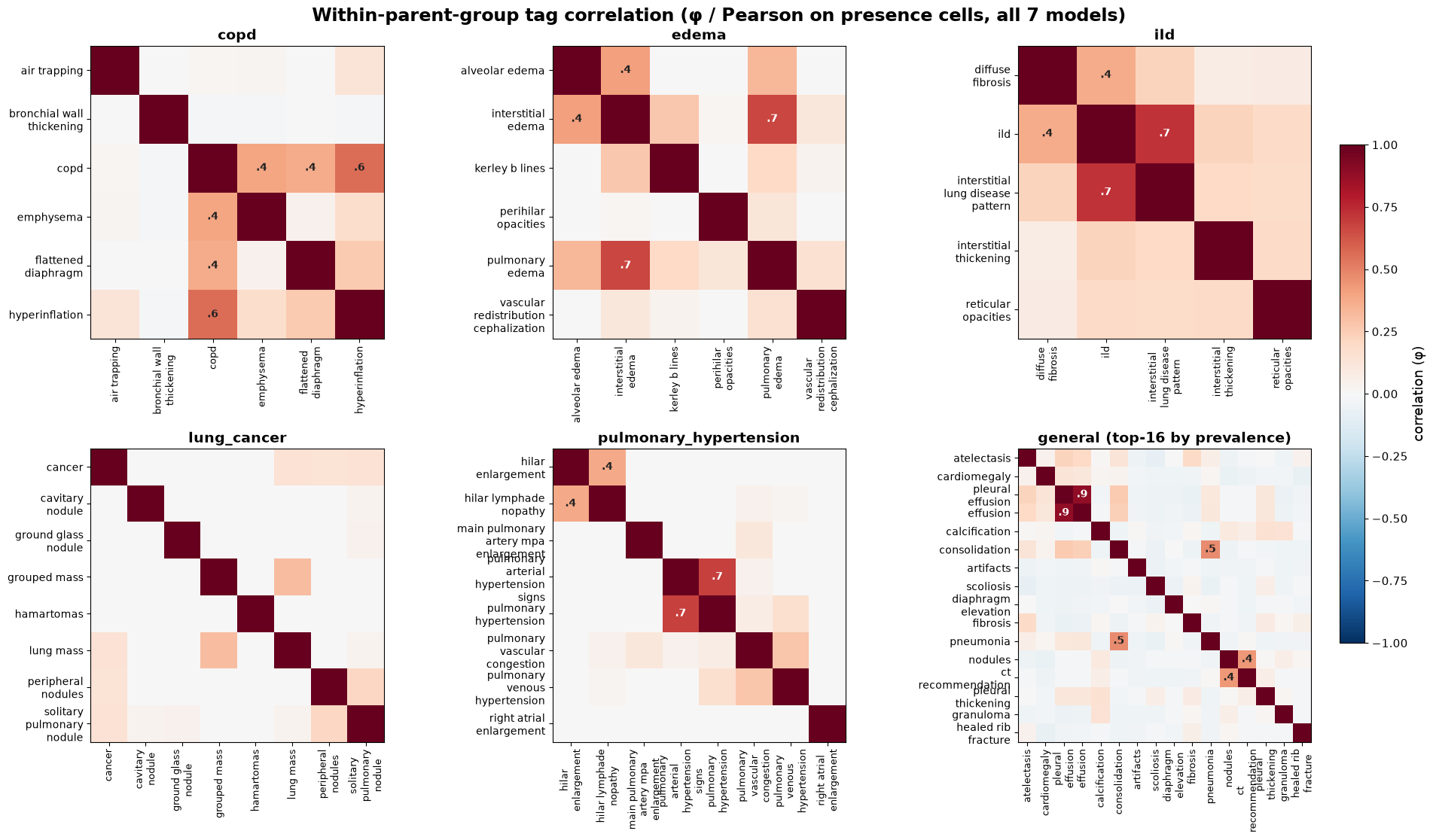}
    \caption{\textbf{Within-group correlation of extracted labels across the seven LLMs.}
Pairwise $\phi$ correlations between binary tag-presence indicators were computed across pooled model--report observations within each parent group. Higher $\phi$ indicates greater co-occurrence, capturing related, overlapping, or potentially synonymous labels.}
    \label{fig:parent_tag_corr_concordance}
\end{figure}

The effect of synonym routing was also assessed by comparing agreement at different levels of the hierarchy. Collapsing child labels into their parent finding groups increased mean pairwise Cohen's $\kappa$ from 0.83 to 0.87, indicating that a substantial component of child-level disagreement disappeared when differences in terminology and granularity were removed. Thus, parent-level agreement was used as an additional diagnostic for distinguishing genuine disagreement from alternative routing within the taxonomy.

\begin{figure}
    \centering
    \includegraphics[width=\linewidth]{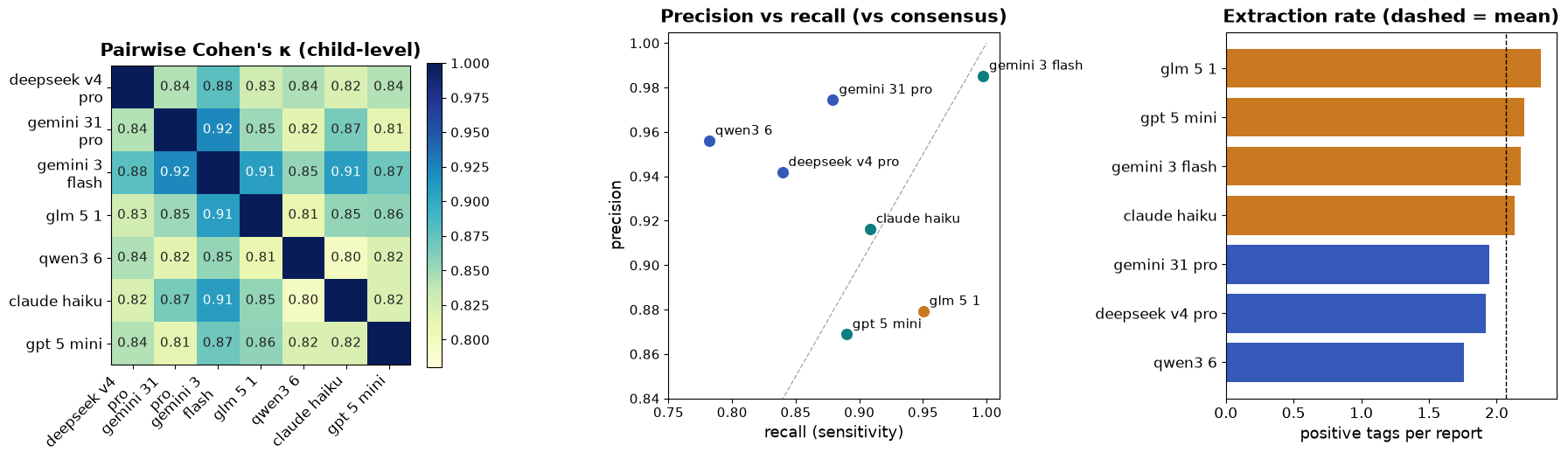}
    \caption{\textbf{Inter-model agreement and extraction characteristics across seven LLMs.}
\textbf{Left:} Pairwise Cohen's $\kappa$ at the child-label level, demonstrating consistently high agreement between models.
\textbf{Middle:} Precision--recall characteristics relative to the seven-model majority consensus, illustrating differences in model extraction behavior from conservative (higher precision, lower recall) to more inclusive extraction.
\textbf{Right:} Mean number of positive tags extracted per report for each model; the dashed line indicates the seven-model mean.}
    \label{fig:cohens_kappa}
\end{figure}

\paragraph{Handling free-form and ambiguous labels.}
Free-form \texttt{other\_findings} labels received additional scrutiny because their vocabulary was substantially less standardized. Device and incidental findings were particularly affected: labels such as central venous catheters, PICC lines, chest tubes, surgical clips, and degenerative changes exhibited low apparent tag-level agreement despite frequently representing differences in naming or schema coverage. Free-form keys were therefore not automatically promoted to independent classes. Instead, recurrent keys were mapped to an existing canonical concept where possible; lexical variants were consolidated; and poorly supported or non-reproducible keys were flagged for taxonomy review.

The \texttt{artifacts} label was handled separately because its disagreement pattern differed from synonym routing. Only 92 reports reached majority consensus for \texttt{artifacts}, although at least one model emitted the label in 228 reports. Moreover, models emitting \texttt{artifacts} generally also emitted the corresponding radiographic abnormalities, indicating that \texttt{artifacts} behaved as an additional, model-dependent catch-all rather than as an alternative name for another finding. Such labels were therefore not merged with co-occurring clinical findings and were instead treated as candidates for stricter definition or removal.

\begin{figure}
    \centering
    \includegraphics[width=\linewidth]{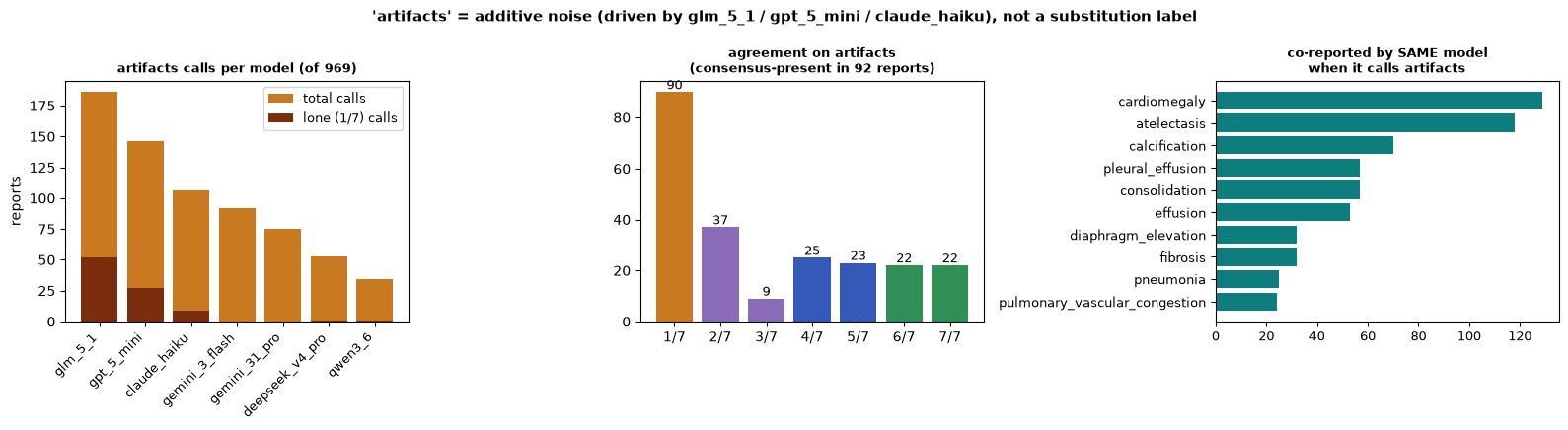}
    \caption{\textbf{Model-specific behavior of the \texttt{artifacts} label.}
The \texttt{artifacts} label shows substantial model-specific variation and is particularly frequent for GLM~5.1. This reflects a model-specific labeling tendency in which GLM~5.1 often assigns the generic \texttt{artifacts} label alongside more specific findings, such as devices or tubes. The high frequency of single-model calls consequently suggests that \texttt{artifacts} represents an inconsistently interpreted label rather than systematic disagreement in the underlying radiographic findings.}
    \label{fig:artifacts_concordance}
\end{figure}

\paragraph{Taxonomy auditing and filtering.}
Following normalization, tags were audited using several complementary indicators of instability. These included the number of models containing the tag in their vocabulary, prevalence across models, pairwise agreement and Cohen's $\kappa$, frequency of single-model positive calls, and the extent to which disagreement disappeared at the parent-group level. Tags occurring in only a minority of model vocabularies were interpreted primarily as schema-coverage differences rather than reliable independent categories.

Finally, majority voting across the seven models was used during concordance analysis to identify systematic routing and extraction differences. A report-tag pair was considered consensus-positive when at least four of seven models marked it as present. This consensus served as an analytical reference rather than ground truth: over-calls and under-calls were defined relative to the $\geq4/7$ majority and were subsequently examined for systematic substitutions. This procedure was particularly useful for separating consensus-relative omissions from vocabulary routing; for example, the apparently frequent under-calling of \texttt{effusion} was largely attributable to models using \texttt{pleural\_effusion} instead. Overall, the post-processing procedure was designed to reduce vocabulary fragmentation without artificially increasing agreement by collapsing clinically distinct findings.

\FloatBarrier
\clearpage
\section{Cross-Dataset Summary and Operating Points}\label{supp:overview}
\subsection{Operating-Point Metrics}\label{supp:operating_points}
For public-dataset sensitivity, specificity, and Youden's $J$, thresholds are selected per label by maximizing $J=\mathrm{Sensitivity}+\mathrm{Specificity}-1$ on the validation set. These operating points summarize the sensitivity--specificity trade-off, rather than a clinically validated deployment threshold. The internal per-label table is described in the source analysis as validation-set performance with class-wise $J$-maximizing thresholds; it should be distinguished from the held-out aggregate comparison in the main text.

\subsection{Top-2 Win Rate Analysis for Public Datasets}
\label{subsec:top2_win_rate}

To provide a holistic view of model competitiveness across the entire label
space of chest radiograph benchmarks, we summarize the per-tag performance
of every method using a \emph{Top-2 Win Rate} metric. For each tag in a given
dataset, we rank all ten competing encoders by their score and award a
``win'' to any model that appears among the top two. The win rate of a
model is then defined as the fraction of tags (expressed as a percentage)
on which it achieves a top-2 placement. This formulation rewards models
that are \textit{consistently} strong across the full tag distribution,
rather than those that excel on only a handful of well-represented findings.

\begin{figure}[t]
    \centering
    \includegraphics[width=0.95\linewidth]{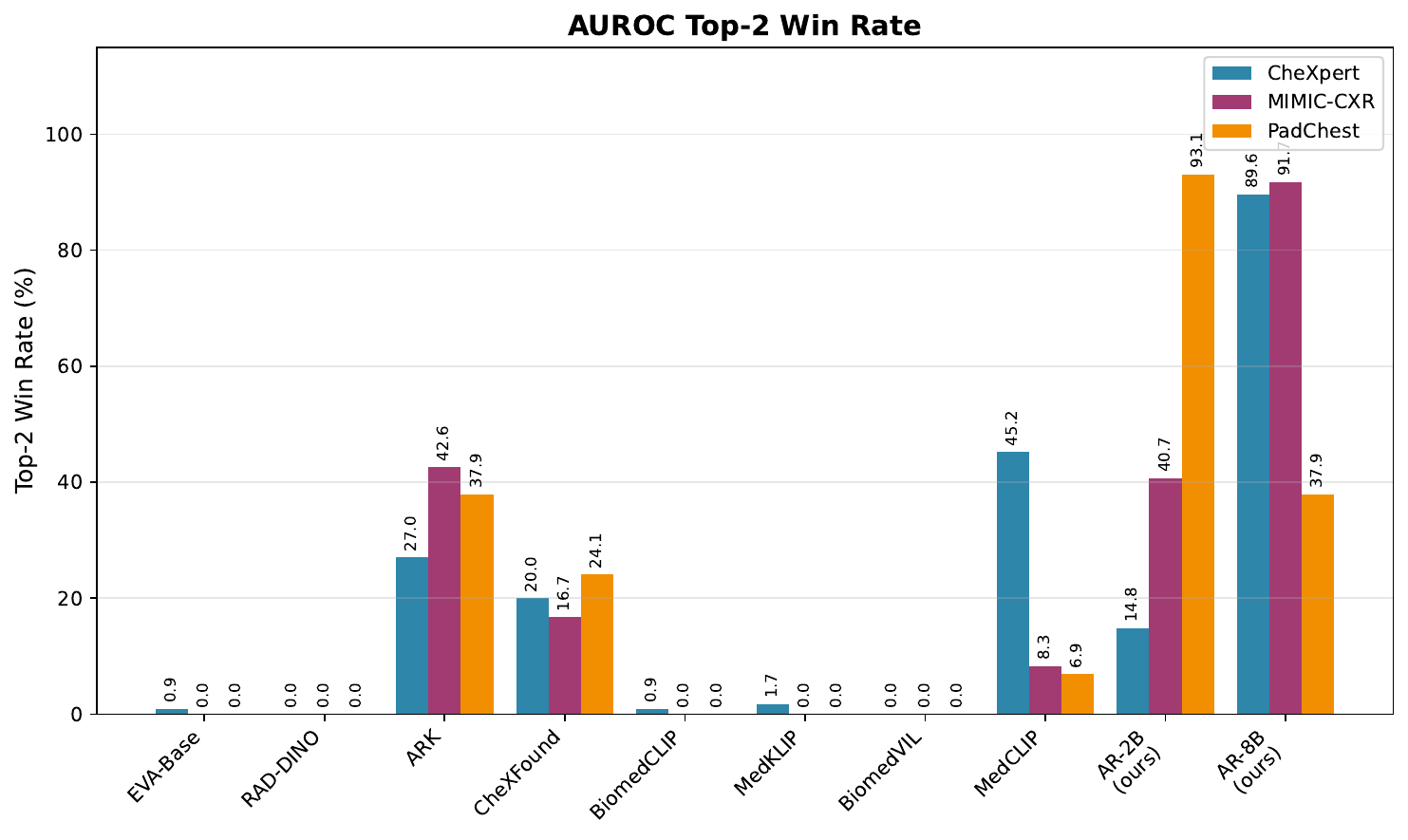}
    \caption{Top-2 Win Rate (\%) under AUROC across CheXpert, MIMIC-CXR,
    and PadChest. Our Med-AR-2B and Med-AR-8B models achieve top-2 placements on a
    larger fraction of tags than prior encoders. Numeric
    values are annotated above each bar for clarity.}
    \label{fig:win_rate_auroc}
\end{figure}

\begin{figure}[t]
    \centering
    \includegraphics[width=0.95\linewidth]{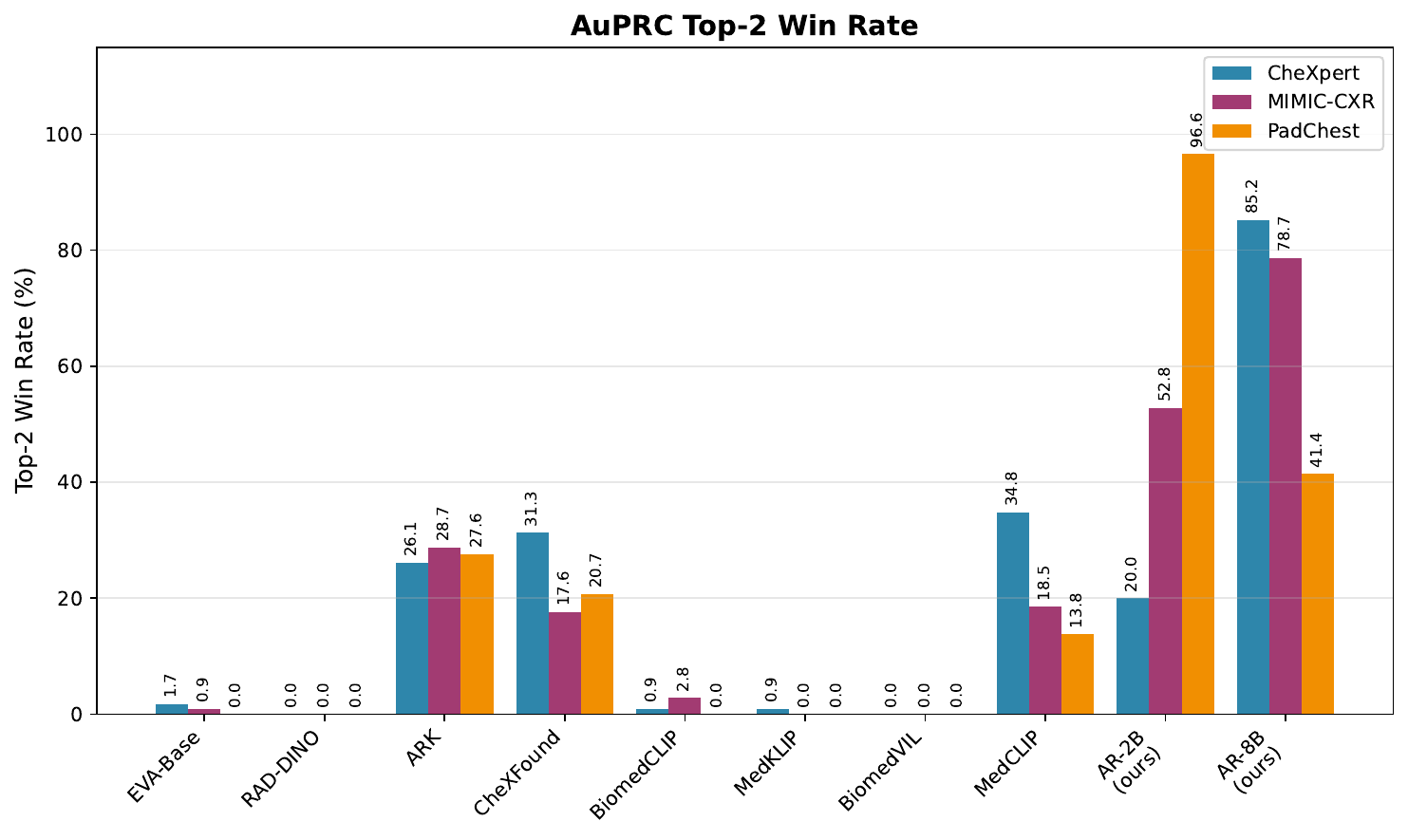}
    \caption{Top-2 Win Rate (\%) under AUPRC across CheXpert, MIMIC-CXR,
    and PadChest. Consistent with the AUROC results, our proposed models
    lead on precision-recall performance, with Med-AR-2B particularly
    excelling on the long-tailed PadChest benchmark.}
    \label{fig:win_rate_auprc}
\end{figure}

Figure~\ref{fig:win_rate_auroc} reports the Top-2 Win Rate measured under
AUROC, while Figure~\ref{fig:win_rate_auprc} reports the same analysis
under AUPRC. We evaluate on three widely used chest X-ray
benchmarks CheXpert, MIMIC-CXR, and PadChest covering a broad spectrum
of pathology labels ranging from common findings to long-tailed diagnoses.
Med-AR-8B has the highest reported top-2 rates on CheXpert (89.6\% AUROC, 85.2\% AUPRC) and MIMIC-CXR (91.7\% AUROC, 78.7\% AUPRC), while Med-AR-2B leads on PadChest (93.1\% AUROC, 96.6\% AUPRC). This summary describes relative placement across labels; it does not measure effect size or establish rare-label clinical performance.

Per-label values are provided by dataset in Sections~\ref{supp:chexpert}--\ref{supp:padchest}.

\FloatBarrier
\clearpage
\begin{landscape}
\section{Internal Dataset: Per-Label Results}\label{supp:internal}
The following table preserves the reported validation-set results for 40 labels. AUROC and AUPRC describe discrimination; sensitivity, specificity, and Youden's $J$ describe class-wise operating points. Performance varies across labels and metrics, with neither pretraining recipe uniformly superior.

\fontsize{6.3}{7.5}\selectfont
\setlength{\tabcolsep}{2.5pt}
\renewcommand{\arraystretch}{1.1}
\setlength\LTcapwidth{\linewidth}
\setlength\LTleft{0pt}
\setlength\LTright{0pt}

\begin{longtable}{l|ccc|ccc|ccc|ccc|ccc}
\caption{Consolidated per-label performance on the internal dataset across AUROC, AUPRC, Sensitivity, Specificity, and Youden's $J$ for Med-CLIP and our autoregressive models (AR-2B, AR-8B). Thresholds for Sensitivity/Specificity/Youden are obtained per abnormality by maximising $J = \mathrm{Sens} + \mathrm{Spec} - 1$. Tail abnormalities are marked with an asterisk (*), and medium-prevalence abnormalities with a dagger ($\dagger$). For each metric, the best value per row is shown in bold and the second-best is underlined.}
\label{tab:internal_full_metrics}\\
\toprule
\multirow{2}{*}{\textbf{Abnormality}}
 & \multicolumn{3}{c|}{\textbf{AUROC}}
 & \multicolumn{3}{c|}{\textbf{AUPRC}}
 & \multicolumn{3}{c|}{\textbf{Sensitivity}}
 & \multicolumn{3}{c|}{\textbf{Specificity}}
 & \multicolumn{3}{c}{\textbf{Youden's $J$}} \\
\cmidrule(lr){2-4}\cmidrule(lr){5-7}\cmidrule(lr){8-10}\cmidrule(lr){11-13}\cmidrule(lr){14-16}
 & \textbf{Med-CLIP} & \makecell[c]{\textbf{AR-2B}\\[-0.4ex]\textbf{(ours)}} & \makecell[c]{\textbf{AR-8B}\\[-0.4ex]\textbf{(ours)}}
 & \textbf{Med-CLIP} & \makecell[c]{\textbf{AR-2B}\\[-0.4ex]\textbf{(ours)}} & \makecell[c]{\textbf{AR-8B}\\[-0.4ex]\textbf{(ours)}}
 & \textbf{Med-CLIP} & \makecell[c]{\textbf{AR-2B}\\[-0.4ex]\textbf{(ours)}} & \makecell[c]{\textbf{AR-8B}\\[-0.4ex]\textbf{(ours)}}
 & \textbf{Med-CLIP} & \makecell[c]{\textbf{AR-2B}\\[-0.4ex]\textbf{(ours)}} & \makecell[c]{\textbf{AR-8B}\\[-0.4ex]\textbf{(ours)}}
 & \textbf{Med-CLIP} & \makecell[c]{\textbf{AR-2B}\\[-0.4ex]\textbf{(ours)}} & \makecell[c]{\textbf{AR-8B}\\[-0.4ex]\textbf{(ours)}} \\
\midrule
\endfirsthead

\toprule
\multirow{2}{*}{\textbf{Abnormality}}
 & \multicolumn{3}{c|}{\textbf{AUROC}}
 & \multicolumn{3}{c|}{\textbf{AUPRC}}
 & \multicolumn{3}{c|}{\textbf{Sensitivity}}
 & \multicolumn{3}{c|}{\textbf{Specificity}}
 & \multicolumn{3}{c}{\textbf{Youden's $J$}} \\
\cmidrule(lr){2-4}\cmidrule(lr){5-7}\cmidrule(lr){8-10}\cmidrule(lr){11-13}\cmidrule(lr){14-16}
 & \textbf{Med-CLIP} & \makecell[c]{\textbf{AR-2B}\\[-0.4ex]\textbf{(ours)}} & \makecell[c]{\textbf{AR-8B}\\[-0.4ex]\textbf{(ours)}}
 & \textbf{Med-CLIP} & \makecell[c]{\textbf{AR-2B}\\[-0.4ex]\textbf{(ours)}} & \makecell[c]{\textbf{AR-8B}\\[-0.4ex]\textbf{(ours)}}
 & \textbf{Med-CLIP} & \makecell[c]{\textbf{AR-2B}\\[-0.4ex]\textbf{(ours)}} & \makecell[c]{\textbf{AR-8B}\\[-0.4ex]\textbf{(ours)}}
 & \textbf{Med-CLIP} & \makecell[c]{\textbf{AR-2B}\\[-0.4ex]\textbf{(ours)}} & \makecell[c]{\textbf{AR-8B}\\[-0.4ex]\textbf{(ours)}}
 & \textbf{Med-CLIP} & \makecell[c]{\textbf{AR-2B}\\[-0.4ex]\textbf{(ours)}} & \makecell[c]{\textbf{AR-8B}\\[-0.4ex]\textbf{(ours)}} \\
\midrule
\endhead

\midrule
\multicolumn{16}{r}{\scriptsize Continued on next page} \\
\endfoot

\bottomrule
\endlastfoot
Hydropneumothorax$^{*}$ & \textbf{0.9899} & \underline{0.9888} & 0.9863 & \textbf{0.4445} & \underline{0.3672} & 0.3013 & \textbf{0.9641} & \underline{0.9596} & 0.9417 & \underline{0.9634} & 0.9623 & \textbf{0.9713} & \textbf{0.9275} & \underline{0.9219} & 0.9130 \\
Hilar Mass$^{*}$ & \textbf{0.8907} & \underline{0.8904} & 0.8764 & \textbf{0.3714} & 0.3275 & \underline{0.3671} & \textbf{0.9107} & \textbf{0.9107} & \underline{0.8571} & \underline{0.7424} & 0.7285 & \textbf{0.7850} & \textbf{0.6531} & 0.6392 & \underline{0.6421} \\
Granuloma$^{\dagger}$ & 0.9185 & \textbf{0.9481} & \underline{0.9367} & 0.0782 & \textbf{0.0863} & \underline{0.0804} & 0.8346 & \underline{0.8508} & \textbf{0.8612} & 0.8395 & \textbf{0.9091} & \underline{0.8815} & 0.6741 & \textbf{0.7599} & \underline{0.7427} \\
Consolidation & 0.8687 & \textbf{0.8902} & \underline{0.8877} & 0.5193 & \textbf{0.5746} & \underline{0.5698} & 0.7924 & \underline{0.8128} & \textbf{0.8142} & 0.7824 & \textbf{0.8091} & \underline{0.8026} & 0.5748 & \textbf{0.6219} & \underline{0.6168} \\
Lesion$^{\dagger}$ & \textbf{0.5796} & 0.5431 & \underline{0.5764} & \underline{0.0792} & 0.0744 & \textbf{0.0810} & \underline{0.6790} & \textbf{0.7433} & 0.6639 & \textbf{0.4683} & 0.3474 & \underline{0.4596} & \textbf{0.1473} & 0.0907 & \underline{0.1235} \\
Fibrosis & 0.8588 & \textbf{0.8866} & \underline{0.8821} & 0.3234 & \textbf{0.3986} & \underline{0.3849} & 0.7578 & \textbf{0.7951} & \underline{0.7865} & 0.7937 & \textbf{0.8168} & \underline{0.8131} & 0.5515 & \textbf{0.6119} & \underline{0.5996} \\
Scoliosis$^{\dagger}$ & 0.8772 & \textbf{0.9162} & \underline{0.9107} & 0.2814 & \textbf{0.3580} & \underline{0.3536} & 0.7769 & \textbf{0.8056} & \underline{0.7859} & 0.8583 & \underline{0.8867} & \textbf{0.9065} & 0.6352 & \underline{0.6923} & \textbf{0.6924} \\
\end{longtable}

\end{landscape}
\clearpage
\begin{landscape}
\section{CheXpert: Per-Label Results}\label{supp:chexpert}
The following tables report the existing results for 115 labels, in the order AUROC, AUPRC, sensitivity, specificity, and Youden's $J$.

\subsection{AUROC}

\scriptsize
\setlength{\tabcolsep}{3pt}
\renewcommand{\arraystretch}{1.08}
\setlength\LTcapwidth{\linewidth}
\setlength\LTleft{0pt}
\setlength\LTright{0pt}

\begin{longtable}{l|cccccccccc}
\caption{Per-label AUROC comparison across baseline, contrastive, and autoregressive models on the CheXpert dataset. Tail abnormalities are marked with an asterisk (*), and medium-prevalence abnormalities are marked with a dagger ($\dagger$). For each row, the best value is shown in bold and the second-best value is underlined.}
\label{tab:per_label_auroc_all_models}\\

\toprule
\multirow{2}{*}{\textbf{Abnormality}} & \textbf{EVA-Base} & \textbf{RAD-DINO} & \textbf{ARK} & \textbf{CheXFound} & \textbf{BiomedCLIP} & \textbf{MedKLIP} & \textbf{BioViL-T} & \textbf{Med-CLIP} & \makecell[c]{\textbf{Med-AR-2B}\\[-0.4ex]\textbf{(ours)}} & \makecell[c]{\textbf{Med-AR-8B}\\[-0.4ex]\textbf{(ours)}} \\
\midrule
\endfirsthead

\toprule
\multirow{2}{*}{\textbf{Abnormality}} & \textbf{EVA-Base} & \textbf{RAD-DINO} & \textbf{ARK} & \textbf{CheXFound} & \textbf{BiomedCLIP} & \textbf{MedKLIP} & \textbf{BioViL-T} & \textbf{Med-CLIP} & \makecell[c]{\textbf{Med-AR-2B}\\[-0.4ex]\textbf{(ours)}} & \makecell[c]{\textbf{Med-AR-8B}\\[-0.4ex]\textbf{(ours)}} \\
\midrule
\endhead

\midrule
\multicolumn{11}{r}{\scriptsize Continued on next page} \\
\endfoot

\bottomrule
\endlastfoot
LVAD* & 0.9166 & 0.9202 & 0.9499 & \underline{0.9741} & 0.8624 & 0.9086 & 0.6904 & 0.9581 & 0.9579 & \textbf{0.9893} \\
Diaphragm Elevation & 0.7963 & 0.7187 & 0.8422 & 0.8996 & 0.7299 & 0.7362 & 0.5589 & \underline{0.8998} & 0.8808 & \textbf{0.9134} \\
Osteopenia$^\dagger$ & 0.9038 & 0.8549 & 0.8978 & 0.9006 & 0.8195 & 0.8803 & 0.6719 & \underline{0.9102} & 0.9052 & \textbf{0.9130} \\
Parenchymal Opacities$^\dagger$ & 0.7188 & 0.7463 & 0.7707 & 0.7331 & 0.6791 & \textbf{0.7871} & 0.7458 & \underline{0.7794} & 0.7440 & 0.7681 \\
Nodules & 0.7665 & 0.7578 & \textbf{0.8370} & 0.8083 & 0.7394 & 0.7590 & 0.7210 & 0.8102 & 0.7904 & \underline{0.8347} \\
Pleural Effusion & 0.8818 & 0.8527 & \underline{0.9083} & \textbf{0.9089} & 0.8191 & 0.8817 & 0.7735 & 0.9053 & 0.8921 & 0.9075 \\
Congestive Heart Failure* & 0.7978 & 0.7765 & 0.8512 & 0.8580 & 0.7945 & 0.8033 & 0.5773 & 0.8646 & \textbf{0.8707} & \underline{0.8667} \\
Prosthetic Valve* & 0.8379 & 0.8322 & 0.8643 & \underline{0.8677} & 0.7147 & 0.8424 & 0.5762 & 0.8577 & 0.8327 & \textbf{0.8779} \\
\end{longtable}
\end{landscape}

\begin{landscape}
\subsection{AUPRC}

\scriptsize
\setlength{\tabcolsep}{3pt}
\renewcommand{\arraystretch}{1.08}
\setlength\LTcapwidth{\linewidth}
\setlength\LTleft{0pt}
\setlength\LTright{0pt}

\begin{longtable}{l|cccccccccc}
\caption{Per-label AUPRC comparison across baseline, contrastive, and autoregressive models on the CheXpert dataset. Tail abnormalities are marked with an asterisk (*), and medium-prevalence abnormalities are marked with a dagger ($^\dagger$). For each row, the best value is shown in bold and the second-best value is underlined.}
\label{tab:per_label_auprc_all_models}\\

\toprule
\textbf{Abnormality} & \textbf{EVA-Base} & \textbf{RAD-DINO} & \textbf{ARK} & \textbf{CheXFound} & \textbf{BiomedCLIP} & \textbf{MedKLIP} & \textbf{BioViL-T} & \textbf{Med-CLIP} & \makecell[c]{\textbf{Med-AR-2B}\\[-0.4ex]\textbf{(ours)}} & \makecell[c]{\textbf{Med-AR-8B}\\[-0.4ex]\textbf{(ours)}} \\
\midrule
\endfirsthead

\toprule
\textbf{Abnormality} & \textbf{EVA-Base} & \textbf{RAD-DINO} & \textbf{ARK} & \textbf{CheXFound} & \textbf{BiomedCLIP} & \textbf{MedKLIP} & \textbf{BioViL-T} & \textbf{Med-CLIP} & \makecell[c]{\textbf{Med-AR-2B}\\[-0.4ex]\textbf{(ours)}} & \makecell[c]{\textbf{Med-AR-8B}\\[-0.4ex]\textbf{(ours)}} \\
\midrule
\endhead

\midrule
\multicolumn{11}{r}{\scriptsize Continued on next page} \\
\endfoot

\bottomrule
\endlastfoot
LVAD* & 0.0814 & 0.0851 & 0.1213 & \underline{0.2439} & 0.0848 & 0.0591 & 0.0093 & 0.1771 & 0.2239 & \textbf{0.3616} \\
Diaphragm Elevation & 0.1911 & 0.1063 & 0.2607 & 0.4024 & 0.1548 & 0.1251 & 0.0426 & \underline{0.4165} & 0.4013 & \textbf{0.4314} \\
Osteopenia$^\dagger$ & \underline{0.1545} & 0.0869 & 0.1384 & 0.1384 & 0.0884 & 0.1144 & 0.0248 & 0.1508 & 0.1462 & \textbf{0.1724} \\
Parenchymal Opacities$^\dagger$ & \textbf{0.0386} & 0.0139 & 0.0139 & 0.0206 & 0.0126 & \underline{0.0255} & 0.0229 & 0.0217 & 0.0168 & 0.0209 \\
Nodules & 0.1140 & 0.1039 & \underline{0.2721} & 0.2329 & 0.0906 & 0.1124 & 0.0725 & 0.2306 & 0.2164 & \textbf{0.2727} \\
Pleural Effusion & 0.8558 & 0.8127 & \textbf{0.8902} & \underline{0.8892} & 0.7774 & 0.8552 & 0.7057 & 0.8850 & 0.8715 & 0.8891 \\
Congestive Heart Failure* & 0.0218 & 0.0234 & 0.0236 & 0.0355 & 0.0172 & 0.0180 & 0.0058 & \textbf{0.0602} & 0.0362 & \underline{0.0431} \\
Prosthetic Valve* & 0.0159 & 0.0152 & 0.0172 & \textbf{0.0188} & 0.0122 & 0.0111 & 0.0044 & 0.0169 & \underline{0.0182} & 0.0155 \\
\end{longtable}
\end{landscape}

\begin{landscape}
\subsection{Sensitivity}

\scriptsize
\setlength{\tabcolsep}{3pt}
\renewcommand{\arraystretch}{1.08}
\setlength\LTcapwidth{\linewidth}
\setlength\LTleft{0pt}
\setlength\LTright{0pt}

\begin{longtable}{l|cccccccccc}
\caption{Per-label \textbf{Sensitivity} comparison across baseline, contrastive, and autoregressive models on the CheXpert dataset. Tail abnormalities are marked with an asterisk (*), and medium-prevalence abnormalities are marked with a dagger ($^\dagger$). For each row, the best value is shown in bold and the second-best value is underlined.}
\label{tab:per_label_sensitivity_chexpert}\\

\toprule
\textbf{Abnormality} & \textbf{EVA-Base} & \textbf{RAD-DINO} & \textbf{ARK} & \textbf{CheXFound} & \textbf{BiomedCLIP} & \textbf{MedKLIP} & \textbf{BioViL-T} & \textbf{Med-CLIP} & \makecell[c]{\textbf{Med-AR-2B}\\[-0.4ex]\textbf{(ours)}} & \makecell[c]{\textbf{Med-AR-8B}\\[-0.4ex]\textbf{(ours)}} \\
\midrule
\endfirsthead

\toprule
\textbf{Abnormality} & \textbf{EVA-Base} & \textbf{RAD-DINO} & \textbf{ARK} & \textbf{CheXFound} & \textbf{BiomedCLIP} & \textbf{MedKLIP} & \textbf{BioViL-T} & \textbf{Med-CLIP} & \makecell[c]{\textbf{Med-AR-2B}\\[-0.4ex]\textbf{(ours)}} & \makecell[c]{\textbf{Med-AR-8B}\\[-0.4ex]\textbf{(ours)}} \\
\midrule
\endhead

\midrule
\multicolumn{11}{r}{\scriptsize Continued on next page} \\
\endfoot

\bottomrule
\endlastfoot

LVAD* & 0.9038 & 0.9423 & \textbf{1.0000} & 0.9615 & 0.9038 & 0.8846 & 0.8269 & 0.9231 & 0.9231 & \underline{0.9808} \\
Diaphragm elevation & 0.7598 & 0.6520 & 0.7917 & \underline{0.8284} & 0.5858 & 0.6324 & \textbf{0.8701} & 0.8064 & 0.8039 & 0.8260 \\
Osteopenia$^\dagger$ & 0.8364 & 0.8121 & 0.8000 & \underline{0.9091} & 0.8182 & 0.8182 & 0.7818 & 0.8727 & 0.8545 & \textbf{0.9394} \\
Parenchymal opacities$^\dagger$ & \textbf{0.9286} & \underline{0.8929} & 0.8036 & 0.6964 & 0.7143 & \underline{0.8929} & 0.7679 & \underline{0.8929} & 0.6964 & 0.6250 \\
Nodules & 0.6770 & \textbf{0.8217} & 0.7855 & 0.7726 & 0.7829 & 0.6977 & 0.7054 & 0.7519 & 0.6693 & \underline{0.7933} \\
Pleural effusion & 0.8198 & 0.7939 & \underline{0.8434} & \textbf{0.8633} & 0.7847 & 0.7970 & 0.8062 & 0.8266 & 0.8227 & 0.8292 \\
Congestive heart failure* & 0.8108 & 0.6216 & 0.7568 & 0.7568 & 0.6757 & \textbf{0.9459} & 0.8108 & \underline{0.9189} & 0.8378 & 0.6757 \\
Prosthetic valve* & 0.9091 & \underline{0.9394} & \textbf{0.9697} & 0.8788 & 0.6061 & 0.9091 & 0.5152 & 0.8485 & 0.9091 & 0.9091 \\
\end{longtable}
\end{landscape}

\begin{landscape}
\subsection{Specificity}

\scriptsize
\setlength{\tabcolsep}{3pt}
\renewcommand{\arraystretch}{1.08}
\setlength\LTcapwidth{\linewidth}
\setlength\LTleft{0pt}
\setlength\LTright{0pt}

\begin{longtable}{l|cccccccccc}
\caption{Per-label \textbf{Specificity} comparison across baseline, contrastive, and autoregressive models on the CheXpert dataset. Tail abnormalities are marked with an asterisk (*), and medium-prevalence abnormalities are marked with a dagger ($^\dagger$). For each row, the best value is shown in bold and the second-best value is underlined.}
\label{tab:per_label_specificity_chexpert}\\

\toprule
\textbf{Abnormality} & \textbf{EVA-Base} & \textbf{RAD-DINO} & \textbf{ARK} & \textbf{CheXFound} & \textbf{BiomedCLIP} & \textbf{MedKLIP} & \textbf{BioViL-T} & \textbf{Med-CLIP} & \makecell[c]{\textbf{Med-AR-2B}\\[-0.4ex]\textbf{(ours)}} & \makecell[c]{\textbf{Med-AR-8B}\\[-0.4ex]\textbf{(ours)}} \\
\midrule
\endfirsthead

\toprule
\textbf{Abnormality} & \textbf{EVA-Base} & \textbf{RAD-DINO} & \textbf{ARK} & \textbf{CheXFound} & \textbf{BiomedCLIP} & \textbf{MedKLIP} & \textbf{BioViL-T} & \textbf{Med-CLIP} & \makecell[c]{\textbf{Med-AR-2B}\\[-0.4ex]\textbf{(ours)}} & \makecell[c]{\textbf{Med-AR-8B}\\[-0.4ex]\textbf{(ours)}} \\
\midrule
\endhead

\midrule
\multicolumn{11}{r}{\scriptsize Continued on next page} \\
\endfoot

\bottomrule
\endlastfoot

LVAD* & 0.8065 & 0.7863 & 0.8010 & \underline{0.9075} & 0.7148 & 0.8121 & 0.4776 & 0.8730 & 0.8487 & \textbf{0.9339} \\
Diaphragm elevation & 0.6971 & 0.6838 & 0.7580 & 0.8405 & 0.7724 & 0.7352 & 0.2505 & \underline{0.8470} & 0.8271 & \textbf{0.8646} \\
Osteopenia$^\dagger$ & \underline{0.8344} & 0.7647 & \textbf{0.8530} & 0.7553 & 0.7418 & 0.7813 & 0.5046 & 0.8146 & 0.8315 & 0.7632 \\
Parenchymal opacities$^\dagger$ & 0.4206 & 0.5634 & 0.6590 & \underline{0.6801} & 0.5716 & 0.5945 & 0.6355 & 0.5439 & 0.6716 & \textbf{0.8011} \\
Nodules & 0.7293 & 0.5739 & \underline{0.7430} & 0.7044 & 0.6098 & 0.6963 & 0.6669 & 0.7201 & \textbf{0.7802} & 0.7214 \\
Pleural effusion & 0.7838 & 0.7552 & 0.8211 & 0.8014 & 0.7058 & 0.8035 & 0.6161 & \textbf{0.8244} & 0.8040 & \underline{0.8214} \\
Congestive heart failure* & 0.6798 & 0.8383 & 0.8273 & \underline{0.8566} & 0.7916 & 0.5695 & 0.3998 & 0.6965 & 0.8279 & \textbf{0.8973} \\
Prosthetic valve* & 0.7597 & 0.7262 & 0.7028 & 0.7465 & \underline{0.7734} & 0.7253 & 0.6287 & \textbf{0.7924} & 0.6649 & 0.7571 \\
\end{longtable}
\end{landscape}

\begin{landscape}
\subsection{Youden's \texorpdfstring{$J$}{J}}

\scriptsize
\setlength{\tabcolsep}{3pt}
\renewcommand{\arraystretch}{1.08}
\setlength\LTcapwidth{\linewidth}
\setlength\LTleft{0pt}
\setlength\LTright{0pt}

\begin{longtable}{l|cccccccccc}
\caption{Per-label \textbf{Youden's J Score} comparison across baseline, contrastive, and autoregressive models on the CheXpert dataset. Tail abnormalities are marked with an asterisk (*), and medium-prevalence abnormalities are marked with a dagger ($^\dagger$). For each row, the best value is shown in bold and the second-best value is underlined.}
\label{tab:per_label_youden_chexpert}\\

\toprule
\textbf{Abnormality} & \textbf{EVA-Base} & \textbf{RAD-DINO} & \textbf{ARK} & \textbf{CheXFound} & \textbf{BiomedCLIP} & \textbf{MedKLIP} & \textbf{BioViL-T} & \textbf{Med-CLIP} & \makecell[c]{\textbf{Med-AR-2B}\\[-0.4ex]\textbf{(ours)}} & \makecell[c]{\textbf{Med-AR-8B}\\[-0.4ex]\textbf{(ours)}} \\
\midrule
\endfirsthead

\toprule
\textbf{Abnormality} & \textbf{EVA-Base} & \textbf{RAD-DINO} & \textbf{ARK} & \textbf{CheXFound} & \textbf{BiomedCLIP} & \textbf{MedKLIP} & \textbf{BioViL-T} & \textbf{Med-CLIP} & \makecell[c]{\textbf{Med-AR-2B}\\[-0.4ex]\textbf{(ours)}} & \makecell[c]{\textbf{Med-AR-8B}\\[-0.4ex]\textbf{(ours)}} \\
\midrule
\endhead

\midrule
\multicolumn{11}{r}{\scriptsize Continued on next page} \\
\endfoot

\bottomrule
\endlastfoot

LVAD* & 0.7103 & 0.7286 & 0.8010 & \underline{0.8690} & 0.6186 & 0.6967 & 0.3045 & 0.7961 & 0.7718 & \textbf{0.9147} \\
Diaphragm elevation & 0.4569 & 0.3358 & 0.5497 & \underline{0.6689} & 0.3582 & 0.3676 & 0.1206 & 0.6534 & 0.6310 & \textbf{0.6906} \\
Osteopenia$^\dagger$ & 0.6708 & 0.5768 & 0.6530 & 0.6644 & 0.5600 & 0.5995 & 0.2864 & \underline{0.6873} & 0.6860 & \textbf{0.7026} \\
Parenchymal opacities$^\dagger$ & 0.3492 & 0.4563 & \underline{0.4626} & 0.3765 & 0.2859 & \textbf{0.4874} & 0.4034 & 0.4368 & 0.3680 & 0.4261 \\
Nodules & 0.4063 & 0.3956 & \textbf{0.5285} & 0.4770 & 0.3927 & 0.3940 & 0.3723 & 0.4720 & 0.4495 & \underline{0.5147} \\
Pleural effusion & 0.6036 & 0.5491 & \underline{0.6645} & \textbf{0.6647} & 0.4905 & 0.6005 & 0.4223 & 0.6510 & 0.6267 & 0.6506 \\
Congestive heart failure* & 0.4906 & 0.4599 & 0.5841 & 0.6134 & 0.4673 & 0.5154 & 0.2106 & \underline{0.6154} & \textbf{0.6657} & 0.5730 \\
Prosthetic valve* & \underline{0.6688} & 0.6656 & \textbf{0.6725} & 0.6253 & 0.3795 & 0.6344 & 0.1439 & 0.6409 & 0.5740 & 0.6662 \\
\end{longtable}
\end{landscape}

\clearpage
\begin{landscape}
\section{MIMIC-CXR: Per-Label Results}\label{supp:mimic}
The evaluated MIMIC-CXR vocabulary contains 110 labels. The available per-label results below cover 108 labels, in the order AUROC, AUPRC, sensitivity, specificity, and Youden's $J$; results for the remaining two labels are not included in these listings.

\subsection{AUROC}

\scriptsize
\setlength{\tabcolsep}{3pt}
\renewcommand{\arraystretch}{1.08}
\setlength\LTcapwidth{\linewidth}
\setlength\LTleft{0pt}
\setlength\LTright{0pt}

\begin{longtable}{l|cccccccccc}
\caption{Per-label AUROC comparison across baseline, contrastive, and autoregressive models on the MIMIC-CXR dataset. Tail abnormalities are marked with an asterisk (*), and medium-prevalence abnormalities are marked with a dagger ($\dagger$). For each row, the best value is shown in bold and the second-best value is underlined.} 
\label{tab:per_label_auroc_all_models_mimic}\\

\toprule
\multirow{2}{*}{\textbf{Abnormality}} & \textbf{EVA-Base} & \textbf{RAD-DINO} & \textbf{ARK} & \textbf{CheXFound} & \textbf{BiomedCLIP} & \textbf{MedKLIP} & \textbf{BioViL-T} & \textbf{Med-CLIP} & \makecell[c]{\textbf{Med-AR-2B}\\[-0.4ex]\textbf{(ours)}} & \makecell[c]{\textbf{Med-AR-8B}\\[-0.4ex]\textbf{(ours)}} \\
\midrule
\endfirsthead

\toprule
\multirow{2}{*}{\textbf{Abnormality}} & \textbf{EVA-Base} & \textbf{RAD-DINO} & \textbf{ARK} & \textbf{CheXFound} & \textbf{BiomedCLIP} & \textbf{MedKLIP} & \textbf{BioViL-T} & \textbf{Med-CLIP} & \makecell[c]{\textbf{Med-AR-2B}\\[-0.4ex]\textbf{(ours)}} & \makecell[c]{\textbf{Med-AR-8B}\\[-0.4ex]\textbf{(ours)}} \\
\midrule
\endhead

\midrule
\multicolumn{11}{r}{\scriptsize Continued on next page} \\
\endfoot

\bottomrule
\endlastfoot
Diaphragm Elevation & 0.8174 & 0.8314 & 0.8680 & 0.9072 & 0.8812 & 0.6922 & 0.5298 & 0.9076 & \underline{0.9209} & \textbf{0.9251} \\
Parenchymal Opacities$^\dagger$ & 0.8764 & 0.8839 & \textbf{0.8993} & 0.8904 & 0.8877 & 0.8674 & 0.7626 & 0.8869 & 0.8893 & \underline{0.8975} \\
Nodules & 0.7210 & 0.7590 & \textbf{0.8355} & 0.8155 & 0.7608 & 0.7351 & 0.5837 & 0.7949 & 0.8038 & \underline{0.8303} \\
Pleural Effusion & 0.9117 & 0.9215 & \textbf{0.9339} & \underline{0.9327} & 0.9134 & 0.9150 & 0.8240 & 0.9287 & 0.9303 & 0.9321 \\
Congestive Heart Failure* & 0.8128 & 0.8403 & \underline{0.8747} & 0.8410 & 0.8649 & 0.7886 & 0.6731 & \textbf{0.8774} & 0.8726 & 0.8725 \\
Support Devices & 0.9173 & 0.9198 & 0.9325 & 0.9317 & 0.9246 & 0.9069 & 0.8886 & 0.9303 & \underline{0.9348} & \textbf{0.9371} \\
PICC Line & 0.7570 & 0.8739 & \underline{0.9749} & 0.9723 & 0.9571 & 0.7100 & 0.6885 & 0.9389 & 0.9703 & \textbf{0.9754} \\
Pleural Thickening & 0.7481 & 0.7853 & 0.8410 & 0.8406 & 0.7738 & 0.7292 & 0.6119 & 0.8408 & \underline{0.8433} & \textbf{0.8538} \\
\end{longtable}
\end{landscape}

\begin{landscape}
\subsection{AUPRC}

\scriptsize
\setlength{\tabcolsep}{3pt}
\renewcommand{\arraystretch}{1.08}
\setlength\LTcapwidth{\linewidth}
\setlength\LTleft{0pt}
\setlength\LTright{0pt}

\begin{longtable}{l|cccccccccc}
\caption{Per-label AUPRC comparison across baseline, contrastive, and autoregressive models on the MIMIC-CXR dataset. Tail abnormalities are marked with an asterisk (*), and medium-prevalence abnormalities are marked with a dagger ($^\dagger$). For each row, the best value is shown in bold and the second-best value is underlined.}
\label{tab:per_label_auprc_all_models_mimic}\\

\toprule
\textbf{Abnormality} & \textbf{EVA-Base} & \textbf{RAD-DINO} & \textbf{ARK} & \textbf{CheXFound} & \textbf{BiomedCLIP} & \textbf{MedKLIP} & \textbf{BioViL-T} & \textbf{Med-CLIP} & \makecell[c]{\textbf{Med-AR-2B}\\[-0.4ex]\textbf{(ours)}} & \makecell[c]{\textbf{Med-AR-8B}\\[-0.4ex]\textbf{(ours)}} \\
\midrule
\endfirsthead

\toprule
\textbf{Abnormality} & \textbf{EVA-Base} & \textbf{RAD-DINO} & \textbf{ARK} & \textbf{CheXFound} & \textbf{BiomedCLIP} & \textbf{MedKLIP} & \textbf{BioViL-T} & \textbf{Med-CLIP} & \makecell[c]{\textbf{Med-AR-2B}\\[-0.4ex]\textbf{(ours)}} & \makecell[c]{\textbf{Med-AR-8B}\\[-0.4ex]\textbf{(ours)}} \\
\midrule
\endhead

\midrule
\multicolumn{11}{r}{\scriptsize Continued on next page} \\
\endfoot

\bottomrule
\endlastfoot
Diaphragm Elevation & 0.2024 & 0.2282 & 0.3259 & 0.4529 & 0.3923 & 0.0722 & 0.0334 & 0.4595 & \underline{0.4844} & \textbf{0.4967} \\
Parenchymal Opacities$^\dagger$ & 0.0581 & 0.0552 & \textbf{0.0691} & 0.0632 & 0.0583 & 0.0502 & 0.0192 & 0.0549 & 0.0630 & \underline{0.0633} \\
Nodules & 0.0991 & 0.1335 & \textbf{0.3179} & 0.2356 & 0.1577 & 0.1118 & 0.0422 & 0.2503 & 0.2624 & \underline{0.2965} \\
Pleural Effusion & 0.8302 & 0.8463 & \textbf{0.8706} & \underline{0.8682} & 0.8332 & 0.8343 & 0.6555 & 0.8605 & 0.8663 & 0.8674 \\
Congestive Heart Failure* & 0.0179 & 0.0190 & \textbf{0.0242} & 0.0186 & 0.0197 & 0.0113 & 0.0049 & \underline{0.0241} & 0.0210 & 0.0220 \\
Support Devices & 0.1770 & 0.1640 & 0.1957 & \underline{0.2010} & 0.1856 & 0.1477 & 0.1401 & 0.1972 & 0.1978 & \textbf{0.2053} \\
Picc Line & 0.1270 & 0.2494 & \textbf{0.6183} & 0.6005 & 0.5522 & 0.0928 & 0.0862 & 0.4965 & 0.6083 & \underline{0.6171} \\
Pleural Thickening & 0.0715 & 0.0999 & 0.1886 & 0.2010 & 0.1159 & 0.0725 & 0.0373 & 0.2072 & \underline{0.2294} & \textbf{0.2429} \\
\end{longtable}
\end{landscape}

\begin{landscape}
\subsection{Sensitivity}

\scriptsize
\setlength{\tabcolsep}{3pt}
\renewcommand{\arraystretch}{1.08}
\setlength\LTcapwidth{\linewidth}
\setlength\LTleft{0pt}
\setlength\LTright{0pt}

\begin{longtable}{l|cccccccccc}
\caption{Per-label \textbf{Sensitivity} comparison across baseline, contrastive, and autoregressive models on the \textbf{MIMIC} dataset. Tail abnormalities are marked with an asterisk (*), and medium-prevalence abnormalities are marked with a dagger ($^\dagger$). For each row, the best value is shown in bold and the second-best value is underlined.}
\label{tab:per_label_sensitivity_mimic}\\

\toprule
\textbf{Abnormality} & \textbf{EVA-Base} & \textbf{RAD-DINO} & \textbf{ARK} & \textbf{CheXFound} & \textbf{BiomedCLIP} & \textbf{MedKLIP} & \textbf{BioViL-T} & \textbf{Med-CLIP} & \makecell[c]{\textbf{Med-AR-2B}\\[-0.4ex]\textbf{(ours)}} & \makecell[c]{\textbf{Med-AR-8B}\\[-0.4ex]\textbf{(ours)}} \\
\midrule
\endfirsthead

\toprule
\textbf{Abnormality} & \textbf{EVA-Base} & \textbf{RAD-DINO} & \textbf{ARK} & \textbf{CheXFound} & \textbf{BiomedCLIP} & \textbf{MedKLIP} & \textbf{BioViL-T} & \textbf{Med-CLIP} & \makecell[c]{\textbf{Med-AR-2B}\\[-0.4ex]\textbf{(ours)}} & \makecell[c]{\textbf{Med-AR-8B}\\[-0.4ex]\textbf{(ours)}} \\
\midrule
\endhead

\midrule
\multicolumn{11}{r}{\scriptsize Continued on next page} \\
\endfoot

\bottomrule
\endlastfoot

Diaphragm elevation & 0.7113 & 0.7307 & 0.7942 & 0.8108 & 0.7555 & 0.7320 & \textbf{0.9171} & 0.8177 & 0.7887 & \underline{0.8522} \\
Parenchymal opacities$^\dagger$ & 0.8623 & \underline{0.8922} & 0.8802 & \textbf{0.9222} & \underline{0.8922} & 0.8144 & 0.7605 & 0.8443 & 0.8802 & 0.8323 \\
Nodules & 0.7195 & 0.6697 & 0.6868 & \underline{0.7274} & 0.7235 & 0.6776 & 0.5531 & 0.6239 & 0.6645 & \textbf{0.7471} \\
Pleural effusion & 0.8516 & 0.8709 & \textbf{0.8747} & 0.8538 & \underline{0.8735} & 0.8706 & 0.7845 & 0.8671 & 0.8535 & 0.8665 \\
Congestive heart failure* & 0.8095 & 0.8254 & 0.8571 & 0.8730 & 0.7619 & \textbf{0.9206} & \textbf{0.9206} & \underline{0.8889} & 0.8413 & 0.8571 \\
Support devices & 0.9427 & 0.9301 & 0.9516 & 0.9498 & 0.9301 & 0.9409 & 0.9480 & 0.9570 & \underline{0.9642} & \textbf{0.9749} \\
PICC line & 0.7813 & 0.8960 & \underline{0.9595} & 0.9538 & 0.9143 & 0.7861 & 0.7813 & 0.8796 & 0.9566 & \textbf{0.9624} \\
Pleural thickening & 0.6724 & \underline{0.7966} & 0.7483 & 0.7414 & 0.6776 & 0.6138 & 0.7845 & 0.7914 & 0.7879 & \textbf{0.8000} \\
\end{longtable}
\end{landscape}

\begin{landscape}
\subsection{Specificity}

\scriptsize
\setlength{\tabcolsep}{3pt}
\renewcommand{\arraystretch}{1.08}
\setlength\LTcapwidth{\linewidth}
\setlength\LTleft{0pt}
\setlength\LTright{0pt}

\begin{longtable}{l|cccccccccc}
\caption{Per-label \textbf{Specificity} comparison across baseline, contrastive, and autoregressive models on the \textbf{MIMIC} dataset. Tail abnormalities are marked with an asterisk (*), and medium-prevalence abnormalities are marked with a dagger ($^\dagger$). For each row, the best value is shown in bold and the second-best value is underlined.}
\label{tab:per_label_specificity_mimic}\\

\toprule
\textbf{Abnormality} & \textbf{EVA-Base} & \textbf{RAD-DINO} & \textbf{ARK} & \textbf{CheXFound} & \textbf{BiomedCLIP} & \textbf{MedKLIP} & \textbf{BioViL-T} & \textbf{Med-CLIP} & \makecell[c]{\textbf{Med-AR-2B}\\[-0.4ex]\textbf{(ours)}} & \makecell[c]{\textbf{Med-AR-8B}\\[-0.4ex]\textbf{(ours)}} \\
\midrule
\endfirsthead

\toprule
\textbf{Abnormality} & \textbf{EVA-Base} & \textbf{RAD-DINO} & \textbf{ARK} & \textbf{CheXFound} & \textbf{BiomedCLIP} & \textbf{MedKLIP} & \textbf{BioViL-T} & \textbf{Med-CLIP} & \makecell[c]{\textbf{Med-AR-2B}\\[-0.4ex]\textbf{(ours)}} & \makecell[c]{\textbf{Med-AR-8B}\\[-0.4ex]\textbf{(ours)}} \\
\midrule
\endhead

\midrule
\multicolumn{11}{r}{\scriptsize Continued on next page} \\
\endfoot

\bottomrule
\endlastfoot

Diaphragm elevation & 0.7651 & 0.7826 & 0.7739 & 0.8533 & 0.8421 & 0.5545 & 0.2131 & 0.8440 & \textbf{0.9018} & \underline{0.8586} \\
Parenchymal opacities$^\dagger$ & 0.7392 & 0.7465 & \underline{0.8114} & 0.7207 & 0.7516 & 0.7588 & 0.6673 & 0.7963 & 0.7685 & \textbf{0.8239} \\
Nodules & 0.6128 & 0.7120 & \underline{0.8134} & 0.7440 & 0.6781 & 0.6684 & 0.5918 & \textbf{0.8142} & 0.7951 & 0.7553 \\
Pleural effusion & 0.8140 & 0.8133 & 0.8389 & \textbf{0.8584} & 0.7923 & 0.8009 & 0.7095 & 0.8325 & \underline{0.8493} & 0.8377 \\
Congestive heart failure* & 0.7424 & 0.7545 & 0.7655 & 0.6752 & \textbf{0.8219} & 0.5473 & 0.4684 & 0.7504 & \underline{0.7846} & 0.7823 \\
Support devices & 0.7846 & 0.8139 & \textbf{0.8325} & 0.8189 & \underline{0.8283} & 0.7751 & 0.7241 & 0.8077 & 0.8251 & 0.8110 \\
PICC line & 0.6081 & 0.7148 & \underline{0.9348} & 0.9337 & 0.9065 & 0.5338 & 0.5221 & 0.8650 & 0.9320 & \textbf{0.9350} \\
Pleural thickening & 0.7081 & 0.6473 & \underline{0.7937} & \textbf{0.8046} & 0.7616 & 0.7330 & 0.3895 & 0.7374 & 0.7692 & 0.7650 \\
\end{longtable}
\end{landscape}

\begin{landscape}
\subsection{Youden's \texorpdfstring{$J$}{J}}

\scriptsize
\setlength{\tabcolsep}{3pt}
\renewcommand{\arraystretch}{1.08}
\setlength\LTcapwidth{\linewidth}
\setlength\LTleft{0pt}
\setlength\LTright{0pt}

\begin{longtable}{l|cccccccccc}
\caption{Per-label \textbf{Youden's J Score} comparison across baseline, contrastive, and autoregressive models on the \textbf{MIMIC} dataset. Tail abnormalities are marked with an asterisk (*), and medium-prevalence abnormalities are marked with a dagger ($^\dagger$). For each row, the best value is shown in bold and the second-best value is underlined.}
\label{tab:per_label_youden_mimic}\\

\toprule
\textbf{Abnormality} & \textbf{EVA-Base} & \textbf{RAD-DINO} & \textbf{ARK} & \textbf{CheXFound} & \textbf{BiomedCLIP} & \textbf{MedKLIP} & \textbf{BioViL-T} & \textbf{Med-CLIP} & \makecell[c]{\textbf{Med-AR-2B}\\[-0.4ex]\textbf{(ours)}} & \makecell[c]{\textbf{Med-AR-8B}\\[-0.4ex]\textbf{(ours)}} \\
\midrule
\endfirsthead

\toprule
\textbf{Abnormality} & \textbf{EVA-Base} & \textbf{RAD-DINO} & \textbf{ARK} & \textbf{CheXFound} & \textbf{BiomedCLIP} & \textbf{MedKLIP} & \textbf{BioViL-T} & \textbf{Med-CLIP} & \makecell[c]{\textbf{Med-AR-2B}\\[-0.4ex]\textbf{(ours)}} & \makecell[c]{\textbf{Med-AR-8B}\\[-0.4ex]\textbf{(ours)}} \\
\midrule
\endhead

\midrule
\multicolumn{11}{r}{\scriptsize Continued on next page} \\
\endfoot

\bottomrule
\endlastfoot

Diaphragm elevation & 0.4764 & 0.5133 & 0.5681 & 0.6641 & 0.5976 & 0.2865 & 0.1302 & 0.6617 & \underline{0.6905} & \textbf{0.7108} \\
Parenchymal opacities$^\dagger$ & 0.6015 & 0.6387 & \textbf{0.6916} & 0.6429 & 0.6438 & 0.5732 & 0.4278 & 0.6406 & 0.6487 & \underline{0.6562} \\
Nodules & 0.3323 & 0.3817 & \underline{0.5002} & 0.4714 & 0.4016 & 0.3460 & 0.1449 & 0.4381 & 0.4596 & \textbf{0.5024} \\
Pleural effusion & 0.6656 & 0.6842 & \textbf{0.7136} & \underline{0.7122} & 0.6658 & 0.6715 & 0.4940 & 0.6996 & 0.7028 & 0.7042 \\
Congestive heart failure* & 0.5519 & 0.5799 & 0.6226 & 0.5482 & 0.5838 & 0.4679 & 0.3890 & \underline{0.6393} & 0.6259 & \textbf{0.6394} \\
Support devices & 0.7273 & 0.7440 & 0.7841 & 0.7687 & 0.7584 & 0.7160 & 0.6721 & 0.7647 & \textbf{0.7893} & \underline{0.7859} \\
PICC line & 0.3894 & 0.6108 & \underline{0.8943} & 0.8875 & 0.8208 & 0.3199 & 0.3034 & 0.7446 & 0.8886 & \textbf{0.8974} \\
Pleural thickening & 0.3805 & 0.4439 & 0.5420 & 0.5460 & 0.4392 & 0.3468 & 0.1740 & 0.5288 & \underline{0.5571} & \textbf{0.5650} \\
\end{longtable}
\end{landscape}

\clearpage
\begin{landscape}
\section{PadChest: Per-Label Results}\label{supp:padchest}
The following tables report the existing results for 29 labels, in the order AUROC, AUPRC, sensitivity, specificity, and Youden's $J$.

\subsection{AUROC}

\scriptsize
\setlength{\tabcolsep}{3pt}
\renewcommand{\arraystretch}{1.08}
\setlength\LTcapwidth{\linewidth}
\setlength\LTleft{0pt}
\setlength\LTright{0pt}

\begin{longtable}{l|cccccccccc}
\caption{Per-label AUROC comparison across baseline, contrastive, and autoregressive models on the PadChest dataset. Tail abnormalities are marked with an asterisk (*), and medium-prevalence abnormalities are marked with a dagger ($\dagger$). For each row, the best value is shown in bold and the second-best value is underlined.}
\label{tab:per_label_auroc_padchest_all_models}\\

\toprule
\multirow{2}{*}{\textbf{Abnormality}} & \textbf{EVA-Base} & \textbf{RAD-DINO} & \textbf{ARK} & \textbf{CheXFound} & \textbf{BiomedCLIP} & \textbf{MedKLIP} & \textbf{BioViL-T} & \textbf{Med-CLIP} & \makecell[c]{\textbf{Med-AR}\\[-0.4ex]\textbf{(ours)}} & \makecell[c]{\textbf{Med-AR-8B}\\[-0.4ex]\textbf{(ours)}} \\
\midrule
\endfirsthead

\toprule
\multirow{2}{*}{\textbf{Abnormality}} & \textbf{EVA-Base} & \textbf{RAD-DINO} & \textbf{ARK} & \textbf{CheXFound} & \textbf{BiomedCLIP} & \textbf{MedKLIP} & \textbf{BioViL-T} & \textbf{Med-CLIP} & \makecell[c]{\textbf{Med-AR-2B}\\[-0.4ex]\textbf{(ours)}} & \makecell[c]{\textbf{Med-AR-8B}\\[-0.4ex]\textbf{(ours)}} \\
\midrule
\endhead

\midrule
\multicolumn{11}{r}{\scriptsize Continued on next page} \\
\endfoot

\bottomrule
\endlastfoot
NSG Tube$^\dagger$ & 0.9695 & 0.9685 & 0.9925 & \underline{0.9928} & 0.9291 & 0.9888 & 0.9653 & 0.9899 & \textbf{0.9934} & 0.9919 \\
Volume Loss* & 0.7783 & 0.7486 & 0.9117 & 0.9137 & 0.6554 & 0.8210 & 0.5479 & 0.9087 & \textbf{0.9226} & \underline{0.9163} \\
Kyphosis$^\dagger$ & 0.8535 & 0.8203 & 0.8683 & \underline{0.9009} & 0.6513 & 0.8491 & 0.5987 & 0.8914 & 0.8988 & \textbf{0.9029} \\
Alveolar Pattern$^\dagger$ & 0.8863 & 0.8728 & \textbf{0.9278} & 0.9270 & 0.7708 & 0.9055 & 0.8291 & 0.9219 & \underline{0.9277} & 0.9204 \\
Pacemaker* & 0.9851 & 0.9819 & \underline{0.9967} & 0.9954 & 0.9244 & 0.9944 & 0.9641 & 0.9954 & \textbf{0.9969} & 0.9955 \\
COPD Signs & 0.8174 & 0.8084 & 0.8324 & 0.8386 & 0.6645 & 0.8168 & 0.6061 & 0.8318 & \textbf{0.8436} & \underline{0.8407} \\
Infiltrates & 0.7314 & 0.7132 & \underline{0.8069} & 0.8068 & 0.6065 & 0.7658 & 0.6513 & 0.7880 & \textbf{0.8125} & 0.8001 \\
Vertebral Degenerative Changes$^\dagger$ & 0.7564 & 0.7351 & 0.7897 & \underline{0.8368} & 0.6221 & 0.7892 & 0.5485 & 0.8247 & \textbf{0.8484} & 0.8366 \\
\end{longtable}
\end{landscape}

\begin{landscape}
\subsection{AUPRC}

\scriptsize
\setlength{\tabcolsep}{3pt}
\renewcommand{\arraystretch}{1.08}
\setlength\LTcapwidth{\linewidth}
\setlength\LTleft{0pt}
\setlength\LTright{0pt}

\begin{longtable}{l|cccccccccc}
\caption{Per-label AUPRC comparison across baseline, contrastive, and autoregressive models on the PadChest dataset. Tail abnormalities are marked with an asterisk (*), and medium-prevalence abnormalities are marked with a dagger ($^\dagger$). For each row, the best value is shown in bold and the second-best value is underlined.}
\label{tab:per_label_auprc_all_models_padchest}\\

\toprule
\textbf{Abnormality} & \textbf{EVA-Base} & \textbf{RAD-DINO} & \textbf{ARK} & \textbf{CheXFound} & \textbf{BiomedCLIP} & \textbf{MedKLIP} & \textbf{BioViL-T} & \textbf{Med-CLIP} & \makecell[c]{\textbf{Med-AR-2B}\\[-0.4ex]\textbf{(ours)}} & \makecell[c]{\textbf{Med-AR-8B}\\[-0.4ex]\textbf{(ours)}} \\
\midrule
\endfirsthead

\toprule
\textbf{Abnormality} & \textbf{EVA-Base} & \textbf{RAD-DINO} & \textbf{ARK} & \textbf{CheXFound} & \textbf{BiomedCLIP} & \textbf{MedKLIP} & \textbf{BioViL-T} & \textbf{Med-CLIP} & \makecell[c]{\textbf{Med-AR-2B}\\[-0.4ex]\textbf{(ours)}} & \makecell[c]{\textbf{Med-AR-8B}\\[-0.4ex]\textbf{(ours)}} \\
\midrule
\endhead

\midrule
\multicolumn{11}{r}{\scriptsize Continued on next page} \\
\endfoot

\bottomrule
\endlastfoot
NSG tube$^\dagger$ & 0.5976 & 0.5641 & 0.8669 & \underline{0.8691} & 0.4898 & 0.8268 & 0.5539 & 0.8283 & \textbf{0.8728} & 0.8630 \\
Volume loss* & 0.1298 & 0.1106 & 0.3903 & 0.3851 & 0.0720 & 0.1751 & 0.0347 & 0.3933 & \textbf{0.4516} & \underline{0.4227} \\
Kyphosis$^\dagger$ & 0.3222 & 0.2504 & 0.3504 & \underline{0.4490} & 0.1112 & 0.2906 & 0.0687 & 0.4186 & \textbf{0.4494} & 0.4390 \\
Alveolar pattern$^\dagger$ & 0.4612 & 0.4173 & \textbf{0.5698} & 0.5612 & 0.2338 & 0.5154 & 0.3237 & 0.5623 & \underline{0.5651} & 0.5504 \\
Pacemaker* & 0.7318 & 0.7209 & 0.9142 & 0.9008 & 0.6551 & 0.8997 & 0.5389 & 0.9164 & \textbf{0.9388} & \underline{0.9374} \\
COPD signs & 0.5282 & 0.5107 & 0.5612 & 0.5695 & 0.3509 & 0.5288 & 0.2584 & 0.5638 & \textbf{0.5862} & \underline{0.5796} \\
Infiltrates & 0.1607 & 0.1481 & 0.2397 & \underline{0.2463} & 0.1010 & 0.1966 & 0.1115 & 0.2163 & \textbf{0.2518} & 0.2418 \\
Vertebral degenerative changes$^\dagger$ & 0.1326 & 0.1187 & 0.1683 & 0.2387 & 0.0766 & 0.1580 & 0.0542 & 0.2163 & \textbf{0.2602} & \underline{0.2461} \\
\end{longtable}
\end{landscape}

\begin{landscape}
\subsection{Sensitivity}

\scriptsize
\setlength{\tabcolsep}{3pt}
\renewcommand{\arraystretch}{1.08}
\setlength\LTcapwidth{\linewidth}
\setlength\LTleft{0pt}
\setlength\LTright{0pt}

\begin{longtable}{l|cccccccccc}
\caption{Per-label \textbf{Sensitivity} comparison across baseline, contrastive, and autoregressive models on the PadChest dataset. Tail abnormalities are marked with an asterisk (*), and medium-prevalence abnormalities are marked with a dagger ($^\dagger$). For each row, the best value is shown in bold and the second-best value is underlined.}
\label{tab:per_label_sensitivity_padchest}\\

\toprule
\textbf{Abnormality} & \textbf{EVA-Base} & \textbf{RAD-DINO} & \textbf{ARK} & \textbf{CheXFound} & \textbf{BiomedCLIP} & \textbf{MedKLIP} & \textbf{BioViL-T} & \textbf{Med-CLIP} & \makecell[c]{\textbf{Med-AR-2B}\\[-0.4ex]\textbf{(ours)}} & \makecell[c]{\textbf{Med-AR-8B}\\[-0.4ex]\textbf{(ours)}} \\
\midrule
\endfirsthead

\toprule
\textbf{Abnormality} & \textbf{EVA-Base} & \textbf{RAD-DINO} & \textbf{ARK} & \textbf{CheXFound} & \textbf{BiomedCLIP} & \textbf{MedKLIP} & \textbf{BioViL-T} & \textbf{Med-CLIP} & \makecell[c]{\textbf{Med-AR-2B}\\[-0.4ex]\textbf{(ours)}} & \makecell[c]{\textbf{Med-AR-8B}\\[-0.4ex]\textbf{(ours)}} \\
\midrule
\endhead

\midrule
\multicolumn{11}{r}{\scriptsize Continued on next page} \\
\endfoot

\bottomrule
\endlastfoot

NSG tube$^\dagger$ & 0.9554 & 0.9705 & \textbf{0.9857} & 0.9777 & 0.8375 & 0.9737 & 0.9450 & 0.9721 & \underline{0.9809} & 0.9785 \\
Volume loss* & 0.6830 & 0.7692 & 0.8276 & \underline{0.8382} & 0.5040 & 0.7944 & 0.6631 & 0.8130 & 0.8236 & \textbf{0.8554} \\
Kyphosis$^\dagger$ & 0.8060 & 0.7610 & 0.8419 & \underline{0.8659} & 0.6247 & 0.8052 & 0.8352 & 0.8479 & 0.8142 & \textbf{0.8831} \\
Alveolar pattern$^\dagger$ & 0.8368 & 0.8318 & 0.8677 & \textbf{0.8878} & 0.7549 & 0.8066 & 0.7026 & 0.8601 & 0.8570 & \underline{0.8715} \\
Pacemaker* & 0.9643 & 0.9475 & 0.9855 & 0.9788 & 0.7969 & 0.9777 & 0.9375 & 0.9810 & \underline{0.9900} & \textbf{0.9922} \\
COPD signs & \underline{0.7921} & 0.7526 & 0.7898 & 0.7789 & 0.5750 & 0.7676 & 0.7691 & 0.7855 & \textbf{0.8083} & 0.7829 \\
Infiltrates & 0.7383 & 0.7278 & \textbf{0.7925} & 0.7260 & 0.5320 & 0.7137 & 0.5665 & 0.7525 & 0.7586 & \underline{0.7629} \\
Vertebral degenerative changes$^\dagger$ & 0.8148 & 0.8131 & 0.8064 & 0.7896 & 0.6563 & 0.7519 & \textbf{0.9279} & 0.7611 & \underline{0.8399} & 0.8106 \\
\end{longtable}
\end{landscape}

\begin{landscape}
\subsection{Specificity}

\scriptsize
\setlength{\tabcolsep}{3pt}
\renewcommand{\arraystretch}{1.08}
\setlength\LTcapwidth{\linewidth}
\setlength\LTleft{0pt}
\setlength\LTright{0pt}

\begin{longtable}{l|cccccccccc}
\caption{Per-label \textbf{Specificity} comparison across baseline, contrastive, and autoregressive models on the PadChest dataset. Tail abnormalities are marked with an asterisk (*), and medium-prevalence abnormalities are marked with a dagger ($^\dagger$). For each row, the best value is shown in bold and the second-best value is underlined.}
\label{tab:per_label_specificity_padchest}\\

\toprule
\textbf{Abnormality} & \textbf{EVA-Base} & \textbf{RAD-DINO} & \textbf{ARK} & \textbf{CheXFound} & \textbf{BiomedCLIP} & \textbf{MedKLIP} & \textbf{BioViL-T} & \textbf{Med-CLIP} & \makecell[c]{\textbf{Med-AR-2B}\\[-0.4ex]\textbf{(ours)}} & \makecell[c]{\textbf{Med-AR-8B}\\[-0.4ex]\textbf{(ours)}} \\
\midrule
\endfirsthead

\toprule
\textbf{Abnormality} & \textbf{EVA-Base} & \textbf{RAD-DINO} & \textbf{ARK} & \textbf{CheXFound} & \textbf{BiomedCLIP} & \textbf{MedKLIP} & \textbf{BioViL-T} & \textbf{Med-CLIP} & \makecell[c]{\textbf{Med-AR-2B}\\[-0.4ex]\textbf{(ours)}} & \makecell[c]{\textbf{Med-AR-8B}\\[-0.4ex]\textbf{(ours)}} \\
\midrule
\endhead

\midrule
\multicolumn{11}{r}{\scriptsize Continued on next page} \\
\endfoot

\bottomrule
\endlastfoot

NSG tube$^\dagger$ & 0.9266 & 0.9147 & 0.9654 & \underline{0.9715} & 0.8855 & 0.9590 & 0.9095 & 0.9643 & \textbf{0.9723} & 0.9712 \\
Volume loss* & 0.7478 & 0.5952 & 0.8366 & 0.8373 & 0.7262 & 0.7140 & 0.4524 & \underline{0.8477} & \textbf{0.8657} & 0.8266 \\
Kyphosis$^\dagger$ & 0.7370 & 0.7264 & 0.7439 & 0.7791 & 0.6014 & 0.7316 & 0.3773 & \underline{0.7823} & \textbf{0.8284} & 0.7720 \\
Alveolar pattern$^\dagger$ & 0.7680 & 0.7514 & 0.8327 & 0.8112 & 0.6522 & \underline{0.8336} & 0.8243 & 0.8334 & \textbf{0.8459} & 0.8142 \\
Pacemaker* & 0.9373 & 0.9372 & \underline{0.9900} & 0.9850 & 0.9082 & 0.9771 & 0.8835 & 0.9887 & \textbf{0.9938} & 0.9878 \\
COPD signs & 0.6898 & 0.7143 & 0.7166 & \textbf{0.7400} & 0.6602 & 0.7105 & 0.3865 & 0.7168 & 0.7214 & \underline{0.7369} \\
Infiltrates & 0.6214 & 0.6082 & 0.6908 & \textbf{0.7482} & 0.6384 & 0.6853 & 0.6577 & 0.6896 & \underline{0.7314} & 0.6992 \\
Vertebral degenerative changes$^\dagger$ & 0.5695 & 0.5372 & 0.6248 & \underline{0.7164} & 0.5338 & 0.6851 & 0.2207 & \textbf{0.7396} & 0.6981 & 0.7126 \\
\end{longtable}
\end{landscape}

\begin{landscape}
\subsection{Youden's \texorpdfstring{$J$}{J}}

\scriptsize
\setlength{\tabcolsep}{3pt}
\renewcommand{\arraystretch}{1.08}
\setlength\LTcapwidth{\linewidth}
\setlength\LTleft{0pt}
\setlength\LTright{0pt}

\begin{longtable}{l|cccccccccc}
\caption{Per-label \textbf{Youden's J Score} comparison across baseline, contrastive, and autoregressive models on the PadChest dataset. Tail abnormalities are marked with an asterisk (*), and medium-prevalence abnormalities are marked with a dagger ($^\dagger$). For each row, the best value is shown in bold and the second-best value is underlined.}
\label{tab:per_label_youden_padchest}\\

\toprule
\textbf{Abnormality} & \textbf{EVA-Base} & \textbf{RAD-DINO} & \textbf{ARK} & \textbf{CheXFound} & \textbf{BiomedCLIP} & \textbf{MedKLIP} & \textbf{BioViL-T} & \textbf{Med-CLIP} & \makecell[c]{\textbf{Med-AR-2B}\\[-0.4ex]\textbf{(ours)}} & \makecell[c]{\textbf{Med-AR-8B}\\[-0.4ex]\textbf{(ours)}} \\
\midrule
\endfirsthead

\toprule
\textbf{Abnormality} & \textbf{EVA-Base} & \textbf{RAD-DINO} & \textbf{ARK} & \textbf{CheXFound} & \textbf{BiomedCLIP} & \textbf{MedKLIP} & \textbf{BioViL-T} & \textbf{Med-CLIP} & \makecell[c]{\textbf{Med-AR-2B}\\[-0.4ex]\textbf{(ours)}} & \makecell[c]{\textbf{Med-AR-8B}\\[-0.4ex]\textbf{(ours)}} \\
\midrule
\endhead

\midrule
\multicolumn{11}{r}{\scriptsize Continued on next page} \\
\endfoot

\bottomrule
\endlastfoot

NSG tube$^\dagger$ & 0.8820 & 0.8852 & \underline{0.9511} & 0.9492 & 0.7230 & 0.9327 & 0.8545 & 0.9364 & \textbf{0.9532} & 0.9497 \\
Volume loss* & 0.4308 & 0.3644 & 0.6642 & 0.6755 & 0.2302 & 0.5084 & 0.1155 & 0.6607 & \textbf{0.6893} & \underline{0.6820} \\
Kyphosis$^\dagger$ & 0.5430 & 0.4874 & 0.5858 & \underline{0.6450} & 0.2261 & 0.5368 & 0.2125 & 0.6302 & 0.6426 & \textbf{0.6551} \\
Alveolar pattern$^\dagger$ & 0.6048 & 0.5832 & \underline{0.7004} & 0.6990 & 0.4071 & 0.6402 & 0.5269 & 0.6935 & \textbf{0.7029} & 0.6857 \\
Pacemaker* & 0.9016 & 0.8847 & 0.9755 & 0.9638 & 0.7051 & 0.9548 & 0.8210 & 0.9697 & \textbf{0.9838} & \underline{0.9800} \\
COPD signs & 0.4819 & 0.4669 & 0.5064 & 0.5189 & 0.2352 & 0.4781 & 0.1556 & 0.5023 & \textbf{0.5297} & \underline{0.5198} \\
Infiltrates & 0.3597 & 0.3360 & \underline{0.4833} & 0.4742 & 0.1704 & 0.3990 & 0.2242 & 0.4421 & \textbf{0.4900} & 0.4621 \\
Vertebral degenerative changes$^\dagger$ & 0.3843 & 0.3503 & 0.4312 & 0.5060 & 0.1901 & 0.4370 & 0.1486 & 0.5007 & \textbf{0.5380} & \underline{0.5232} \\
\end{longtable}
\end{landscape}

\end{document}